\documentclass[final]{cvpr}

\def\cvprPaperID{9568} % *** Enter the CVPR Paper ID here
\def\confName{ICCV}
\def\confYear{2021}

\usepackage[font=footnotesize,labelfont=bf]{caption}
\usepackage[belowskip=-5pt,aboveskip=3pt]{caption} % re-arrange tables on last page, select
\usepackage{times}
\usepackage{epsfig}
\usepackage{graphicx}
\usepackage{amsmath}
\usepackage{amssymb}
\usepackage{mathtools}
\usepackage{setspace}
\usepackage{booktabs}
\usepackage{diagbox}
\usepackage{subfigure}
\usepackage{graphics} % for pdf, bitmapped graphics files
\usepackage{graphicx}
\usepackage{grffile} %file name with multi-dots
\usepackage{amsmath} % assumes amsmath package installed
\usepackage{amssymb}  % assumes amsmath package installed
\usepackage{color}
\usepackage{bm}
\usepackage{url}

\usepackage[noend]{algpseudocode}
\usepackage{algorithm}
\usepackage{makecell}
\usepackage[square, comma,numbers,sort&compress]{natbib}
\usepackage{multirow}
\usepackage{enumitem}
\usepackage[table,dvipsnames]{xcolor}
\usepackage[pagebackref=true,breaklinks=true,colorlinks,bookmarks=false]{hyperref}

\definecolor{goodgreen}{rgb}{0,0.69,0.3137}

\newcommand{\tbred}[1]{\textbf{\textcolor{red}{#1}}}
\newcommand{\tbblue}[1]{\textbf{\textcolor{blue}{#1}}}
\newcommand{\tbgreen}[1]{\textbf{\textcolor{goodgreen}{#1}}}
\hypersetup{
	linkcolor=BrickRed
	,citecolor=Green
	,filecolor=Mulberry
	,urlcolor=NavyBlue
	,menucolor=BrickRed
	,runcolor=Mulberry
	,linkbordercolor=BrickRed
	,citebordercolor=Green
	,filebordercolor=Mulberry
	,urlbordercolor=NavyBlue
	,menubordercolor=BrickRed
	,runbordercolor=Mulberry
}
\begin{document}

%%%%%%%%% TITLE
\title{Fooling LiDAR Perception via Adversarial Trajectory Perturbation} 
\author{Yiming Li${^{\dagger, *}}$ \quad Congcong Wen$^{\dagger,\S, *}$ \quad Felix Juefei-Xu$^{\ddagger}$ \quad Chen Feng$^{\dagger}$\\
	$^{\dagger}$New York University \quad $^{\S}$University of Chinese Academy of Sciences \quad $^{\ddagger}$Alibaba Group\\
	{\tt\small yimingli@nyu.edu, cfeng@nyu.edu}
}

\maketitle

\renewcommand{\thefootnote}{\fnsymbol{footnote}} %将脚注符号设置为fnsymbol类型，即特殊符号表示
\footnotetext[1]{~indicates equal contribution. The work is done during Congcong's visit at NYU. Chen Feng is the corresponding author.}
\footnotetext{}
\begin{abstract}
	LiDAR point clouds collected from a moving vehicle are functions of its trajectories, because the sensor motion needs to be compensated to avoid distortions. When autonomous vehicles are sending LiDAR point clouds to deep networks for perception and planning, could the motion compensation consequently become a wide-open backdoor in those networks, due to both the adversarial vulnerability of deep learning and GPS-based vehicle trajectory estimation that is susceptible to wireless spoofing? We demonstrate such possibilities for the first time: instead of directly attacking point cloud coordinates which requires tampering with the raw LiDAR readings, only adversarial spoofing of a self-driving car's trajectory with small perturbations is enough to make safety-critical objects undetectable or detected with incorrect positions. Moreover, polynomial trajectory perturbation is developed to achieve a temporally-smooth and highly-imperceptible attack. Extensive experiments on 3D object detection have shown that such attacks not only lower the performance of the state-of-the-art detectors effectively, but also transfer to other detectors, raising a red flag for the community. The code is available on \url{https://ai4ce.github.io/FLAT/}.
\end{abstract}

%%%%%%%%% BODY TEXT
\section{Introduction}\label{sec:intro}
\begin{figure}[t]
	\includegraphics[width=0.49\textwidth]{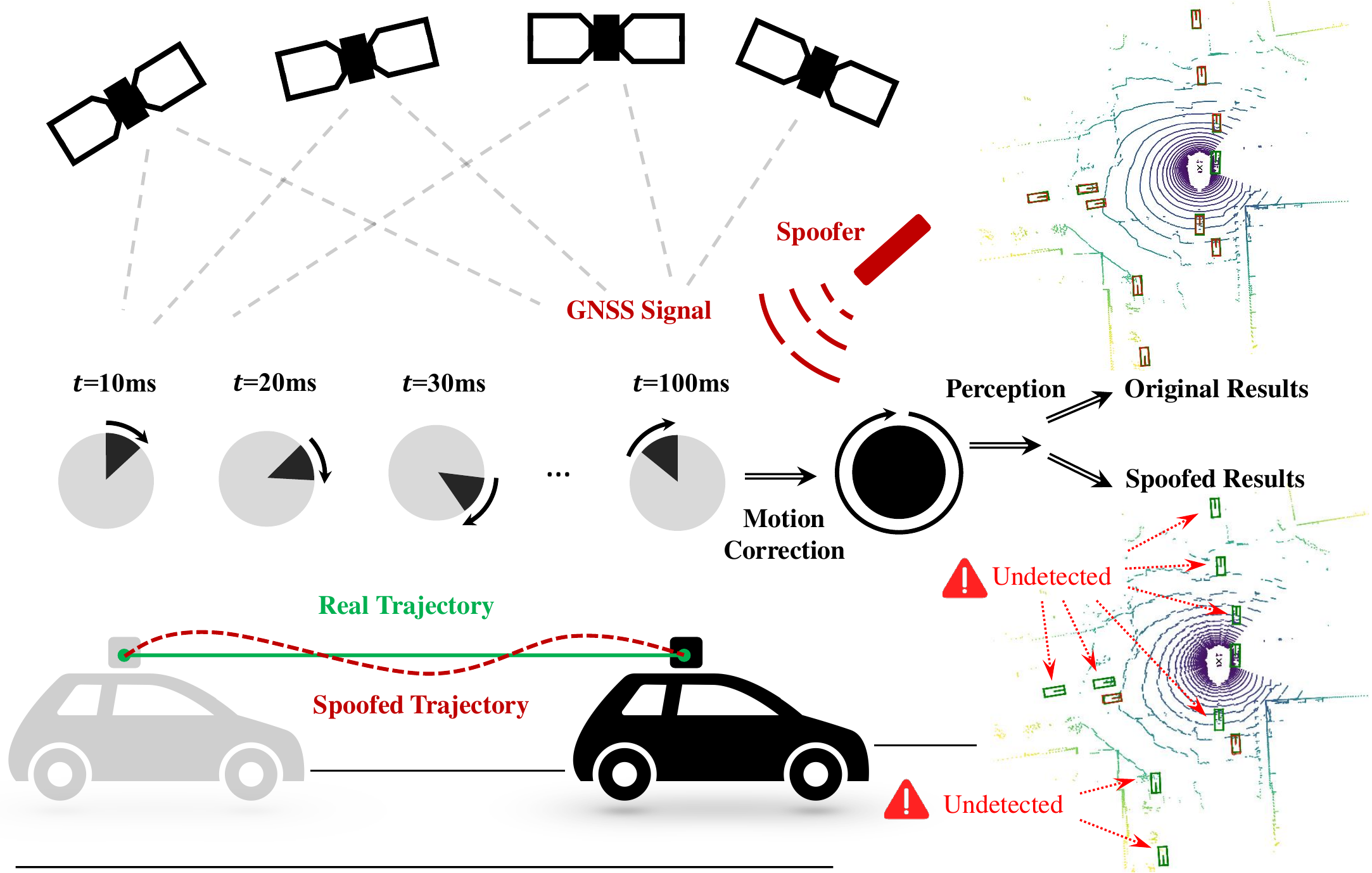}
	\caption{Illustration of our trajectory-based attack and the motion correction process. The top right and bottom right figures are respectively original and distorted LiDAR sweep as well as the detection results. \tbgreen{Green}/\tbred{red} boxes denote the ground truth/prediction. In this example, our method (named as FLAT) makes the detector miss eight out of eleven vehicles.}
	\label{fig:fig1}
	\vspace{-3mm}
\end{figure}

Autonomous driving systems are generally equipped with all kinds of sensors to perceive the complex environment~\cite{Geiger2012AreWR}. Among the sensors, LiDAR has played a crucial role due to its plentiful geometric information sampled by incessantly spinning a set of laser emitters and receivers. However, LiDAR scans are easily distorted by vehicle's motion, \ie, the points in a full sweep are sampled at different timestamps when vehicle is at different locations and orientations, as shown in Fig.~\ref{fig:fig1}. Imagine that a self-driving car is on a highway at a speed of 30m/s, its LiDAR with a 20Hz scanning frequency would move 1.5 meters during a full sweep, severely distorting the captured point cloud.

Such distortions are typically compensated by querying the vehicle/LiDAR's pose at any time from a continuous vehicle ego-motion tracking module that fuses pose estimation from Global Navigation Satellite System (GNSS, \eg, GPS, GLONASS, and BeiDou), Inertial Navigation System (INS), and SLAM-based localization using LiDAR or cameras. Well-known LiDAR datasets like KITTI~\cite{Geiger2012AreWR} and nuScenes~\cite{Caesar2020nuScenesAM} have already corrected such motion distortions prior to the dataset release. Researchers then made impressive progresses by processing those distortion-free point clouds using deep neural networks (DNNs) for many tasks, \eg, 3D object detection, semantic/instance segmentation, motion prediction, multiple object tracking. 

However, using DNNs on LiDAR point clouds creates a potentially dangerous and less cognizant vulnerability in self-driving systems. First, the above perception tasks are functions of LiDAR point clouds implemented via DNNs. Second, the motion compensation makes LiDAR point clouds a function of the vehicle trajectories. This functional composition leads to a simple but surprising fact that \textit{those perception tasks are now also functions of the trajectories}. Thus, such a connection exposes the well-known adversarial robustness issue of DNNs to hackers who could now fool a self-driving car's safety-critical LiDAR-depending perception modules by calculatedly spoofing the area's wireless GNSS signals, which is still a serious and unresolved security problem demonstrated on many practical systems~\cite{Shen2020DriftWD,Narain2019SecurityOG,Zeng2018AllYG}. Luckily, given that the aforementioned non-GNSS pose sources such as INS and visual SLAM are fused together for vehicle ego-motion estimation, large variations (meter-level) of GNSS trajectory spoofing could be detected and filtered, ensuring a safe localization and mapping for self-driving cars. However, \textit{what if a spoofed trajectory only has dozens of centimeters offset? Would current point cloud DNNs be robust enough under such small variations?}

In this paper, we initiate the first work to reveal and investigate such a vulnerability. Different from existing works directly attacking point cloud coordinates with 3D point perturbations or adversarial object generation~\cite{Wicker2019RobustnessO3,Tu2020PhysicallyRA,Xiang2019Generating3A,Hamdi2020AdvPCTA}, we propose to fool LiDAR-based perception modules by attacking the vehicle trajectory, which could be detrimental and easily deployable in the physical world. Our investigation includes how to obtain LiDAR sweeps with simulated motion distortion from real-world datasets, convert LiDAR point clouds as a differentiable function of the
vehicle trajectories, and eventually calculate the adversarial trajectory perturbation and make them less perceptible. Our principal contributions are as follows:
\begin{itemize}[nosep,nolistsep]
	\item We propose an effective approach for simulating motion distortion using a sequence of real-world LiDAR sweeps from existing dataset.
	\item We propose a novel view of LiDAR point clouds as a differentiable function of the vehicle trajectories, based on the real-world motion compensation process.
	\item We propose to \textbf{F}ool \textbf{L}iDAR perception with \textbf{A}dversarial \textbf{T}rajectory (\textbf{FLAT}), which has better feasibility and transferability (code will be released).
	\item We conduct extensive experiments on 3D object detection as a downstream task example, and show that the advanced detectors can be effectively blinded.
\end{itemize}

\section{Related Work}\label{sec:related}

\noindent\textbf{GNSS/INS and LiDAR Motion Compensation.} 
Motion distortion is also known as motion blur or rolling shutter effect of LiDAR on ego-motion vehicles~\cite{manivasagam2020lidarsim}. To compensate for such distortions, GNSS/INS is often used to provide the pose of the LiDAR at any moment when a point is scanned. This opens backdoors for a self-driving system. First, space weather is a substantial error source for GNSS and also significantly affects systems such as differential GPS. The influence of ionosphere disturbances on GPS kinematic precise point positioning (PPP) can be larger than 2 to 10 meters at different latitudes~\cite{yang2020global}, while solar radio burst could cause GPS positioning errors as large as 300 meters vertically and 50 meters horizontally~\cite{muhammad2015performance}.
Besides those naturally occurring events, malicious attacks such as GPS jamming or spoofing could also be used to arbitrarily modify the GPS trajectory~\cite{Shen2020DriftWD,Narain2019SecurityOG,Zeng2018AllYG}. Moreover, when GNSS and INS are fused, such attacks can affect not only positions but also the rotational component (gyroscope bias compensation) of a vehicle's trajectory~\cite{xu2018performance}.
Of course, nowadays for motion compensation, LiDAR poses are usually fused between GNSS/INS and LiDAR-/camera-based localization, so large pose variations from extreme space weather or ``urban canyon'' effect could be filtered. \textit{But as long as GNSS is a part of the equation, the backdoors could remain open, especially when the spoofed trajectories only have small variations from the ground truth}, as we show later.

%--------------------------------------------------------
% \subsection{}
\noindent\textbf{Image-based Adversarial Attack.} Despite the great success achieved by deep learning in both academia and industry, researchers have found that deep networks are susceptible to carefully-designed adversarial perturbation which is hard to distinguish. Since such vulnerability is firstly pointed out in image classification task~\cite{Szegedy2014IntriguingPO}, broad attentions have been paid to adversarial robustness in various downstream tasks, \eg, semantic segmentation~\cite{Xie2017AdversarialEF}, object detection~\cite{Pang2018TowardsRD}, visual tracking~\cite{Guo2019SPARKSO}, \etc. Adversarial attack is divided into white box~\cite{MoosaviDezfooli2016DeepFoolAS} and black box~\cite{Brendel2018DecisionBasedAA} based on whether the model parameters are known. Besides, attacks can be categorized as targeted~\cite{Yan2020CoolingShrinkingAB} and untargeted~\cite{Wu2019UntargetedAA} according to whether the adversary has a particular goal. Meanwhile, many attempts have been made in defense mechanisms, such as adversarial training~\cite{Madry2018TowardsDL}, certified defense~\cite{Raghunathan2018CertifiedDA,Yu2019AND}, adversarial example detector~\cite{Yu2019AND}, and ensemble diversity~\cite{Pang2019ImprovingAR}. Extensive studies in image-based adversarial attack and defense have largely promoted the development of trustworthy machine learning in 2D computer vision and inspired similar investigations in 3D vision.

\noindent\textbf{Point Cloud Attack and Defense.} 
% LiDAR has been widely applied in autonomous driving because of its rich geometric information. Cao2019AdversarialSA,Sun2020TowardsRL,Shin2017IllusionAD  Cao2019AdversarialOA
Recently, researchers have explored the vulnerability of DNNs taking point clouds of 3D objects as input. For the object-level point cloud attack, \citet{Xiang2019Generating3A} proposed point perturbation as well as cluster generation to attack the widely-used PointNet~\cite{Qi2017PointNetDL}. Besides, critical points removal~\cite{Wicker2019RobustnessO3, Yang2019AdversarialAA}, adversarial deformation~\cite{Zhou2020LGGANLG}, geometric-level attack~\cite{Lang2020GeometricAA} are proposed for fooling the point cloud-based deep model. However, none of them directly target LiDAR point clouds of self-driving scenes which have domain gaps than object-level point clouds. Moreover, to implement the above point cloud attacks towards a self-driving car requires tampering with its software for altering point cloud coordinates. Differently, our paper reveals \textit{a simple yet dangerous possibility} of spoofing the trajectory to attack deep modules through the LiDAR motion correction process yet \textit{without any need of software hacking}. As for the works on scene-level point cloud attack, \citet{Tu2020PhysicallyRA} proposed to generate 3D adversarial shapes placed on the rooftop of a target vehicle, making the target invisible for the detectors. Some works~\cite{Cao2019AdversarialSA,Sun2020TowardsRL,Shin2017IllusionAD,Cao2019AdversarialOA} created spoofing obstacles/faked points in front of the car to influence the vehicle's decision, but 
they can only modify the limited area of the scene. Differently, our method does not need to physically alter any shapes in the scene, and can \textit{easily scale up the attack to the whole scene}. In a word, existing works directly manipulate on the point coordinates either physically or virtually, while our attack is realized by spoofing the vehicle trajectories.

\noindent\textbf{Affected Downstream Tasks.}
Theoretically, every downstream task that require LiDAR point cloud as input could be affected by the attacks proposed in this paper. This could include geometric vision tasks such as registration, pose estimation, and mapping, as well as pattern recognition tasks such as 3D object detection~\cite{Lang2019PointPillarsFE,Yang20203DSSDP3,Shi2019PointRCNN3O, Zhou2018VoxelNetEL,Liang2019MultiTaskMF,Wang2020PillarbasedOD}, semantic segmentation~\cite{Jiang2020PointGroupDP,Hu2020RandLANetES}, motion prediction~\cite{Wu2020MotionNetJP, Luo2018FastAF, Zeng2019EndToEndIN}, and multiple object tracking~\cite{Weng20193DMT,Weng2020GNN3DMOTGN}. While the first group of tasks could be less severely affected by GNSS spoofing via data fusion as mentioned above, the second group has higher vulnerability, because small but calculated perturbations in the point coordinates could affect deep networks as demonstrated in the above related works. In this paper, without loss of generality, we choose to focus on the 3D object detection task to illustrate the severity of this backdoor, because miss detection of safety-critical objects surrounding a self-driving car could be a matter of life or death.

%--------------------------------------------------------
\section{Motion Distortion in LiDAR}\label{sec:motion_distortion}
%By emitting laser pulse and measuring the distance to each object based on the reflection, LiDAR can learn abundant geometric knowledge about its surrounding to ensure the safety of self-driving cars. This situation is exacerbated when the car's speed is much faster than the LiDAR scanning speed.
LiDAR measurements are obtained along with the rotation of its beams, so the measurements in a full sweep are captured at different timestamps, introducing motion distortion which jeopardizes the vehicle perception. Autonomous systems generally utilize LiDAR's location and orientation obtained from the localization system to correct distortion~\cite{Shan2020ProbabilisticEM,Frossard2020StrObeSO,Han2020StreamingOD}. Most LiDAR-based datasets~\cite{Caesar2020nuScenesAM,Geiger2012AreWR} have finished synchronization before release. Hence, the performance of current 3D perception algorithms in the distorted point cloud remains unexplored. We briefly introduce the nomenclature in this work before  detailed illustrations.

\textbf{World Frame.} We use a coordinate frame $W$ fixed in the world  with the orthonormal basis  $\{\mathbf{x}_W , \mathbf{y}_W , \mathbf{z}_W\}$ to describe the global displacement of the self-driving car.

\textbf{Object Frame.} The car can be associated with a right-handed, orthonormal coordinate frame which can describe its rigid body motion. Such a frame attached to the car is called object frame~\cite{ma2012invitation}. In the following, we use \textit{frame} to denote the object frame of the car at different timestamps.

\textbf{Sweep\&Packet.} LiDAR points in a complete 360$^\circ$ is called a \textit{sweep}, and the points are emitted as a stream of \textit{packets}, each covering a sector of the 360$^\circ$ coverage~\cite{Han2020StreamingOD}.
%--------------------------------------------------------
\subsection{Linear Pose Interpolation}
\begin{figure}[t]
	\includegraphics[width=0.48\textwidth]{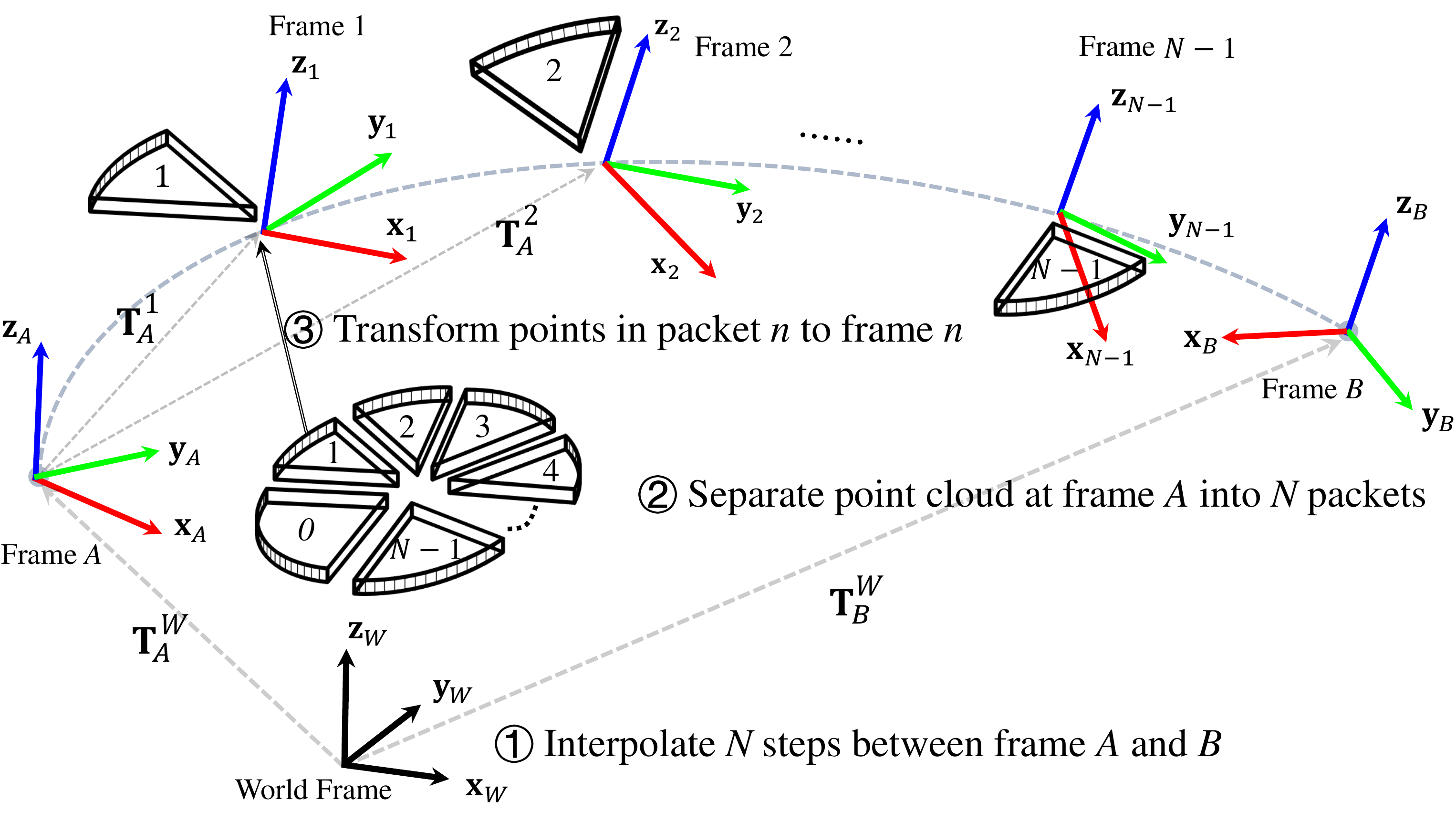}
	\caption{Diagram of motion distortion simulation. Firstly, we interpolate $N$ steps in 6DoF pose between two adjacent frames $A$ and $B$. Secondly, we divide a sweep at frame $A$ into $N$ packets, and the $n$-th sector corresponds to the $n$-th interpolated frame. Thirdly, we transform the point cloud in the $n$-th packet at frame $A$ into frame $n$ with homogeneous transformation ${\mathbf{T}^n_{A}}$. Finally, motion-distorted point cloud can be generated by aggregating the point cloud from frame $0$ to frame $N-1$.}
	\label{fig:motion_distortion}
\end{figure}
In this section, we recover the raw point cloud before synchronization with a sequence of real-world LiDAR point clouds from nuScenes~\cite{Caesar2020nuScenesAM}. Each sample of nuScenes provides a sweep and its corresponding ego pose. We assume the vehicle is moving smoothly and carry out linear pose interpolation between two adjacent frames $A$ and $B$, as shown in Fig.~\ref{fig:motion_distortion}. $\{\mathbf{x}_A , \mathbf{y}_A, \mathbf{z}_A\}$ and $\{\mathbf{x}_B , \mathbf{y}_B, \mathbf{z}_B\}$ are their orthonormal basis. Using $\mathbf{t}_A\in \mathbb{R}^{3}$ and $\mathbf{t}_B\in \mathbb{R}^{3}$ to represent global translation of frame $A$ and $B$, the global translation of the $n$-th ($n = 0, 1, 2, ..., N-1$) interpolated frame is:
\begin{equation}\label{eq:translation}
\mathbf{t}(n) = \mathbf{t}_A + \frac{\mathbf{t}_B-\mathbf{t}_A}{N}\times n \ ,
\end{equation}
where $N$ is the total interpolation steps. For the orientation, we implement spherical linear interpolation (slerp)~\cite{Shoemake1985AnimatingRW}. Using $\mathbf{q}_A=\left[q^A_{w}, q^A_{x}, q^A_{y}, q^A_{z}\right]^{\top}$ and $\mathbf{q}_B=\left[q^B_{w}, q^B_{x}, q^B_{y}, q^B_{z}\right]^{\top}$ to represent quaternion at frame $A$ and $B$, then the quaternion at the $n$-th interpolated frame is: 
\begin{equation}\label{eq:quaternion}
\mathbf{q}(n)
= \frac{\sin ((1-n) \theta)}{\sin \theta} \mathbf{q}_A+\frac{\sin (n \theta)}{\sin \theta} \mathbf{q}_B \ ,
\end{equation}
where $\theta=\cos ^{-1}\left(\mathbf{q}_A \cdot \mathbf{q}_B\right)$ is the rotation angle between $A$ and $B$. Convert the quaternion $\mathbf{q}(n)=\left[q^n_{w}, q^n_{x}, q^n_{y}, q^n_{z}\right]^{\top}$ to the rotation matrix $\mathbf{R}(n) \in SO(3)$, then

the homogeneous transformation matrix from the world frame $W$ to the $n$-th interpolated frame denoted as $\mathbf{T}^W_{n} \in SE(3)$ is:
\begin{equation}\label{eq:transformation}
\mathbf{T}^W_{n}=\left[\begin{array}{cc}
\mathbf{R}(n) & \mathbf{t}(n) \\
\mathbf{0}_{1 \times 3} & 1
\end{array}\right] \ .
\end{equation}

%--------------------------------------------------------
\subsection{Motion Distortion Simulation}
After linear pose interpolation, we can get $N$ more frames. To simulate motion distortion, we assume that the sweep at frame $A$ is aggregated from $N$ frames, and in each frame, the beam scans a degree of $\frac{360}{N}$. Therefore, we firstly divide the sweep at frame $A$ into $N$ packets, as shown in Fig.~\ref{fig:motion_distortion}. Then we transform the LiDAR packet $n$ to the coordinate of frame $n$ at which the point cloud is assumed to be captured. The points in packet $n$ at frame $A$ (denoted as $\prescript{A}{}{\mathbf{P}_n} \in \mathbb{R}^{4\times m_n}$) is represented as a set of 3D points $\{P_m| m = 1,2,..., m_n\}$, where each point $P_m$ is a vector of its homogeneous coordinate $(x, y, z, 1)$, and $m_n$ denotes the number of points in packet $n$. We transform $\prescript{A}{}{\mathbf{P}_n}$ to its corresponding capture frame $n$:
\begin{equation}
\prescript{n}{}{\mathbf{P}} = {\mathbf{T}^n_A} \prescript{A}{}{\mathbf{P}_{n}} \ ,
\end{equation}
where $\prescript{n}{}{\mathbf{P}} \in \mathbb{R}^{4\times m_n}$ is the point cloud of packet $n$ in the coordinate of frame $n$ (can be treated as the point cloud captured at frame $n$) and ${\mathbf{T}^n_A}$ is the homogeneous transformation from frame $n$ to frame $A$ and is as follows:
\begin{equation}\label{eq:A_T_n}
{\mathbf{T}^n_A} = {\mathbf{T}^n_W} {\mathbf{T}^W_A} = ({\mathbf{T}^W_n})^{-1}{\mathbf{T}^W_A} \ ,
\end{equation}
where ${\mathbf{T}^W_n}$ is from Eq.~\eqref{eq:transformation} and ${\mathbf{T}^W_A}$ can also be calculated given $\mathbf{t}_A$ and $\mathbf{q}_A$. Finally, a set of motion-distorted point clouds $^A \widetilde{\mathbf{P}} \in \mathbb{R}^{4\times m} $ is generated by aggregating multiple LiDAR packets captured at different frames:
\begin{equation}\label{eq:6}
^A\widetilde{\mathbf{P}} = [\prescript{0}{}{\mathbf{P}}; \prescript{1}{}{\mathbf{P}}; ...;  \prescript{N-1}{}{\mathbf{P}}] \ ,
\end{equation}
where $[\cdot;\cdot;...;\cdot]$ is the concatenation operation along the row. It is noted that $m = \sum_{n=0}^{N-1} m_n$.
%--------------------------------------------------------
\subsection{Motion Compensation with Ego-Pose}
Up to now, we have generated motion-distorted point cloud as shown in Eq.~\ref{eq:6}, and then we can represent the motion compensation using the mathematical formula, \ie, a LiDAR sweep can be written as a differentiable function of the vehicle trajectory\footnote{In this work, the translation and the orientation are collectively referred to as ``trajectory''. }. The undistorted point cloud ${^A\mathbf{P}}  \in \mathbb{R}^{4\times m}$ is obtained by transforming point clouds from frame $0 \thicksim N-1$ back to the coordinate of frame $A$:
\begin{equation}\label{eq:pc_generating}
{\prescript{A}{}{\mathbf{P}}} = [{\mathbf{T}^A_0}\prescript{0}{}{\mathbf{P}}; {\mathbf{T}^A_1}\prescript{1}{}{\mathbf{P}}; ...; {\mathbf{T}^A_{N-1}}\prescript{N-1}{}{\mathbf{P}}] \ ,
\end{equation}
where ${\mathbf{T}^A_n}$ is the transformation from frame $A$ to frame $n$, which is the inverse matrix of ${\mathbf{T}^n_A}$ in Eq.~\eqref{eq:A_T_n}.

%--------------------------------------------------------
\section{Adversarial Trajectory Perturbation}
% In this section, we first introduce the point cloud generation function with respect to SDV's trajectory. Afterwards, objectives of adversarial attacks with and without regularizations are clarified.
%--------------------------------------------------------
\subsection{Point Cloud Representation w.r.t. Trajectory}
The studies of object-level point cloud generally treat point cloud as a set of 3D points sampled from mesh models instantaneously~\cite{Xiang2019Generating3A, Hamdi2020AdvPCTA}. In autonomous driving, however, the 3D points are captured in a dynamic setting through raycasting, thus LiDAR not only records points' $(x, y, z)$ coordinates, but also the timestamps at which the points are captured. To this end, we propose a novel representation of point cloud as a function of vehicle trajectory through Eq.~\eqref{eq:pc_generating} which can be written as a general form:
\begin{equation}\label{eq:trajectory}
{\mathbf{P}} = {f}(\mathbf{T}, L(\prescript{n}{}{\mathbf{P}})) ,\ \ n=0,1,...,N-1 \ ,
\end{equation}
where ${\mathbf{P}} \in \mathbb{R}^{4\times m} $ is a full sweep with $m$ points, and is represented as a differentiable function $f$ in terms of vehicle trajectory $\mathbf{T} \in \mathbb{R}^{N \times 4\times 4}$ and the list of LiDAR packets $L(\prescript{n}{}{\mathbf{P}}) = [\prescript{0}{}{\mathbf{P}}, \prescript{1}{}{\mathbf{P}},..., \prescript{N-1}{}{\mathbf{P}}]$ ($\prescript{n}{}{\mathbf{P}} \in \mathbb{R}^{4\times m_n}$). Noted that the trajectory between two adjacent frames is represented as a set of homogeneous transformation matrices. Different from previous works tampering with point coordinates~\cite{Xiang2019Generating3A}\footnote{In the following, we will use \textit{coordinate attack} to denote the modification of point coordinate.}, we propose to represent point cloud as a function of trajectory, and attack the trajectory instead of 3D points. Our method has the following advantages.

\textbf{Physical Feasibility.} Since the motion compensation is naturally occurring in self-driving, it is physically-realizable and straightforward to attack the trajectory, \eg, by wireless GNSS spoofing. In contrast, coordinate attack requires software hacking which is infeasible in practice.

\textbf{Better Transferability.}  Our learned trajectory perturbation with the same size ($N \times 4 \times 4$) can be easily transferred to different sweeps. Yet, coordinate attacks cannot be transferred across sweeps because different sweeps could have different numbers of points, and the perturbation in the point space could have different dimensions.

\textbf{Novel Parameterization.} We attack the 6-DoF pose of each packet, but coordinate attack modifies a single point’s xyz position without orientations. Hence, our method has better performance due to new attacked parameters. 

%--------------------------------------------------------
%--------------------------------------------------------
\subsection{Objective Function}
Since the point cloud is represented as a differentiable function of the trajectory, the gradient can be back-propagated to the trajectory smoothly for adversarial learning. The adversarial objective is to minimize the negative loss function of the deep model with parameter $\theta$:
\begin{equation}
\min \widetilde{\mathcal{L}}(\theta, {f}(\mathbf{T}, L(\prescript{n}{}{\mathbf{P}})), \mathbf{y}) \ ,
\end{equation}
where $\mathbf{y}$ is the desired output of the network, ${f}(\mathbf{T}, L(\prescript{n}{}{\mathbf{P}}))$ is the input point cloud as a function of the trajectory.  Assume that we are assigned to attack a downstream task with loss function ${\mathcal{L}}$, \eg, cross-entropy loss in classification, then $\widetilde{\mathcal{L}} = -{\mathcal{L}}$. Afterwards, the adversarial perturbation $\bm{\delta} \in \mathbb{R}^{N \times 4\times 4}$ can be obtained via projected gradient descent (PGD) with multiple iterations~\cite{Madry2018TowardsDL}:
\begin{equation}\small
\bm{\delta}^{t+1}=\mathcal{P}\left(\bm{\delta}^{t}-\alpha \operatorname{sgn}\left(\nabla_{\bm{\delta}} \widetilde{\mathcal{L}}(\theta, {f}(\mathbf{T}+\bm{\delta}, L(\prescript{n}{}{\mathbf{P}})), \mathbf{y})\right)\right) \ ,
\end{equation}
where $t$ denotes iteration number, $\mathcal{P}$ indicates the projection onto the convex set of interest, and this work uses clipping which is the case of $\ell_{\infty}$ norm to ensure an acceptable perturbation magnitude. The $n$-th element in $\bm{\delta}$ in the matrix form is:
\begin{equation}\label{eq:perturbation}
\bm{\delta}(n)=\left[\begin{array}{cc}
\widetilde{\mathbf{R}}(n) & \widetilde{\mathbf{t}}(n) \\
\mathbf{0}_{1 \times 3} & 1
\end{array}\right] \ .
\end{equation}

\noindent \textbf{Polynomial Trajectory Perturbation.} To make the attack highly imperceptible, the polynomial trajectory perturbation is proposed: we define the perturbation in translations as a third-order polynomial of time as follows:
\begin{equation}\small
\widetilde{\mathbf{t}}(n) = {\bm{\beta}}^T  \mathbf{n} \ , 
\end{equation}
where $\mathbf{n}= [1, n, n^2, n^3]^{\top} $  and $\bm{\beta} = [\bm{\beta_x}, \bm{\beta_y}, \bm{\beta_z}]$ is the polynomial coefficients. Now $\bm{\delta}$ is a differentiable function of $\bm{\beta}$, and the gradient will be back-propagated to $\bm{\beta}$, so the adversarial coefficients are calculated by: 
\begin{equation}\small
\bm{\beta}^{t+1}=\mathcal{P}\left(\bm{\beta}^{t}-\alpha \operatorname{sgn}\left(\nabla_{\bm{\beta}} \widetilde{\mathcal{L}}(\theta, {f}(\mathbf{T}+\bm{\delta}, L(\prescript{n}{}{\mathbf{P}})), \mathbf{y})\right)\right) \ ,
\end{equation}
hence, we only need to \textit{manipulate several key points to bend a polynomial-parameterized trajectory} which can be easily achieved in reality, achieving a real-time attack.

%--------------------------------------------------------
%--------------------------------------------------------
\subsection{Attack with Regularization}
\textbf{Regularization on Trajectory.} To realize an imperceptible attack, we propose a trajectory smoothness regularizer to repress the total variation in vehicle poses. Given a trajectory perturbation $\bm{\delta} \in \mathbb{R}^{N \times 4\times 4}$, we calculate the difference in translation and rotation separately. The smoothness $\mathcal{S}(\bm{\delta})$ equals to the following formula:
% \begin{equation}\small
% 	\mathcal{S}(\bm{\delta})=\sum^{N-1}_{n=1}(\lambda_{\mathbf{t}}\left(\left(\widetilde{\mathbf{t}}_{n}-\widetilde{\mathbf{t}}_{n-1}\right)^{p}\right)^{\frac{1}{p}} + \lambda_{\mathbf{R}}\left(\left(\widetilde{\mathbf{R}}_{n}-\widetilde{\mathbf{R}}_{n-1}\right)^{p}\right)^{\frac{1}{p}}) \ ,
% \end{equation}
\begin{equation}\small
\lambda_{\mathbf{t}}(\sum^{N-1}_{n=1}(\widetilde{\mathbf{t}}(n)-\widetilde{\mathbf{t}}(n-1))^{p})^{\frac{1}{p}}
+\lambda_{\mathbf{R}}(\sum^{N-1}_{n=1}(\widetilde{\mathbf{R}}(n)-\widetilde{\mathbf{R}}(n-1))^{p})^{\frac{1}{p}} \ ,
\end{equation}
where $\lambda_{\mathbf{t}}$ and $\lambda_{\mathbf{R}}$ are used to balance the influence of translations and rotations, $p$ denotes the norm ($p=2$ in this work). With trajectory smoothness regularization, the objective of adversarial attacks is as follows:
\begin{equation}\label{eq:smoothness}
\min \widetilde{\mathcal{L}}(\theta, {f}(\mathbf{T}+\bm{\delta}, L(\prescript{n}{}{\mathbf{P}})), \mathbf{y}) + \lambda_s \mathcal{S}(\bm{\delta}) \ ,
\end{equation}
where $\lambda_s$ is for adjusting the smoothness degree. By optimizing over Eq.~\eqref{eq:smoothness}, we aim to find an imperceptible adversarial perturbation $\bm{\delta}$ with desirable smoothness.
%--------------------------------------------------------
%--------------------------------------------------------

\textbf{Regularization on Point Cloud.} We also propose a regularizer on point cloud to repress its change. We use two metrics to measure the variation of point cloud before and after distortion, \ie, $\ell_{p}$ norm and Chamfer distance~\cite{Fan2017APS}. Using $\mathcal{D}_L(\bm{\delta})$ to denote $\ell_{p}$ distance, $\mathcal{D}_C(\bm{\delta})$ to denote Chamfer distance before and after perturbation $\bm{\delta}$. The adversarial objective with regularization is defined as:
\begin{align}\label{eq:distortion}
\min \widetilde{\mathcal{L}}(\theta, {f}(\mathbf{T}+\bm{\delta}, L(\prescript{n}{}{\mathbf{P}})), \mathbf{y}) + \lambda_d \mathcal{D}(\bm{\delta}) \ ,
\end{align}
where $\mathcal{D}(\bm{\delta})$ can be either $\mathcal{D}_L(\bm{\delta})$ or $\mathcal{D}_C(\bm{\delta})$. $\lambda_d$ is to control the degree of distortion. By optimizing over Eq.~\eqref{eq:distortion}, we try to search for a powerful adversarial perturbation $\bm{\delta}$ leading to subtle distortion in the point cloud.
% \textbf{$\ell_{p}$ norm.} For data with fixed shape, $\ell_{p}$ norm is generally adopted to measure adversarial perturbation. The $\ell_{p}$ norm of the perturbation $\bm{\delta}$ is defined as:
% \begin{equation}
% \mathcal{D}_L\left(\bm{\delta}\right)=\left(\sum_{i}\left(x_{i}-\widetilde{x}_{i}\right)^{p}\right)^{\frac{1}{p}} \ ,
% \end{equation}
% where $x_{i}$ is the $i$-th point coordinate in the undistorted point cloud ${\mathbf{P}}={f}(\mathbf{T}, L(\mathbf{P}^n))$, and $\widetilde{x}_{i}$ is its
% corresponding point in distorted point cloud $\widetilde{\mathbf{P}}={f}(\mathbf{T}+\bm{\delta}, L(\mathbf{P}^n))$.

% \textbf{Chamfer Distance.}
% Chamfer distance~\cite{Fan2017APS} calculate the average distance of the nearest point
% pairs in original point cloud ${\mathbf{P}}$ and distorted point cloud $\widetilde{\mathbf{P}}$:
% \begin{align}
% \mathcal{D}_{C}\left(\bm{\delta}\right)&=\frac{1}{\|\widetilde{\mathbf{P}}\|_{0}} \sum_{\widetilde{x} \in \widetilde{\mathbf{P}}} \min _{x \in {\mathbf{P}}}\|x-\widetilde{x}\|_{2}^{2} \ \\
% &+ \frac{1}{\|{\mathbf{P}}\|_{0}} \sum_{x \in {\mathbf{P}}} \min _{\widetilde{x} \in \widetilde{\mathbf{P}}}\|x-\widetilde{x}\|_{2}^{2} \ .
% \end{align}

\begin{table*}[t]
% \small
\footnotesize
\setlength{\tabcolsep}{2.5mm}
\begin{center}
\caption{\textbf{Quantitative results of white box attack:} AP (IoU=0.7) of 3D bounding boxes on nuScenes~\cite{Caesar2020nuScenesAM}. We report results of the car category under different levels of difficulty and ranges of depth following~\citep{wang2020train}. In four attack settings of our method (\textbf{FLAT}), the best and second best attack qualities among four attacking targets are respectively highlighted using \tbred{red} and \tbblue{blue} color. In attacking translations/rotations, the step size for each iteration is 0.1 and 0.01 respectively, and the number of attack iteration is 20 for both two settings.
}
\vspace{0.1cm}
\label{tab:flat}
\begin{spacing}{1}
\begin{tabular}{c|c|c|ccc|ccc}
\toprule
\multicolumn{3}{c|}{Attack Approach \quad  $\backslash$ \quad  Case}  & Easy & Moderate & Hard & 0-30m & 30-50m & 50-70m \\ \midrule
\multicolumn{3}{c|}{PointRCNN~\cite{Shi2019PointRCNN3O}}  & 47.44 & 21.56 & 20.91 & 47.44 & 2.16 & 0.17 \\ \midrule
\multirow{4}{*}{Baseline} 
&\multicolumn{2}{l|}{Coordinate Attack~\cite{Xiang2019Generating3A}}  & 16.42 & 6.58 & 5.90 & 15.20 & 0.48 & 0.03 \\ 
&\multicolumn{2}{l|}{Random Attack (Point Cloud)}  & 30.09 & 12.39 & 10.84 & 25.65 & 0.79 & 0.06 \\ 
&\multicolumn{2}{l|}{Random Attack (Translation)}  & {17.00} & 8.58 & 8.90 & 20.43 & 1.09 & 0.09 \\ 
&\multicolumn{2}{l|}{Random Attack (Rotation)}  & 12.30 & 4.87 & 5.13 & 13.31 & {0.01} & {0.00} \\ 
&\multicolumn{2}{l|}{Random Attack (Full Trajectory)}  & 5.66 & 2.43 & 2.78 & 7.67 & 0.02 & {0.00} \\ \midrule
\multirow{4}{*}{\textbf{FLAT (Translation)}} &\multirow{2}{*}{Classification} & \multirow{1}{*}{Stage-1}  & \tbblue{12.94} & \tbblue{6.58} & \tbblue{7.22} & \tbblue{16.82} & \tbred{0.86} & {0.06}  \\ 
  && \multirow{1}{*}{Stage-2} & \tbred{11.72} & \tbred{5.87} & \tbred{6.06} & \tbred{13.91} & \tbblue{0.87} & \tbblue{0.04} \\ 
 & \multirow{2}{*}{Regression} & \multirow{1}{*}{Stage-1}  & 17.46 & 8.24 & 8.57 & 19.36 & 1.09 & \tbred{0.03} \\ 
 & & \multirow{1}{*}{Stage-2}  & 26.09 & 12.78 & 12.53 & 27.15 & 2.17 & 0.17  \\ \midrule
 \multirow{4}{*}{\textbf{FLAT (Polynomial)}} &\multirow{2}{*}{Classification} & \multirow{1}{*}{Stage-1}  & \tbblue{17.94} & \tbblue{9.36} & \tbblue{9.56} & \tbblue{20.73} & 1.90 & 0.19 \\ 
  && \multirow{1}{*}{Stage-2} & \tbred{12.51} & \tbred{6.37} & \tbred{6.54} & \tbred{14.51} & \tbred{1.38} & \tbblue{0.16} \\ 
 & \multirow{2}{*}{Regression} & \multirow{1}{*}{Stage-1}  &  22.60 & 11.01 & 10.96 & 24.36 & \tbblue{1.67} & \tbred{0.15} {} \\ 
 & & \multirow{1}{*}{Stage-2}  & 26.04 & 12.76 & 12.51 & 27.19 & 2.17 & 0.17  \\ \midrule
 \multirow{4}{*}{\textbf{FLAT (Rotation)}} &\multirow{2}{*}{Classification} & \multirow{1}{*}{Stage-1}  &6.32 & 2.43 & 2.51 & 7.32 & \tbblue{0.02} & \tbred{0.00}  \\ 
  && \multirow{1}{*}{Stage-2} & \tbred{2.35} & \tbred{0.80} & \tbred{0.61} & \tbred{2.03} & \tbred{0.01} & \tbred{0.00} \\ 
 & \multirow{2}{*}{Regression} & \multirow{1}{*}{Stage-1}  & \tbblue{5.50} & \tbblue{1.87} & \tbblue{1.76} & \tbblue{5.45} & \tbblue{0.02} & \tbred{0.00} \\ 
 & & \multirow{1}{*}{Stage-2}  & 26.30 & 12.89 & 12.59 & 27.35 & 2.17 & \tbblue{0.17}  \\ \midrule
 \multirow{4}{*}{\textbf{FLAT (Full Trajectory)}} &\multirow{2}{*}{Classification} & \multirow{1}{*}{Stage-1}  & 1.52 & 0.45 & 0.51 & 1.71 & \tbblue{0.01} & \tbred{0.00}  \\ 
  && \multirow{1}{*}{Stage-2} & \tbred{0.19} & \tbred{0.01} & \tbred{0.02} & \tbred{0.26} & \tbred{0.00} & \tbred{0.00} \\ 
 & \multirow{2}{*}{Regression} & \multirow{1}{*}{Stage-1}  & \tbblue{1.01} & \tbblue{0.35} & \tbblue{0.32} & \tbblue{1.27} & \tbblue{0.01} & \tbred{0.00} \\ 
 & & \multirow{1}{*}{Stage-2}  & 26.03 & 12.70 & 12.48 & 27.13 & 2.16 & \tbblue{0.17}  \\\bottomrule
\end{tabular}
\end{spacing}
\end{center}
\vspace{-30pt}
\end{table*}
% \begin{table*}[t]
% \setlength{\tabcolsep}{3.5mm}
% \begin{center}
% \caption{AP (IoU=0.7) of 3D bounding boxes on the validation set of nuScenes~\cite{Caesar2020nuScenesAM} after only attacking \textbf{INS}. In our method (FLAT), the number of attack iteration is 20 and the step size for each iteration 0.01. Other settings are similar to Table~\ref{tab:trans-only}.
% }
% \vspace{0.1cm}
% \label{tab:rot-only}
% \begin{spacing}{1}
% \begin{tabular}{c|c|c|ccc|ccc}
% \toprule
% \multicolumn{3}{c|}{Attack Approach \quad  $\backslash$ \quad  Case}  & Easy & Moderate & Hard & 0-30m & 30-50m & 50-70m \\ \midrule
% \multicolumn{3}{c|}{Original Result of PointRCNN~\cite{Shi2019PointRCNN3O}}  & 47.44 & 21.56 & 20.91 & 47.44 & 2.16 & \tbblue{0.17} \\ \midrule
% \multirow{1}{*}{Baseline} &\multicolumn{2}{c|}{Random Attack}  & 12.30 & 4.87 & 5.13 & 13.31 & \tbred{0.01} & \tbred{0.00} \\ \midrule
% \multirow{4}{*}{\textbf{FLAT(Ours)}} &\multirow{2}{*}{Classification} & \multirow{1}{*}{Stage-1}  &6.32 & 2.43 & 2.51 & 7.32 & \tbblue{0.02} & \tbred{0.00}  \\ 
%   && \multirow{1}{*}{Stage-2} & \tbred{2.35} & \tbred{0.80} & \tbred{0.61} & \tbred{2.03} & \tbred{0.01} & \tbred{0.00} \\ 
%  & \multirow{2}{*}{Regression} & \multirow{1}{*}{Stage-1}  & \tbblue{5.50} & \tbblue{1.87} & \tbblue{1.76} & \tbblue{5.45} & \tbblue{0.02} & \tbred{0.00} \\ 
%  & & \multirow{1}{*}{Stage-2}  & 26.30 & 12.89 & 12.59 & 27.35 & 2.17 & \tbblue{0.17}  \\ \bottomrule
% \end{tabular}
% \end{spacing}
% \end{center}
% \vspace{-3em}
% \end{table*}
\section{Experiments}\label{sec:exp}
%--------------------------------------------------------
\subsection{Target 3D Deep Model}
In this work, we select widely-studied LiDAR-based 3D detection, which aims to estimate 3D bounding boxes of the objects in point cloud, as a downstream task example to verify our attack pipeline. Currently, LiDAR-based detection has two mainstreams: 1) point-based method directly consuming raw point cloud data, 2) voxel-based method which requires non-differentiable voxelization in the preprocessing stage. For the white box attack, we use point-based PointRCNN~\cite{Shi2019PointRCNN3O}. For the black box transferability test, we adopt voxel-based PointPillar++~\cite{Hu2020WhatYS}.
% \begin{itemize}[nosep,nolistsep]

% 	\item
\textbf{PointRCNN.} Our white box model, PointRCNN, uses PointNet++~\cite{Qi2017PointNetDH} as its backbone and includes two stages: stage-1 for proposal generation based on each foreground point, and stage-2 for proposal refinement in the canonical coordinate. Since PointRCNN uses raw point cloud as the input, the gradient can smoothly reach the point cloud, then arrive at vehicle trajectory. In this work, we individually attack the classification as well as regression branches in stage-1 and stage-2, with four attack targets in total.

% 	\item
\textbf{PointPillar++.} PointPillar~\cite{Lang2019PointPillarsFE} proposes a fast point cloud encoder using pseudo-image representation. It divides point cloud into bins and uses PointNet~\cite{Qi2017PointNetDL} to extract the feature for each pillar. Due to the non-differentiable preprocesssing stage, the gradient cannot reach the point cloud. Peiyun \etal~\cite{Hu2020WhatYS} proposed to augment PointPillar with the visibility map, achieving better precision. In this work, we use PointPillar++ to denote PointPillar with visibility map in~\cite{Hu2020WhatYS}. We use perturbation learned from the white box PointRCNN to attack black box PointPillar++, in order to examine the transferability of our attack pipeline.
% \end{itemize}
%--------------------------------------------------------
%--------------------------------------------------------
\subsection{Dataset and Evaluation Metrics}
\textbf{Dataset.} nuScenes~\cite{Caesar2020nuScenesAM} is a large-scale multimodel autonomous driving dataset captured by a real SDV with a full $360^{\circ}$ field of view in various challenging urban driving scenarios. Including 1000 scenes collected in Boston and Singapore in different weather, nuScenes has much more annotations (7 times) and images (100 times) than the pioneer KITTI~\cite{Geiger2012AreWR}. Besides, nuScenes provides a temporal sequence of samples in each scene, facilitating linear pose interpolation for motion distortion simulation, yet the 3D detection dataset in KITTI solely offers independent frames without temporal connection. 
Considering the above factors, nuScenes is employed in this work. 
%There are 28,130 samples for training and 6,019 samples for validation in nuScenes. 
We use PointRCNN model released in~\cite{wang2020train}, and the open-source PointPillar++ model in~\cite{Hu2020WhatYS}. Both models are trained on nuScenes training set. We report the results of white box on 1,000 samples from the validation set, and the results of black box on the whole validation set.

\textbf{Metrics.} 
For the white box PointRCNN, we report the 3D bounding boxes average precision (AP) with the IoU thresholds at 0.7 on car category. Following~\cite{wang2020train}, we evaluate the detector in three scenarios (Easy, Moderate, and Hard) based on the difficulty level of the surrounding cars. In addition, the performance within three depth ranges, \ie,
$0\sim 30$, $30 \sim 50$, and $50 \sim 70$ meters, are also assessed. For the black box PointPillar++, we follow the original paper~\cite{Hu2020WhatYS} to employ official 3D detection evaluation protocol in nuScenes~\cite{Caesar2020nuScenesAM}, \ie, average mAP over ten categories at four distance thresholds. For evaluating attack quality, we utilize the performance drop after attack.
%-------------------------------------------------------
\subsection{Experimental Setup}
 \textbf{Implementation details.} nuScenes samples keyframes at 2Hz from the original 20Hz data, so we assume that a sweep consumes 0.5 second\footnote{In fact, the LiDAR rotation period is 0.05 second, however, we only have access to annotations on keyframes at 2Hz, so we have to assume the capture frequency of LiDAR is also 2Hz.} and implement linear pose interpolation between two adjacent keyframes. We set total interpolation step $N$ as 100. For PGD, we restrict the perturbation to 10cm in translation, and to 0.01 in rotation. The step size for each attack iteration is 0.1/0.01 in translation/rotation, and the number of iterations is 20. More experimental settings like the step size and the number of iterations are reported in the supplementary.

\textbf{Baselines.} To demonstrate the superiority of our attack pipeline, we employ two baseline methods for comparison.  
\begin{itemize}[nosep,nolistsep]
	\item \textbf{Random Attack.} We add Gaussian noise with a 10cm standard deviation to the original point cloud, Gaussian noise with a 10cm standard deviation to the translations and Gaussian noise with 0.01 standard deviation to rotations.  
	\item \textbf{Coordinate Attack.} Coordinate attack~\cite{Xiang2019Generating3A} is directly manipulating the point set to deceive the detector. We attack classification of stage-2 and restrict the change in point coordinate to 10cm. The binary search step number is 10 and the number of iteration for each binary search step is 100.
\end{itemize}

\textbf{Attack Settings.} The polynomial trajectory attack is temporally-smooth and the others are temporally-discrete:
\begin{itemize}[nosep,nolistsep]
	\item \textbf{Attack Translation Only.} We merely modify the translation vector. The perturbation is a $100\times3$ matrix. 
	\item \textbf{Polynomial Perturbation.} We add a polynomial perturbation into the vehicle trajectory (translation part).
	\item \textbf{Attack Rotation Only.} We only interfere with the rotation matrix. The perturbation is a $100\times3\times3$ tensor.  
	\item \textbf{Attack Full Trajectory.} We tamper with the transformation matrix. The perturbation is a $100\times4\times4$ tensor.
\end{itemize}
\subsection{White Box Attack}\label{subsec:whitebox}
To explore the vulnerability of different stages and branches in PointRCNN, we separately attack its four modules, \ie, classification/regression branch of stage-1/2. Qualitative results are displayed in Fig.~\ref{fig:white_box} and more examples can be found in the supplementary. 
%We will report the results of three attack configurations respectively.
\vspace{-3em}
\begin{figure*}[t]
\centering
	\includegraphics[width=0.9\textwidth]{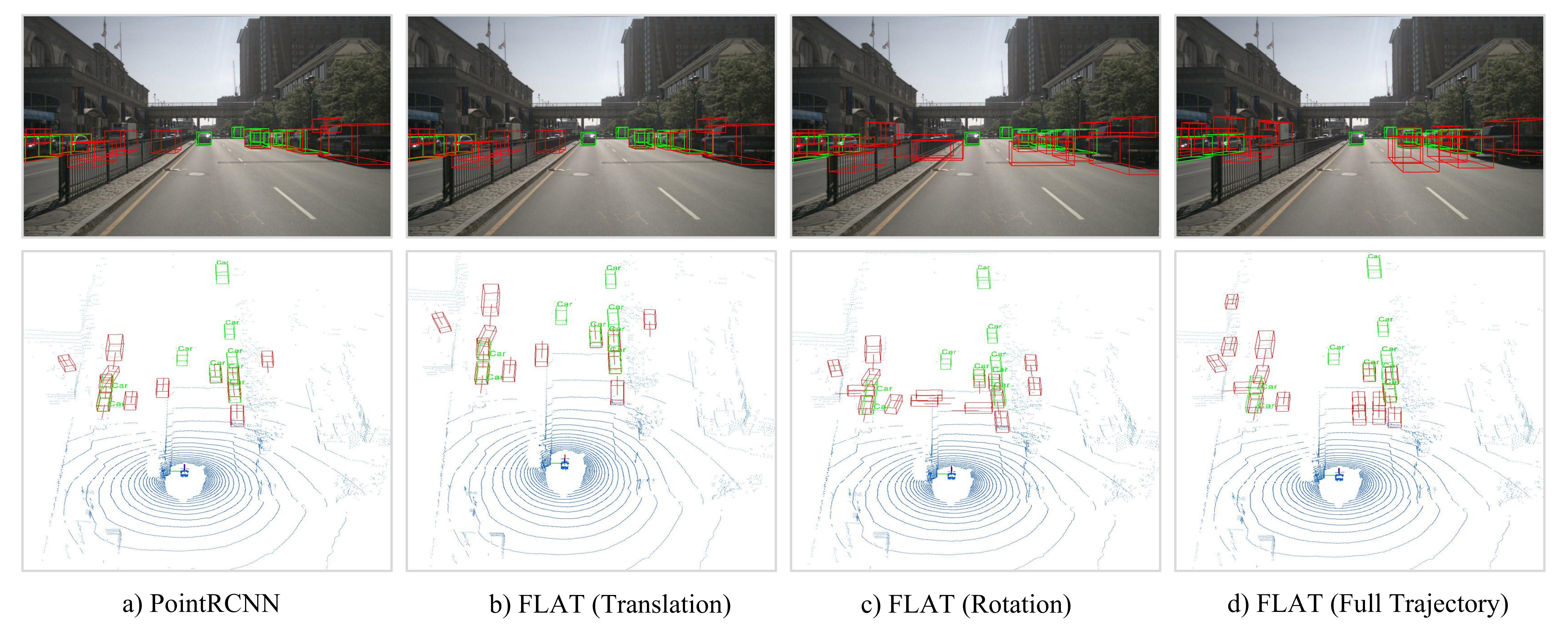}
	\caption{\textbf{Qualitative evaluations of white box attack.} False positives are increased and output are drifted by carefully crafting the trajectory.}
	\label{fig:white_box}
\end{figure*}
\vspace{1cm}
\begin{table*}[t]
\footnotesize
	\setlength{\tabcolsep}{2.5mm}
	\centering
	\caption{\textbf{Quantitative results of black box attack on nuScenes~\cite{Caesar2020nuScenesAM}:} the worst and second worst performances are highlighted by \tbred{red} and \tbblue{blue} color.}
	%\vspace{0.1cm}
	\label{tab:blackbox}
	\begin{spacing}{1}
		\begin{tabular}{c|cccccccccc|cc}
			\toprule
			{Category}                  & Car  & Pedes. & Barri. & Traff. & Truck & Bus  & Trail. & Const. & Motor. & Bicyc. & mAP  \\ \midrule
			{PointPillar++~\cite{Hu2020WhatYS}}    & 80.0 & 66.9   & 34.5   & 27.9   & 35.8  & 54.1 & 28.5   & 7.5    & 18.5   & \tbred{0.0}    & 35.4 \\ \midrule
			%Random Attack (GNSS$\&$INS)   & 49.7 & 20.0   & 16.5   & 5.5  & 19.9  & 22.2 & 20.2   & 1.4    & 0.9   & \tbred{0.0}    & 35.4 \\ 
			\textbf{FLAT (Translation)}  & 57.7 & 26.9   & 21.3   & 12.6    & 25.3  & 30.0 & 25.0   & 3.3    & 3.5    & \tbred{0.0}      & 20.6 \\ 
			\textbf{FLAT (Polynomial)} & {57.9} & {26.9}   & {21.3}   & {12.7}    & {25.4}  & {30.2} & {25.2}   & {3.4}    & {3.5}    & \tbred{0.0}      & {20.7} \\
			\textbf{FLAT (Rotation)}       & \tbblue{47.7} & \tbblue{21.1}   & \tbblue{18.6}   & \tbblue{8.9}    & \tbblue{20.2}  & \tbblue{23.5} & \tbblue{20.5}   & \tbblue{1.7}    & \tbblue{1.3}    & \tbred{0.0}      & \tbblue{16.4} \\ 
			\textbf{FLAT (Full Trajectory)} & \tbred{45.0} & \tbred{18.6}   & \tbred{16.4}   & \tbred{5.9}    & \tbred{18.3}  & \tbred{22.0} & \tbred{19.7}   & \tbred{1.3}    & \tbred{0.5}    & \tbred{0.0}      & \tbred{14.8} \\ 
			\bottomrule
		\end{tabular}
	\end{spacing}
	\vspace{-1.0em}
\end{table*}

\textbf{Attack Translation Only.} The quantitative results are exhibited in Table~\ref{tab:flat}. Simply adding random noise with a small standard deviation can largely decrease the performance, \eg, AP in the easy case is reduced by $30.44$ (around $64\%$). This phenomenon should stand as a warning for the autonomous driving community. Also, attacking classification branch of stage-2 is the most effective way to fool the detector: AP is respectively lowered by $35.72$ ($75.3\%$), $15.69$ ($72.8\%$) and $14.85$ ($71.0\%$) in the scenarios of easy, moderate and hard. This is because the stage-2 outputs the final predictions and is extremely safety-relevant. Moreover, we can find that attacking the classification branch is more effective than attacking regression, which is reasonable because classifying the objects takes priority over estimating their sizes in the detection task. Compared to the random attack, our method is more detrimental thanks to our making use of the vulnerability pointed out by the gradient. In easy, moderate, and hard scenarios, adversarial perturbation generated by attacking stage-2's classification can yield additional drops of $5.28$, $2.71$, and $2.84$ in AP, compared to the random perturbation. Besides, our attack is better than the coordinate attack due to the parameterization, validating the superiority of the trajectory attack.
\vspace{-0.19em}
%\noindent\textbf{Remark:} Directly manipulating point cloud is not as good as tampering with the trajectory. One potential reason is that we are attacking generating process of point cloud, and dividing point cloud into multiple sectors. Such grouping operation can make learning of perturbation easier. In contrast, learning a strong perturbation in point cloud space is much more difficult due to the large number of the points.
\begin{table}[t]
	\setlength{\tabcolsep}{1.0mm}
% 	\small
    % \footnotesize
    \scriptsize
	\centering
	\caption{\textbf{AP of FLAT with and without regularization.} $\lambda$ indicates regularization strength 
	%($\lambda_s$ in trajectory and $\lambda_d$ in point cloud).
	$\mathcal{S}$ denotes average trajectory variation and $\mathcal{D}$ denotes average point cloud distance ($\ell_{p}$ norm and Chamfer distance).}
	%\vspace{0.3cm}
	\label{tab:regular}
	\begin{spacing}{1}
		\begin{tabular}{c|c|c|c|ccc}
			\toprule
			{\multirow{2}{*}{Regularization}} & \multirow{2}{*}{$\lambda$}  & \multirow{2}{*}{$\mathcal{S}$/$\mathcal{D}$} & \multicolumn{3}{c}{AP} \\
			 &  &  &  Easy & Moderate & Hard  \\ \midrule
			{\multirow{2}{*}{Trajectory}} & 0 & {0.18} & 0.19 & 0.01 & 0.02  \\
			& \multirow{1}{*}{\textbf{0.01}}  & \textbf{0.17} & \textbf{0.28} & \textbf{0.04} & \textbf{0.02}   \\ \midrule
			\multirow{4}{*}{$\ell_{p}$ norm/Chamfer} & 0 & 27.54/28.70 & 0.19/0.19 & 0.01/0.01 & 0.02/0.02 \\
			 & \textbf{0.01} & \textbf{14.43}/\textbf{13.85} & \textbf{5.66}/\textbf{3.11} & \textbf{2.45}/\textbf{1.33} & \textbf{2.65}/\textbf{1.38}  \\
		     & 0.1 & 2.35/3.54 & 21.84/14.59 & 9.98/6.45 & 9.75/ 6.59  \\
			 & 1 & 0.95/2.04 & 23.65/19.19 & 11.22/9.21 & 11.33/9.40  \\  \bottomrule
		\end{tabular}
	\end{spacing}
	\vspace{-3em}
\end{table}
% \begin{figure*}[t]
% 	\includegraphics[width=1\textwidth]{figures/fig_reg.pdf}
% 	\caption{Qualitative results of attack with and without regularization. A minor perturbation in point cloud can fool the detector.}
% 	\label{fig:reg}
% \end{figure*}
%--------------------------------------------------------

\begin{figure*}[t]
\centering
	\includegraphics[width=0.94\textwidth]{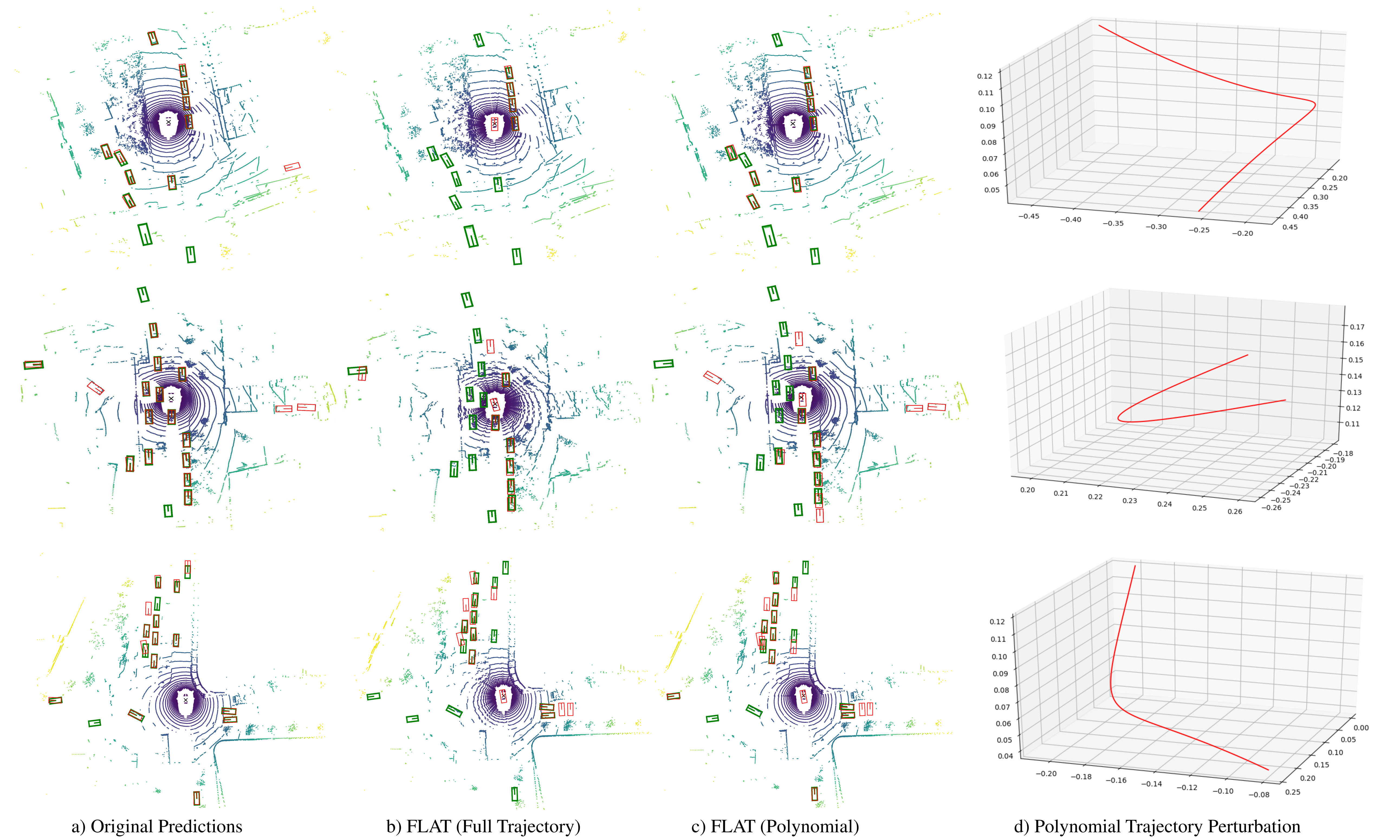}
	\caption{\textbf{Point cloud visualization and qualitative results of black box attack.} a) Raw detections of the original detector PointPillar++~\cite{Hu2020WhatYS}. b) The output of the detector after attacking the full trajectory. c) The output of the detector after polynomial trajectory perturbation in the euclidean space. d) The polynomial translation perturbation visualized in $xyz$ space, the units of three axes are all meters. \tbgreen{Green}/\tbred{red} boxes denote the ground truth/prediction.}
	\label{fig:black_box}
	\vspace{-1em}
\end{figure*}

\textbf{Polynomial Trajectory Perturbation.} When attacking polynomial coefficients instead of the individual trajectory point, 
the performance is still on par with the discrete setting, as shown in Table~\ref{tab:flat}, yet the attack is highly imperceptible especially in the point cloud space as shown in Fig.~\ref{fig:black_box}.

\textbf{Attack Rotation Only.}
Coordinate attack is manipulating each point's xyz position. In contrast, our method treat each LiDAR packet as a rigid body, therefore we can attack the rotation. From Table~\ref{tab:flat} we can find that attacking classification in stage-2 is still the most powerful way to fool PointRCNN. Besides, attacking rotation achieves more performance drop compared to attacking translation, \eg, in easy, moderate, and hard situations, AP is respectively decreased by $45.09$ ($95.0\%$), $20.76$ ($96.3\%$) and $20.30$ ($97.1\%$) in comparison with original PointRCNN. Moreover, fooling regression in stage-1 achieves the second best attacking quality, proving the effectiveness of attacking the fundamental proposal generation. Besides, attacking regression of stage-2  has the worst attacking quality, proving that attacking the refinement of box size has no significant effect. Compared to the random attack, attacking stage-2's classification has realized additional AP drop ($9.95$, $4.07$, $4.52$), validating the merits of adversarial learning.

%--------------------------------------------------------
\textbf{Attack Full Trajectory.}
As shown in Table~\ref{tab:flat}, fooling the full trajectory has achieved the best attacking quality, \eg, while attacking classification of stage-2, AP can be decreased to nearly zero. Compared to original PointRCNN, AP is respectively lowered by $47.25$ ($99.6\%$) in easy scenarios, indicating that the detector is completely blinded.

%--------------------------------------------------------
\textbf{Attack with Regularization.}
As shown in Table~\ref{tab:regular}, for regularizing trajectory smoothness, average trajectory variation is slightly lowered and the performance is still on par with attacking without regularization. For regularizing point cloud change, when $\lambda=0.01$, point distance is decreased by $47.6\%$ ($\ell_{p}$ norm) and $51.7\%$ (Chamfer distance), while the APs in three scenarios are still very low, thus an advanced imperceptible attack can be realized with regularization. Qualitative examples are in the supplementary.

\subsection{Black Box Attack}
We choose voxel-based PointPillar++~\cite{Hu2020WhatYS} to test the transferability across different input representation. The quantitative results on ten categories are shown in Table~\ref{tab:blackbox}: the performance is still largely dropped by our attack, \eg, mAP on ten categories can be reduced by $20.6$ ($58.2\%$), and the AP on car is decreased by $35.0$ ($43.8\%$). Meanwhile, our method has demonstrated satisfactory transferability across categories. With only adversarial learning against the car detector, the detection of other categories are also deceived. For small object like pedestrian/motorcycle, AP can be dropped by $48.3$ ($72.2\%$)/$18.0$ ($97.3\%$), which has demonstrated the superior attacking quality of our method against small object detector. This superiority is mainly because the small object with less points is more susceptible to the perturbation compared to the large object like bus (dropped by $32.1$, $59.3\%$) or truck (dropped by $17.5$, $48.9\%$). Several qualitative examples are displayed in Fig.~\ref{fig:black_box} and more examples are in the supplementary.
\section{Conclusion}\label{sec:conclusion}
We proposed a generic and feasible DNN attack pipeline based on the trajectory against LiDAR perception. We conduct experiments on the well-studied 3D object detection task. In white box attack, even only with a 10cm perturbation in translations, the precision can be dropped by around 70$\%$. While attacking the full trajectory, the precision can be decreased to nearly zero, yet the attack is less perceptible (especially the point clouds). Our attack also shows good transferability across various input representations and target categories, raising a red flag for perception systems using LiDAR and DNN jointly.
% \vspace{-3em}
\vspace{1em}

\textbf{Acknowledgment.} The research is supported by NSF FW-HTF program under DUE-2026479. The authors gratefully acknowledge the useful comments and suggestions from Yong Xiao, Wenxiao Wang, Chenzhuang Du, Wang Zhao, Ziyuan Huang, Hang Zhao and Siheng Chen, and also thank Yan Wang, Shaoshuai Shi and Peiyun Wu for their helpful open-source code.
\vspace{-3em}
{\small
\bibliographystyle{ieee_fullname}
\bibliography{egbib}
}

\renewcommand{\thetable}{\Roman{table}}
\renewcommand{\thefigure}{\Roman{figure}}
\renewcommand\thesection{\Roman {section}}

\section*{Appendix}
\setcounter{section}{0}
\setcounter{figure}{0}
\setcounter{table}{0}

\section{Overview}
In the supplementary, we first present more results of our FLAT attack against black box PointPillar++ [11]. Then more qualitative evaluations of white box attack are shown, and the performances under different parameters (the number of attack iteration, the step size for each iteration, and the interpolation step) are discussed. Finally, more examples of  \textit{\textbf{polynomial trajectory perturbation}} (including both the point cloud after our attack and and the trajectory perturbation) are presented to validate the imperceptibility in point cloud space and the strong smoothness in trajectory.
\section{Black Box Attack}\label{sec:black}
\subsection{Qualitative Evaluation}
\begin{figure}[t]
	\includegraphics[width=0.47\textwidth]{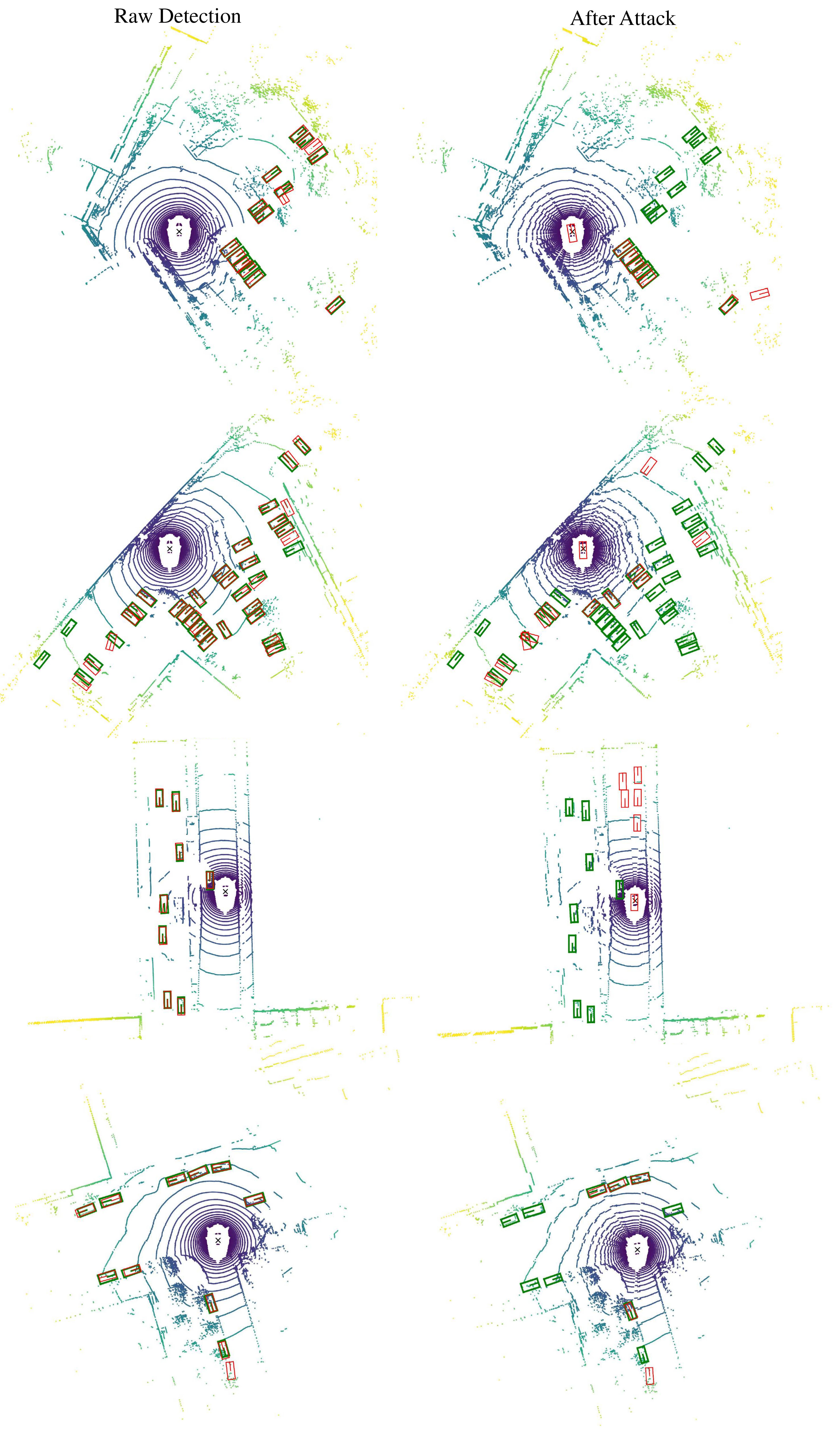}
	\caption{Qualitative results of black box attack when attacking the full trajectory. The left and right figures are respectively original and distorted LiDAR sweep as well as the detection results. \tbgreen{Green}/\tbred{red} boxes denote the ground truth/prediction respectively.}
	\label{fig:supp_blackbox1}
	% 	\vspace{-3mm}
\end{figure}

\begin{figure*}[t]
	\includegraphics[width=0.97\textwidth]{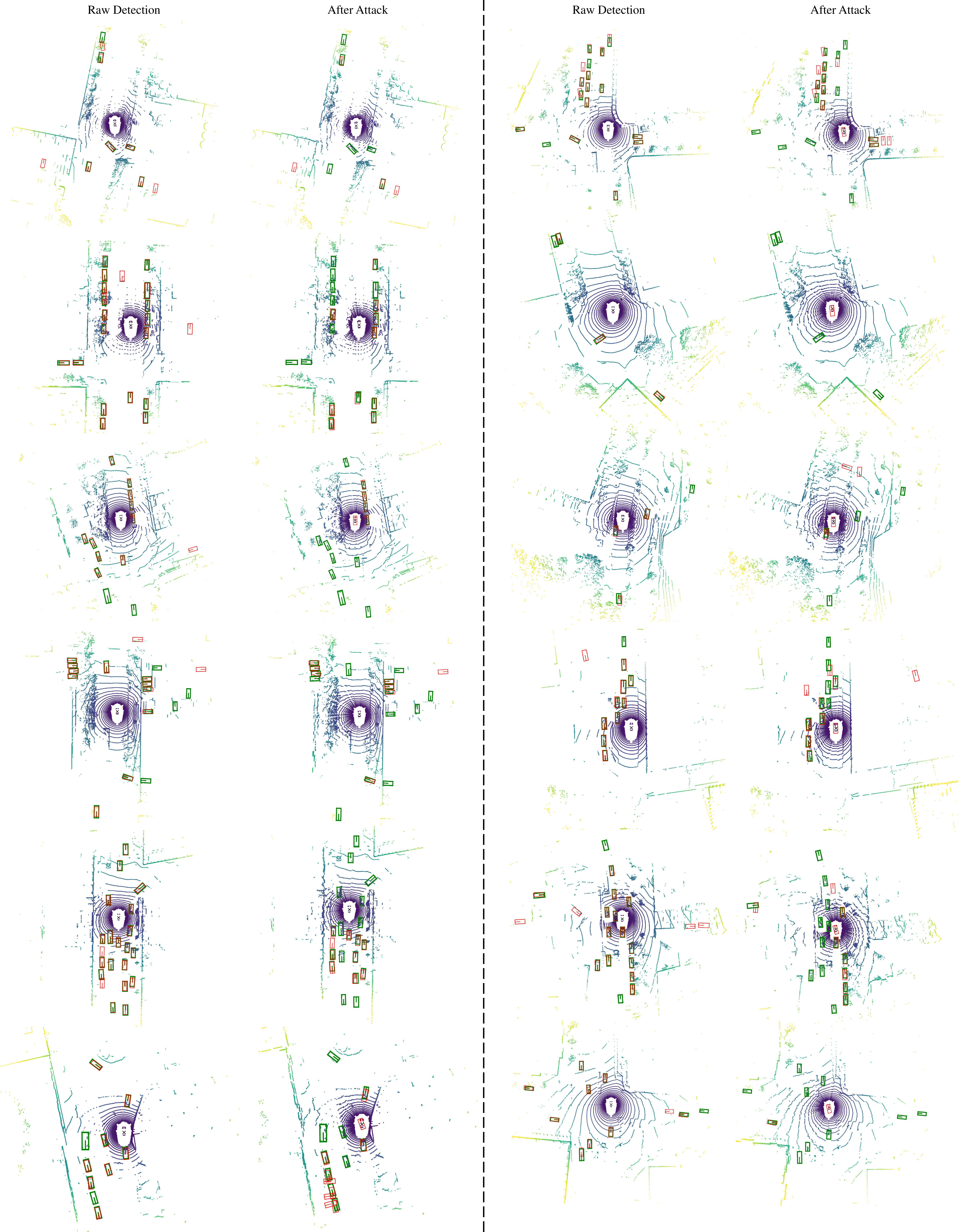}
	\caption{Qualitative results of the black box attack when attacking the full trajectory. \tbgreen{Green}/\tbred{red} boxes denote the ground truth/prediction respectively.}
	\label{fig:supp_blackbox2}
	\vspace{3mm}
\end{figure*}

Additional qualitative examples\footnote{Noted that we transfer the adversarial perturbation of the white-box \textit{full trajectory} attack targeting \textit{classification of stage-2}} are presented in Fig.~\ref{fig:supp_blackbox1} and Fig.~\ref{fig:supp_blackbox2}. Although our FLAT attack cannot know the model parameters of PointPillar++, a subtle perturbation can make the detector lose many safety-critical objects, \eg, as for the four sweeps in Fig.~\ref{fig:supp_blackbox1}, the adversarial perturbation in the full trajectory can make the detector lose 7, 20, 8 and 6 objects respectively (from top to bottom), severely damaging the self-driving car's perception module. Moreover, false positives are also increased by our perturbation (6 more in the third row of Fig.~\ref{fig:supp_blackbox1}), making the car mistakenly believe that there are obstacles in the free space.
\subsection{Quantitative Evaluation}
\begin{figure*}[t]
	\begin{center}

		\subfigure[t] {
			\begin{minipage}{0.186\textwidth}
				\centering
				\includegraphics[width=1\columnwidth]{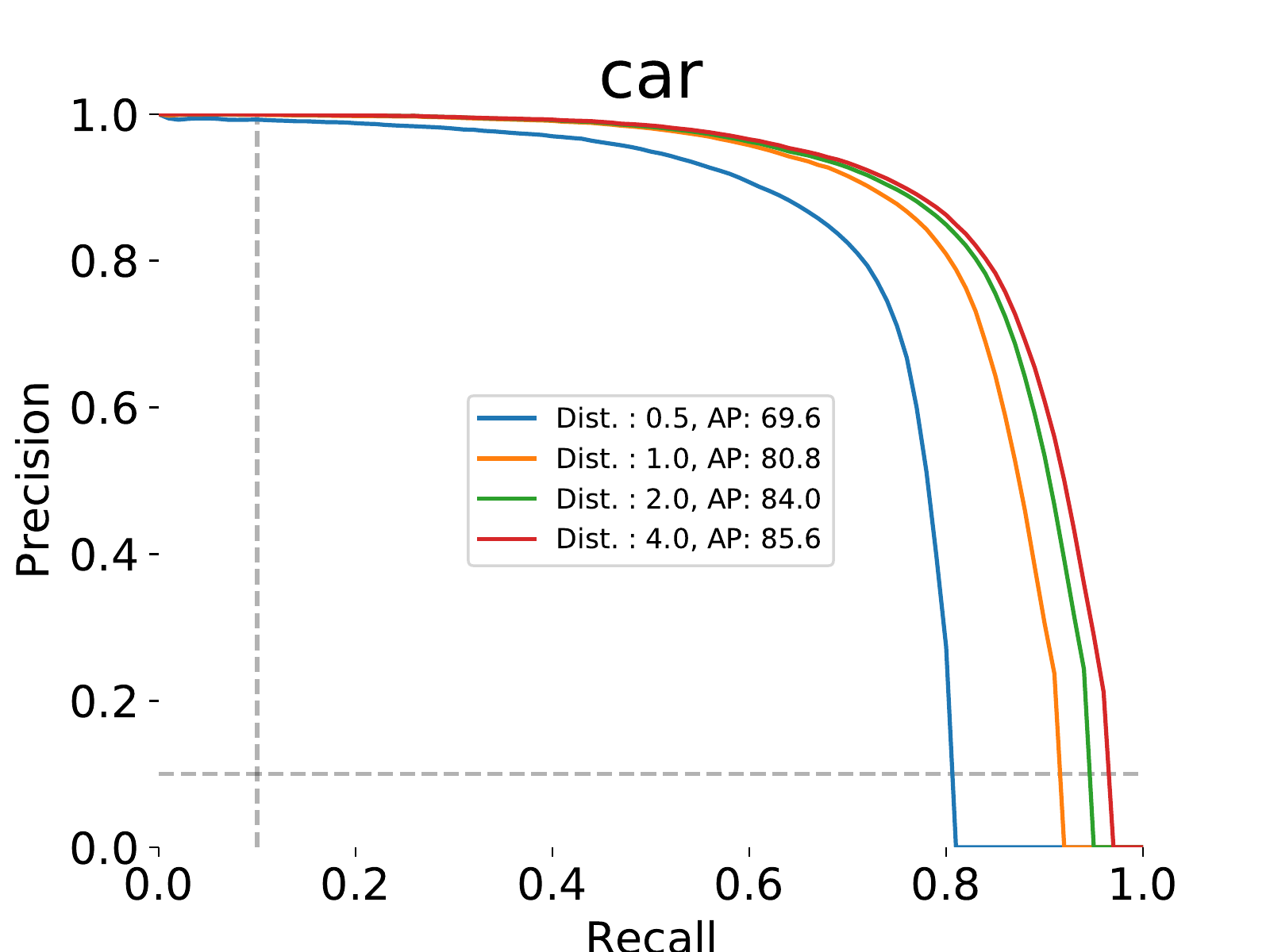}
				\\
				\includegraphics[width=1\columnwidth]{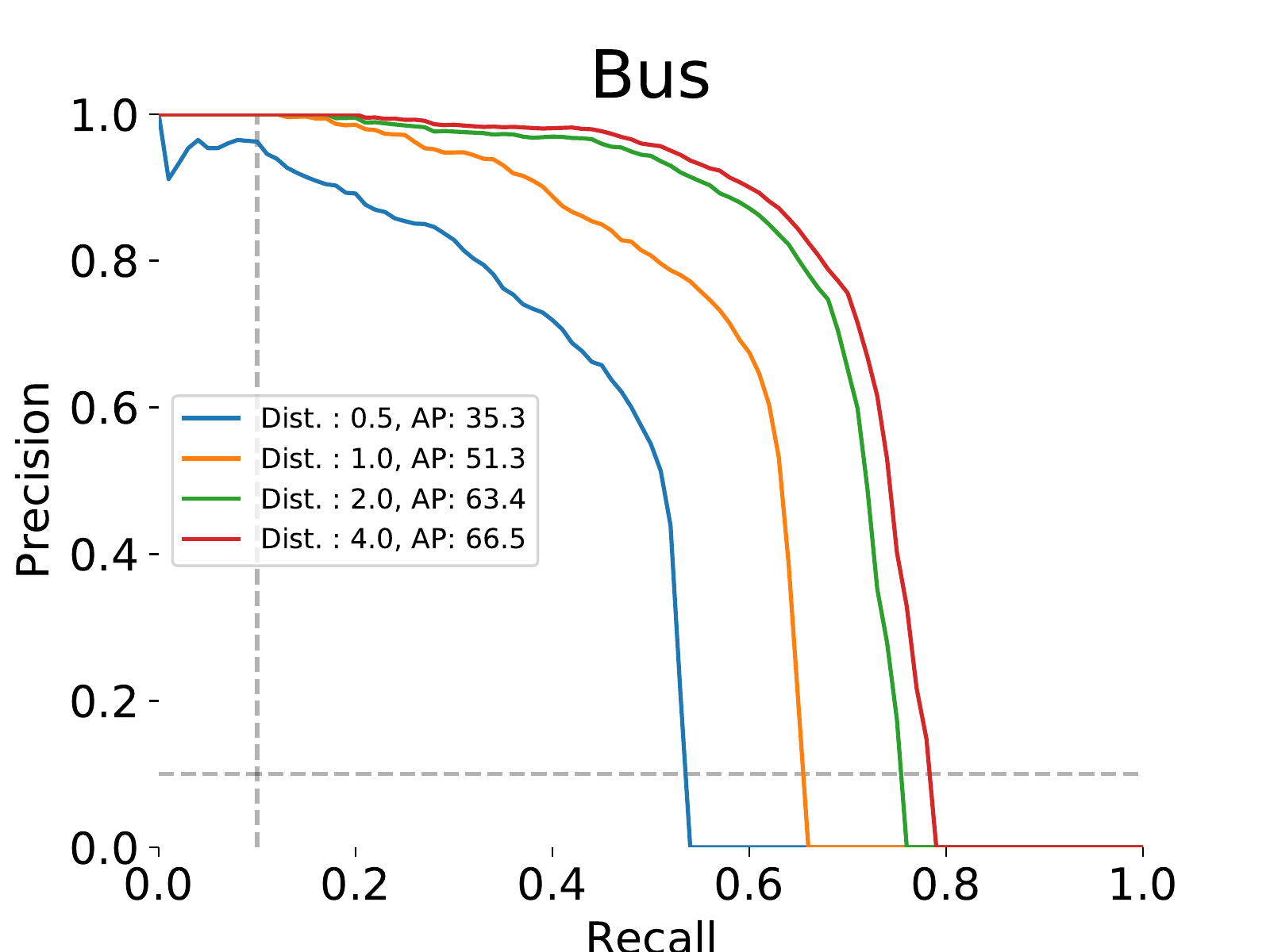}
			\end{minipage}
		}
		\subfigure[t] {
			\begin{minipage}{0.186\textwidth}
				\centering
				\includegraphics[width=1\columnwidth]{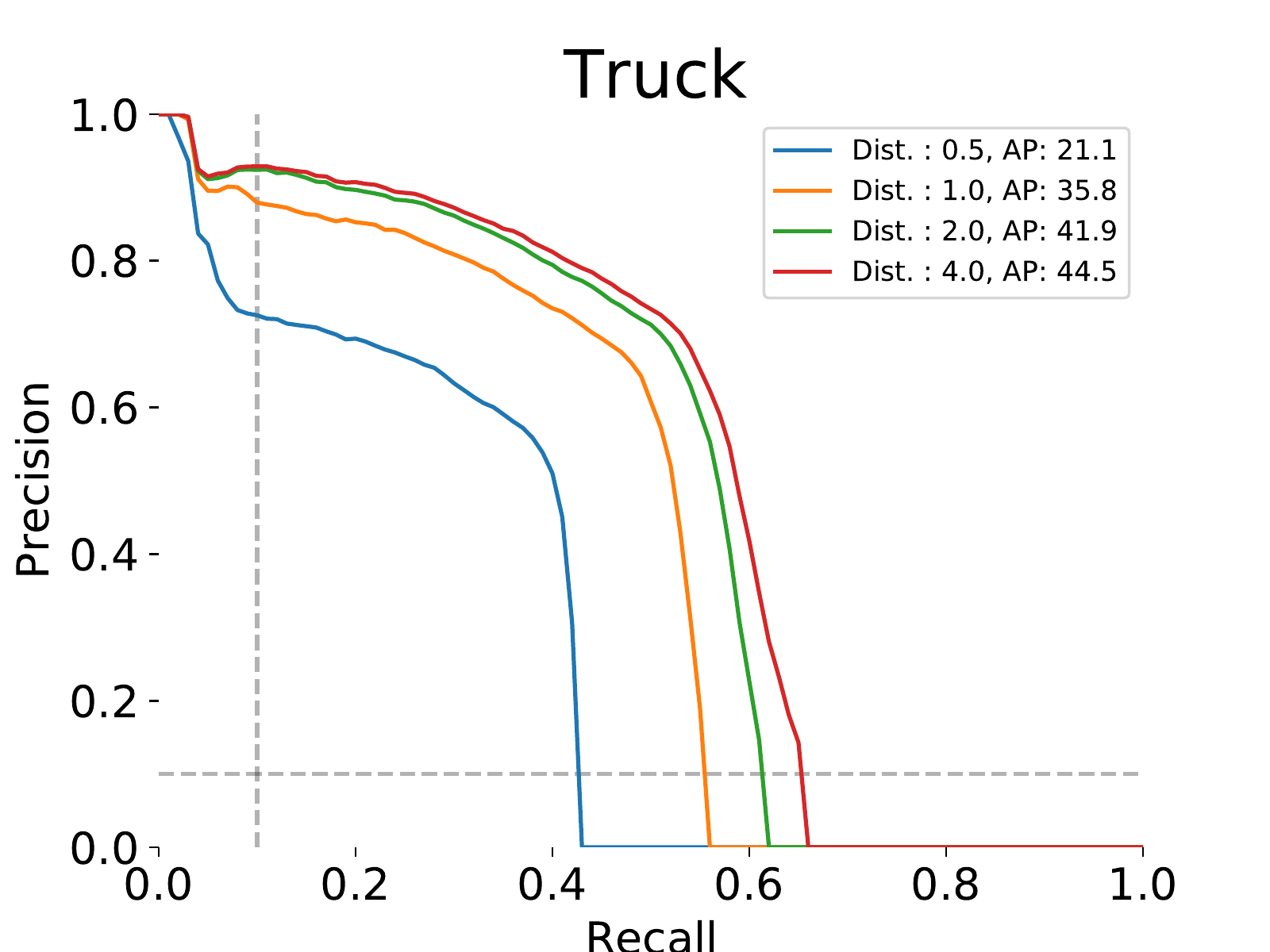}
				\\
				\includegraphics[width=1\columnwidth]{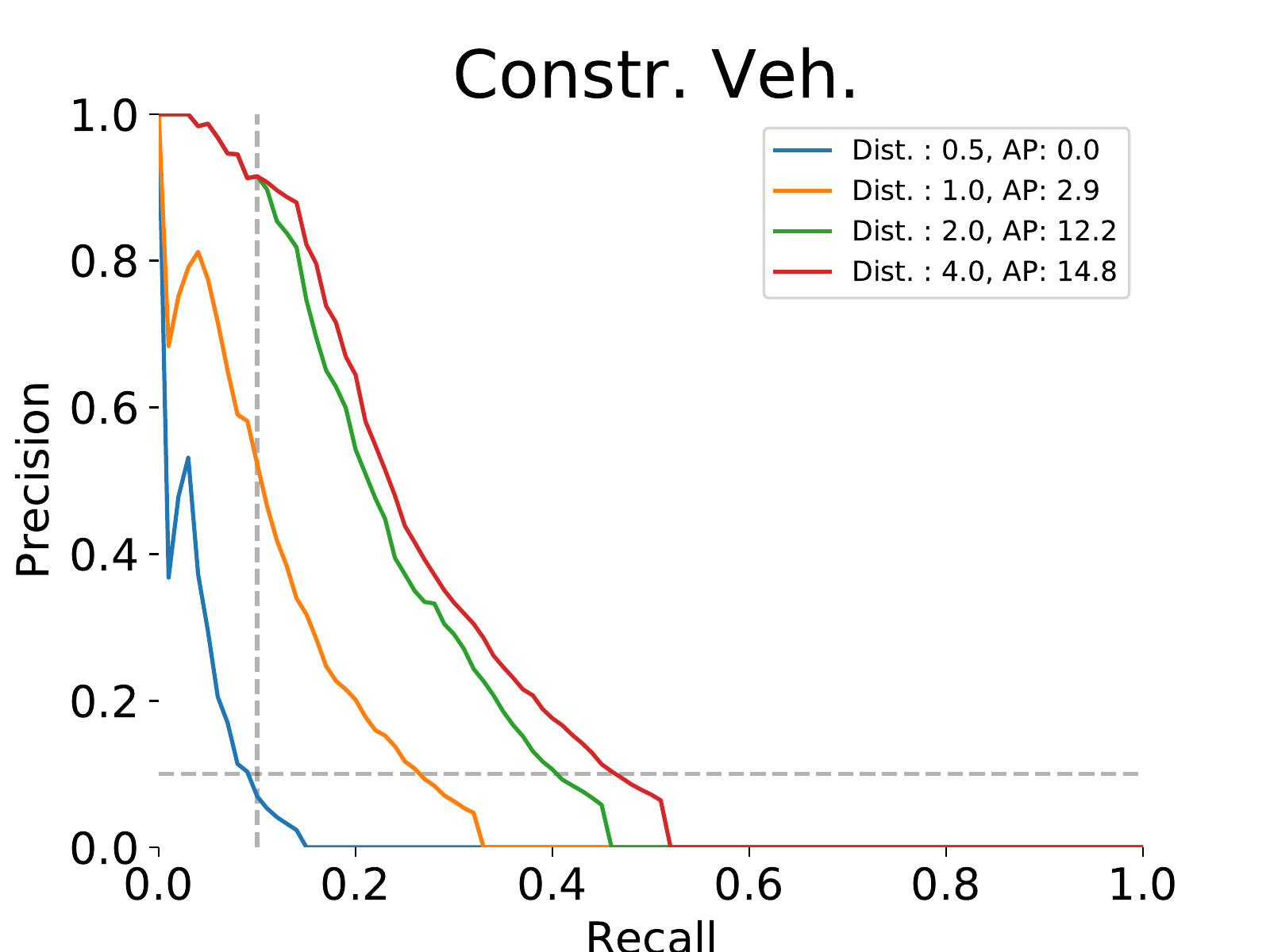}
			\end{minipage}
		}
		\subfigure[t] {
			\begin{minipage}{0.186\textwidth}
				\centering
				\includegraphics[width=1\columnwidth]{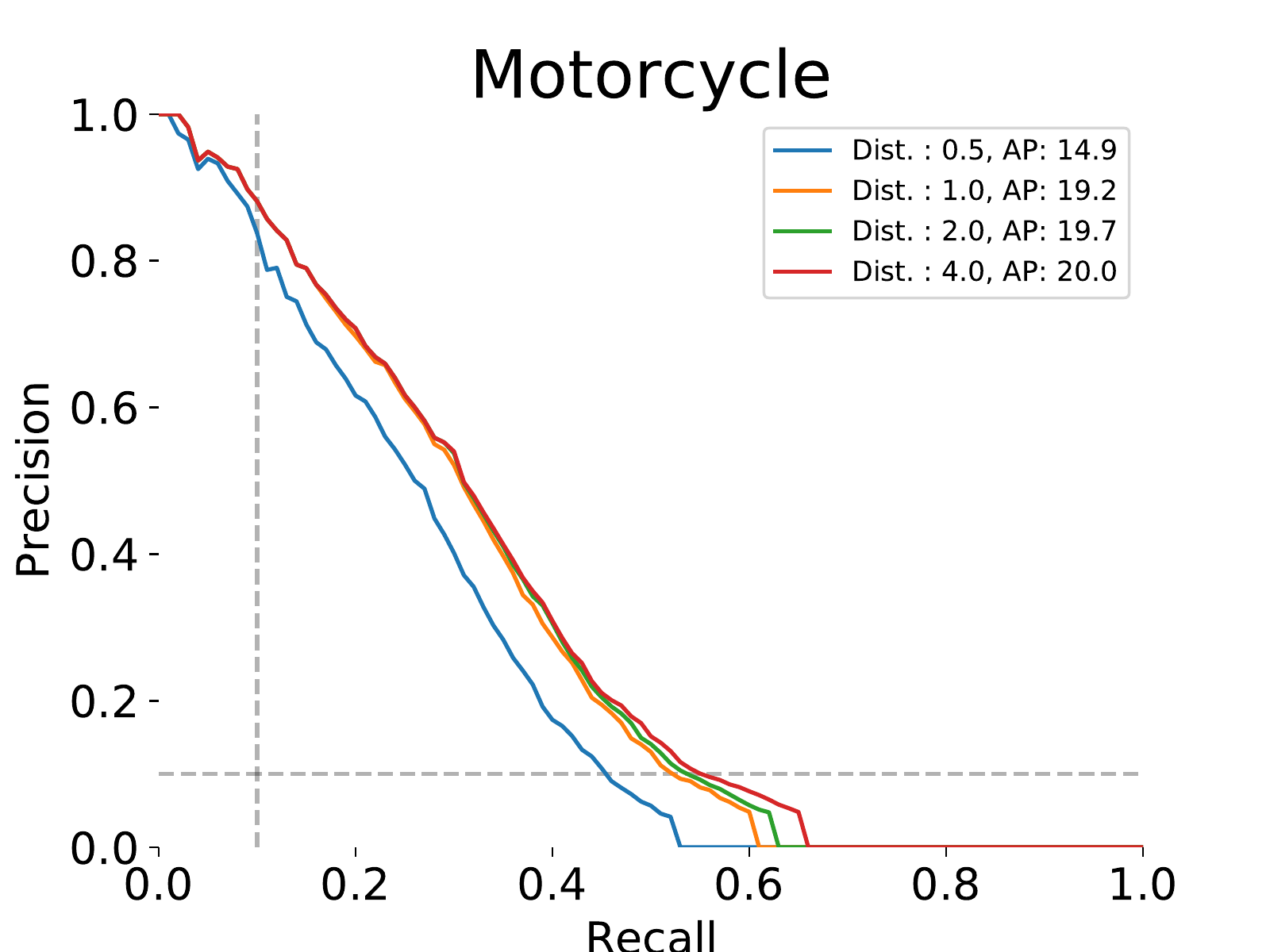}
				\\
				\includegraphics[width=1\columnwidth]{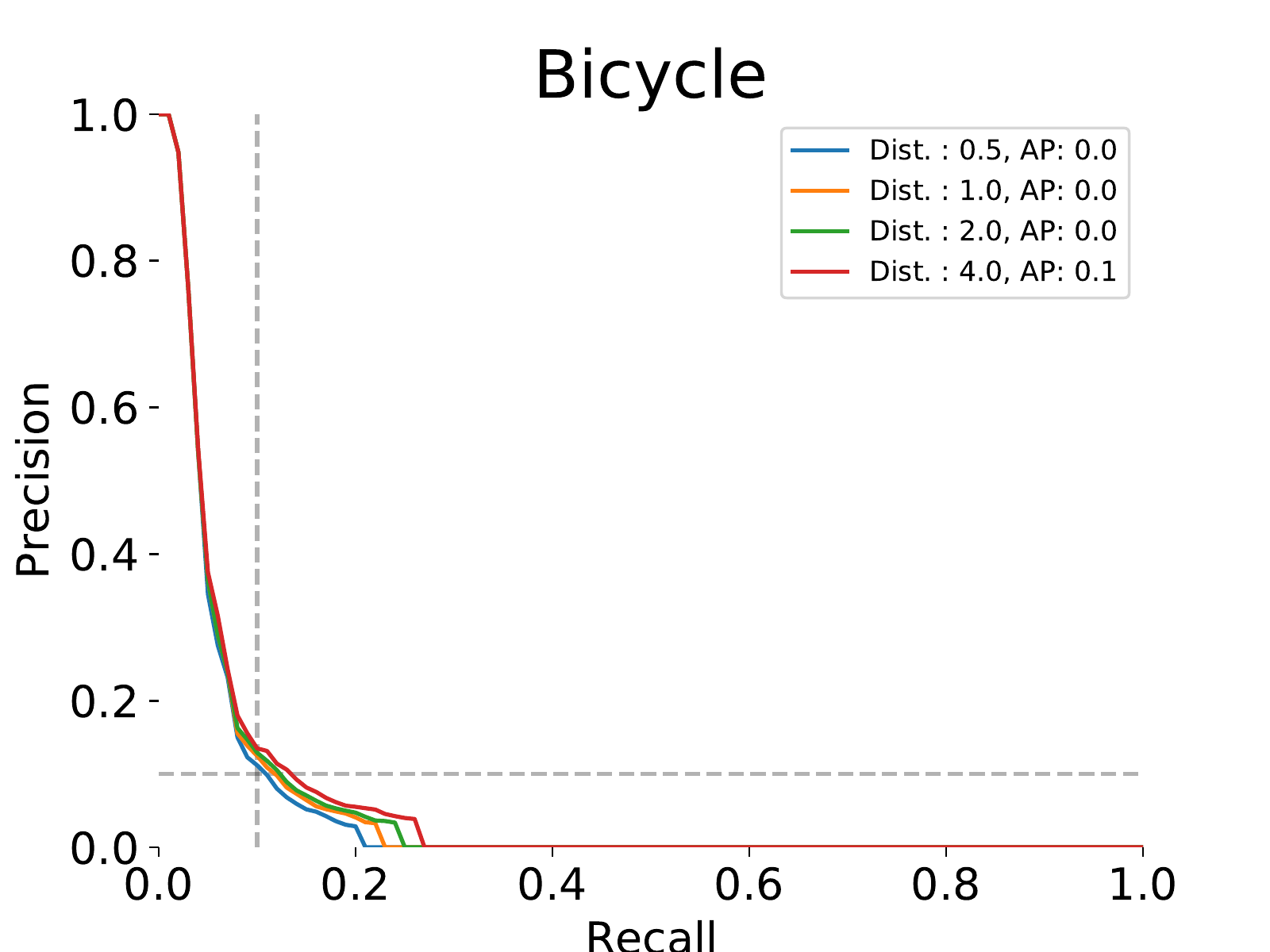}
			\end{minipage}
		}
		\subfigure[t] {
			\begin{minipage}{0.186\textwidth}
				\centering
				\includegraphics[width=1\columnwidth]{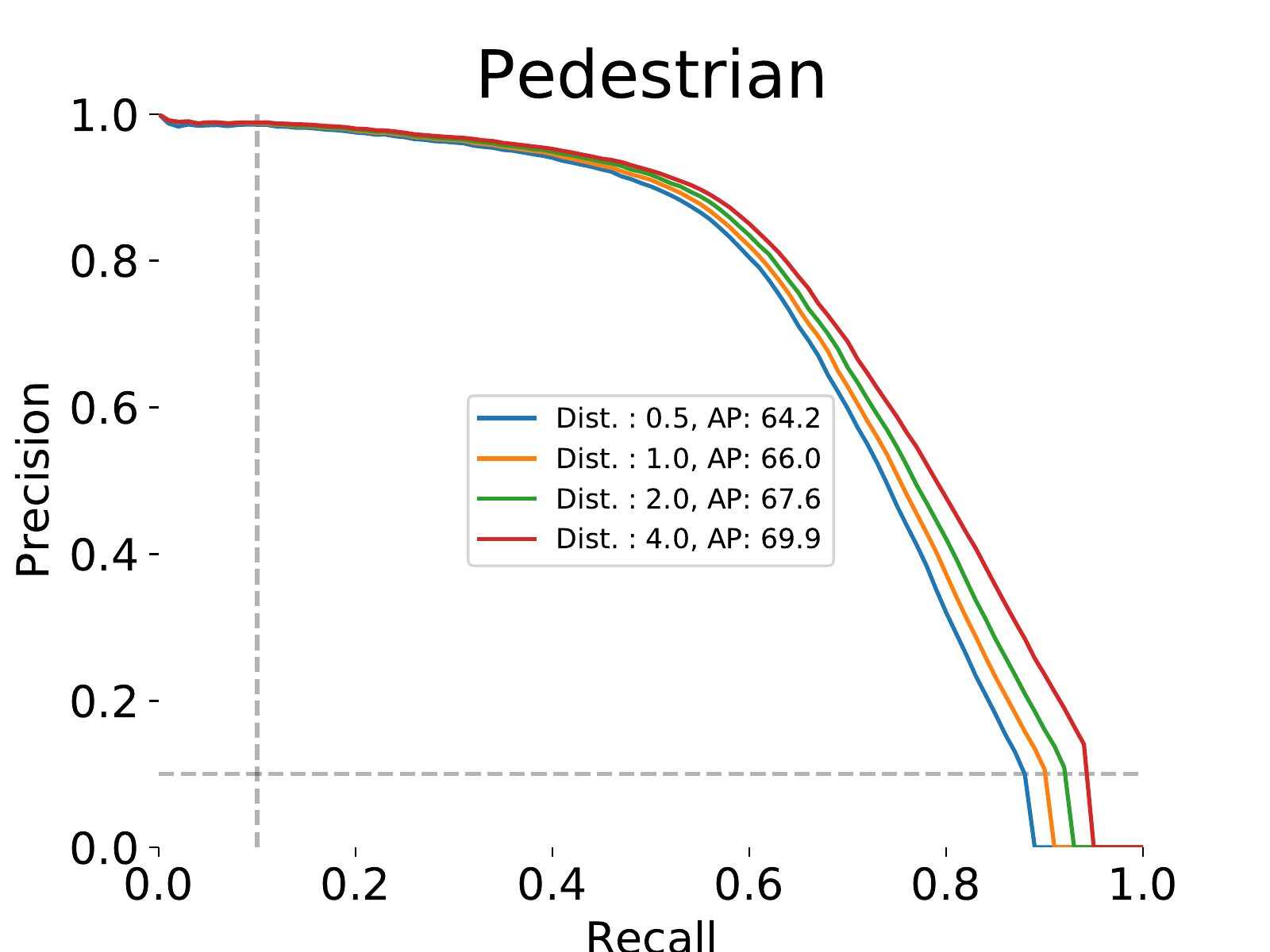}
				\\
				\includegraphics[width=1\columnwidth]{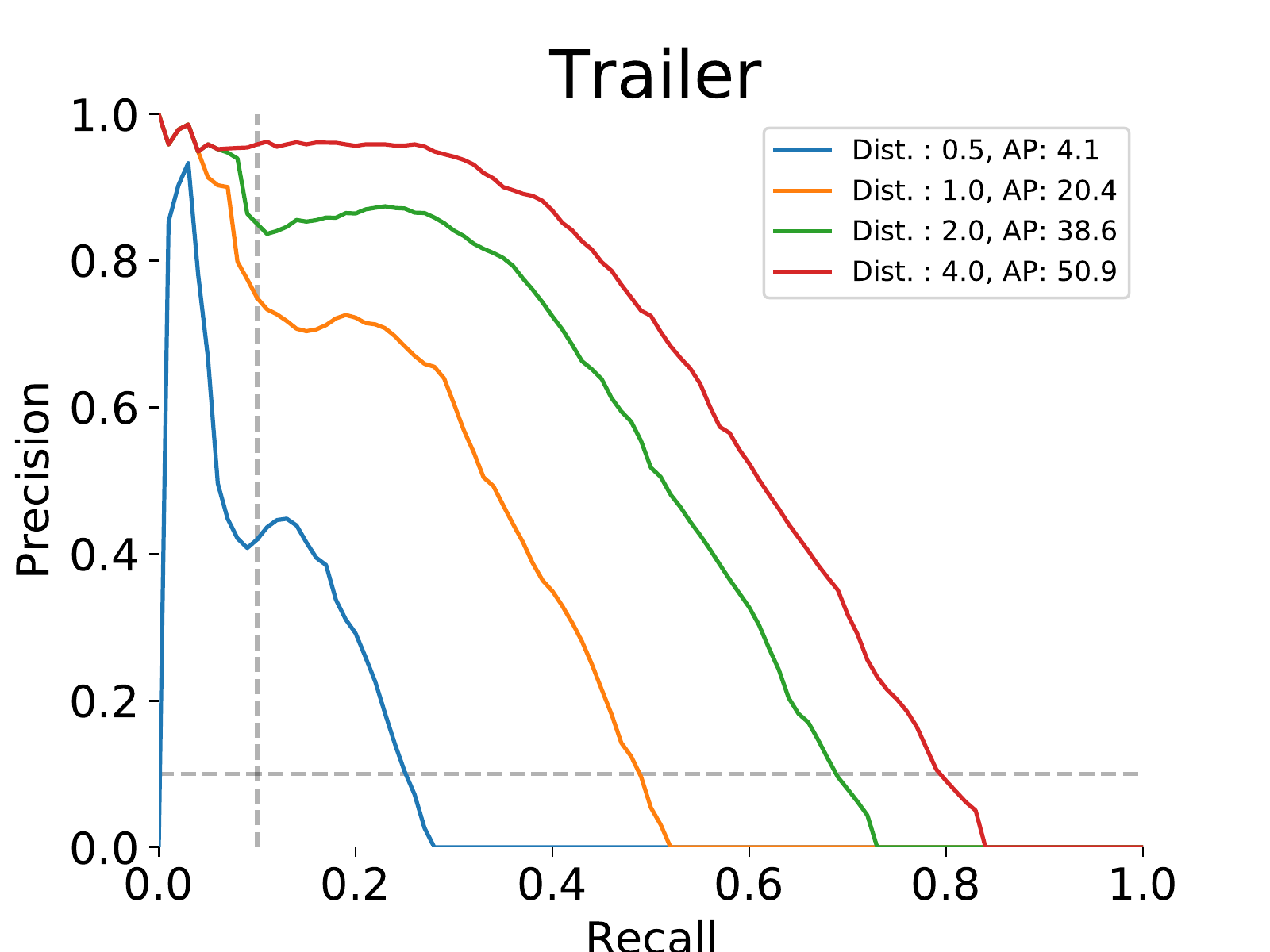}
			\end{minipage}
		}
		\subfigure[t] {
			\begin{minipage}{0.186\textwidth}
				\centering
				\includegraphics[width=1\columnwidth]{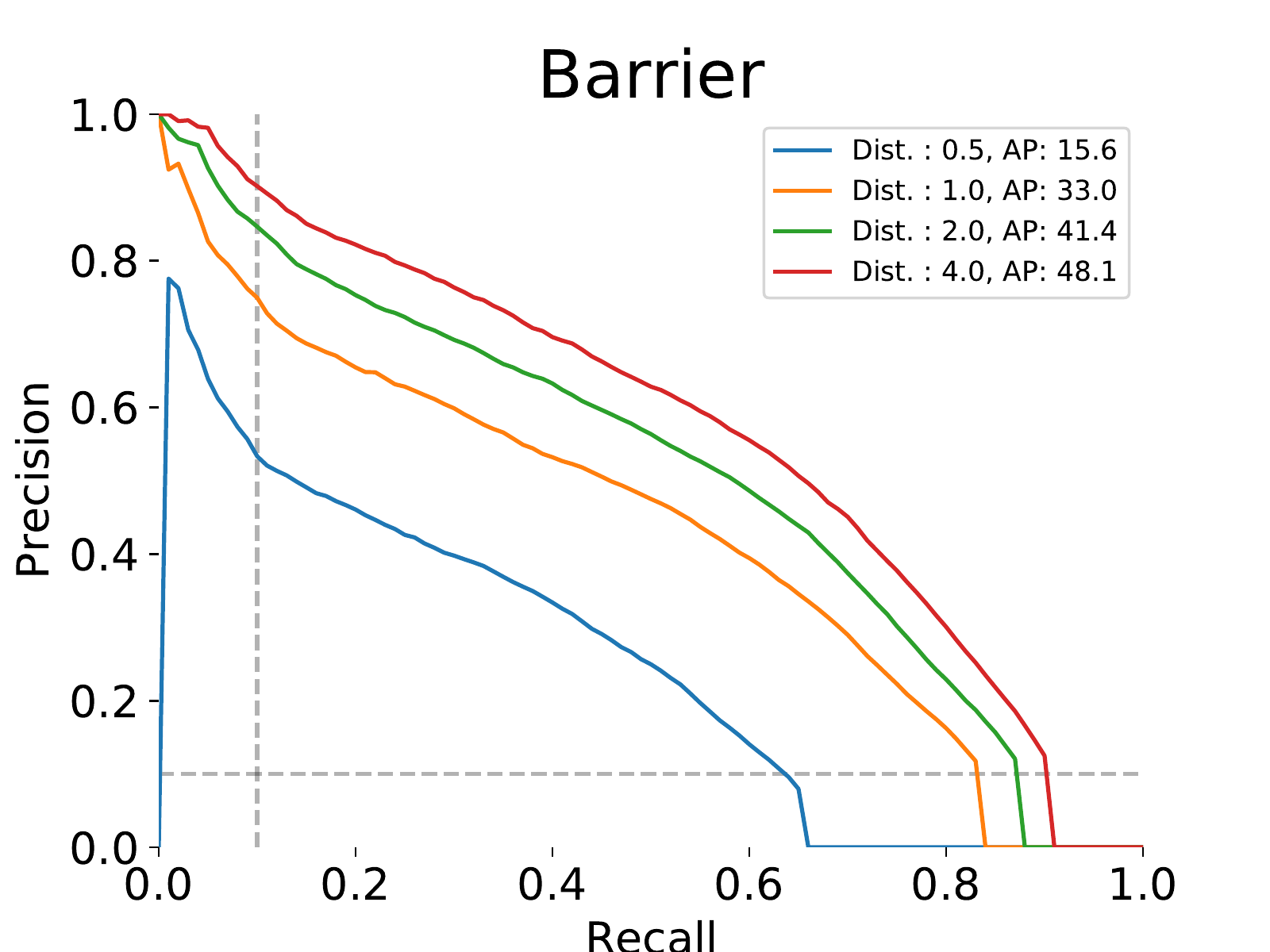}
				\\
				\includegraphics[width=1\columnwidth]{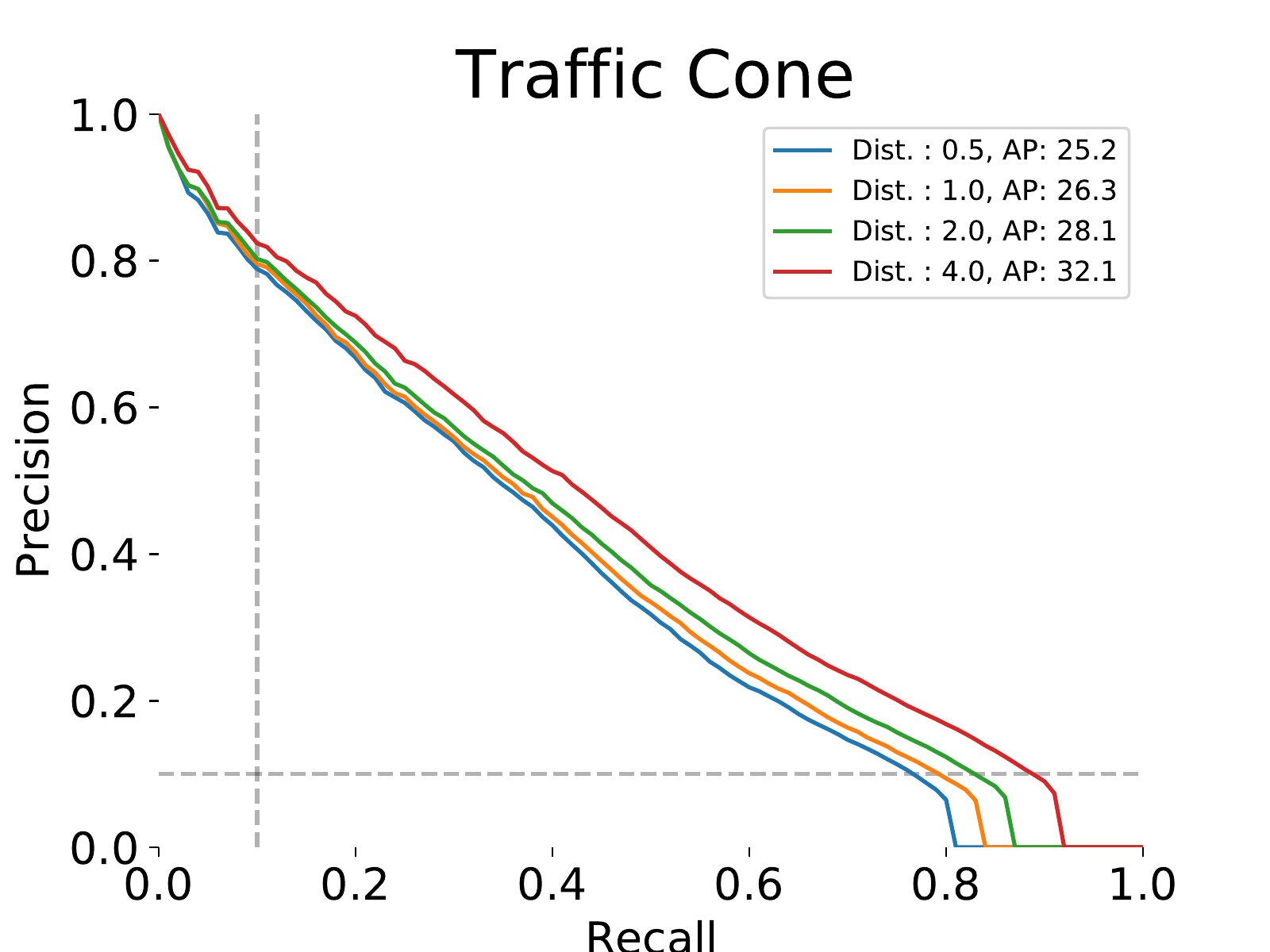}
			\end{minipage}
		}		
	\end{center}
	
	\caption{Per category precision-recall plot of \textbf{PointPillar++}  on the nuScenes validation set [2].}
	\label{fig:black_original}
	%\label{fig:onecol}
\end{figure*}

\begin{figure*}[t]
	\begin{center}

		\subfigure[t] { 
			\begin{minipage}{0.186\textwidth}
				\centering
				\includegraphics[width=1\columnwidth]{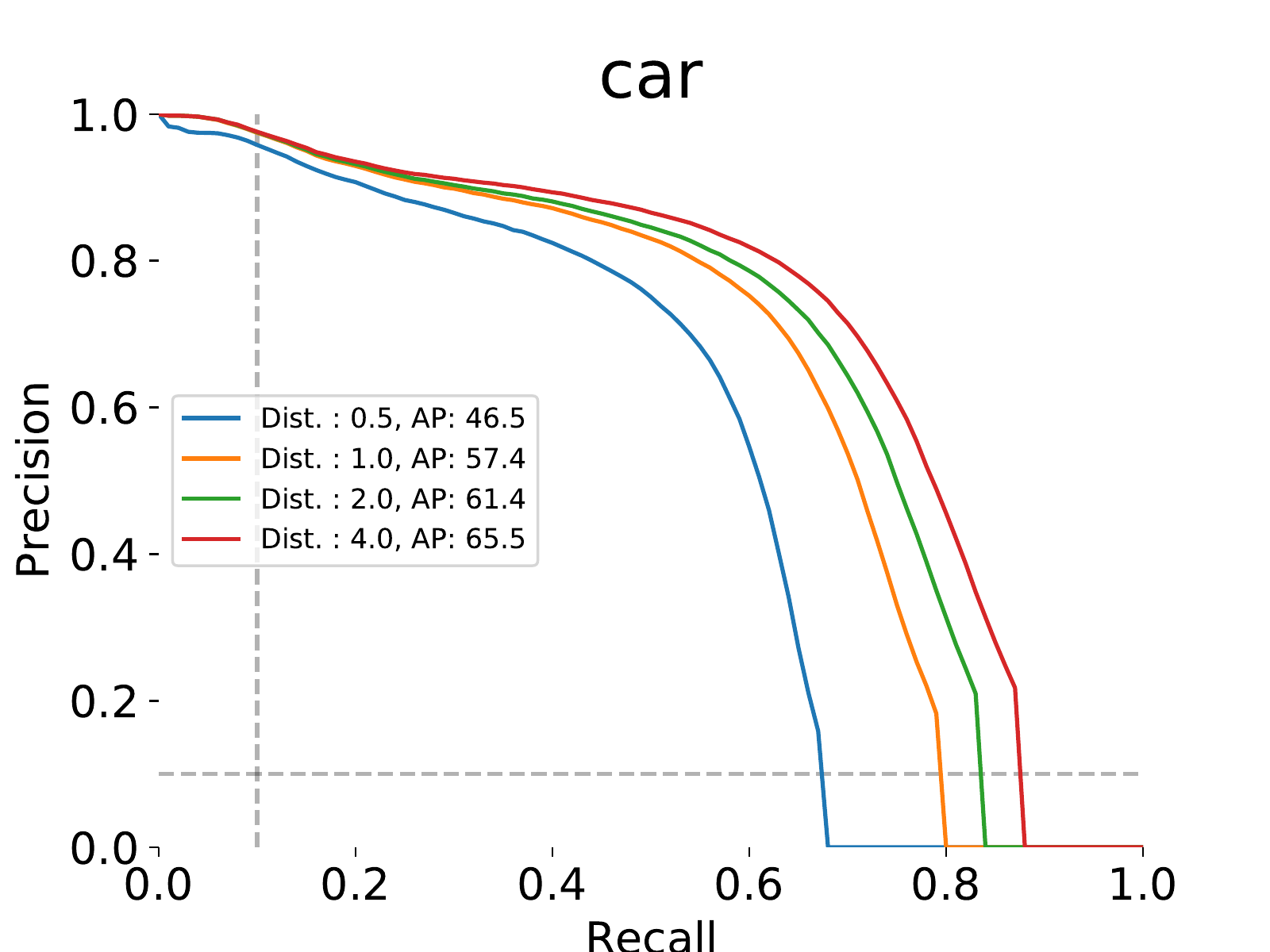}
				\\
				\includegraphics[width=1\columnwidth]{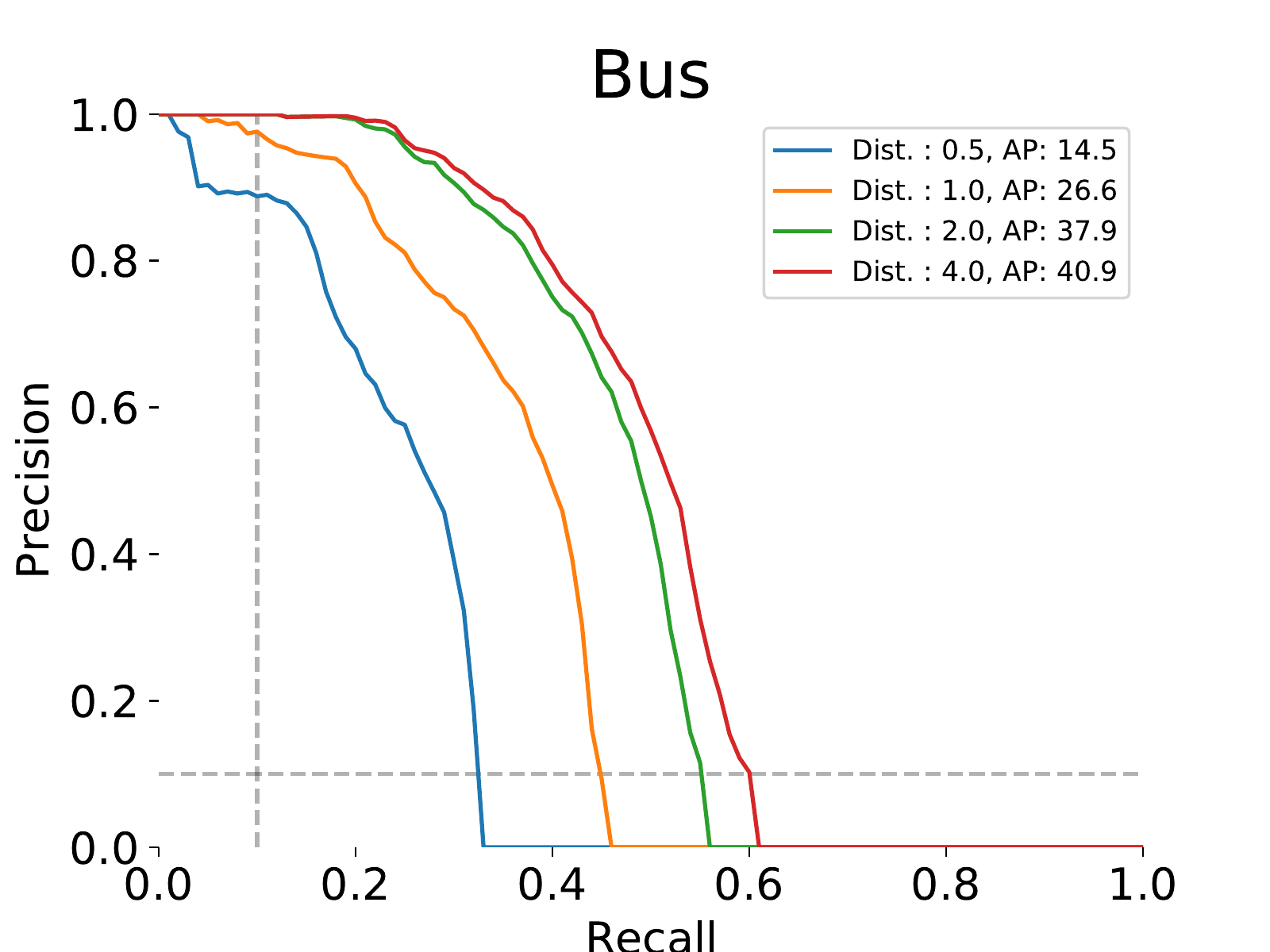}
			\end{minipage}
		}
		\subfigure[t] { 
			\begin{minipage}{0.186\textwidth}
				\centering
				\includegraphics[width=1\columnwidth]{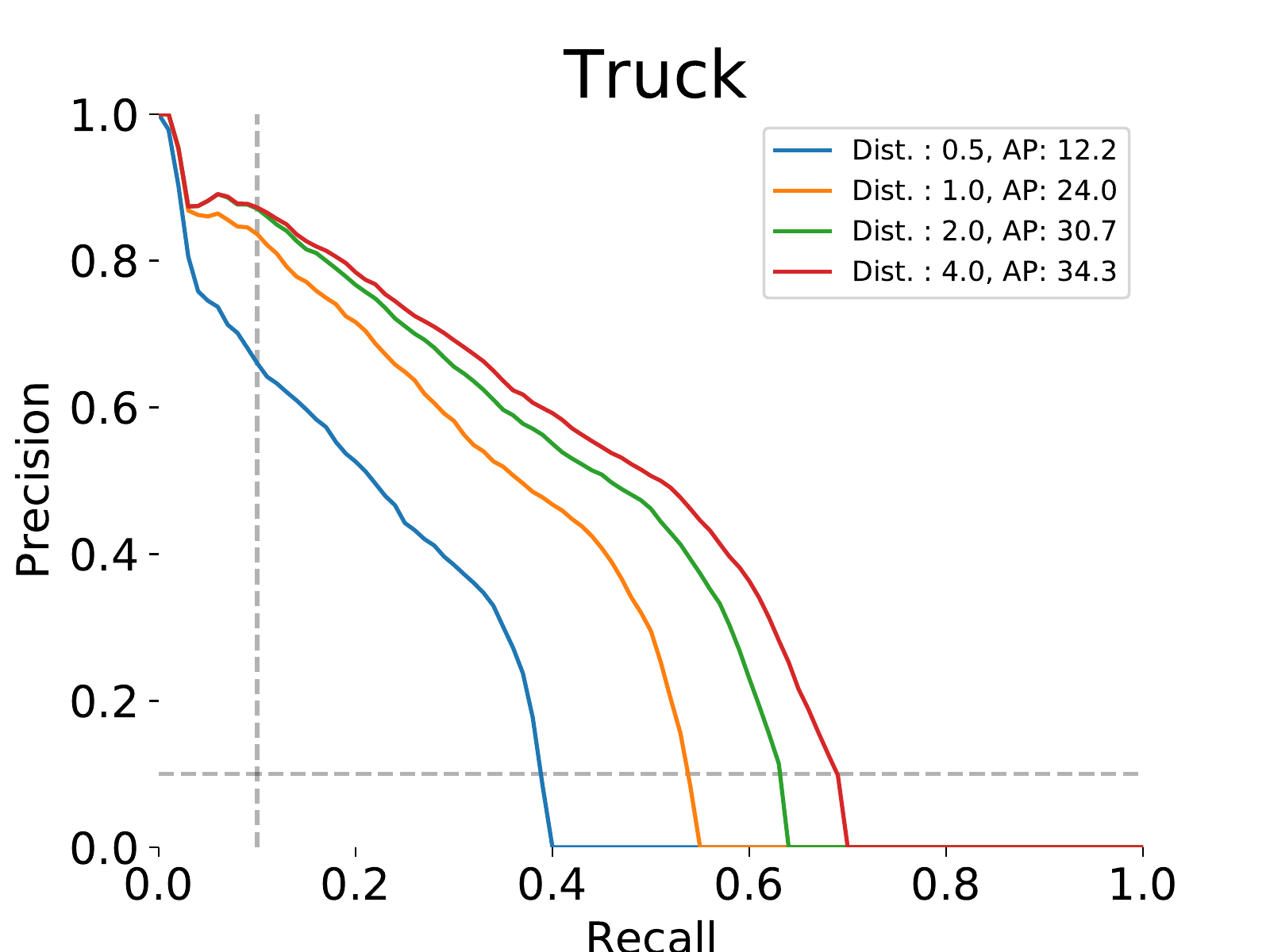}
				\\
				\includegraphics[width=1\columnwidth]{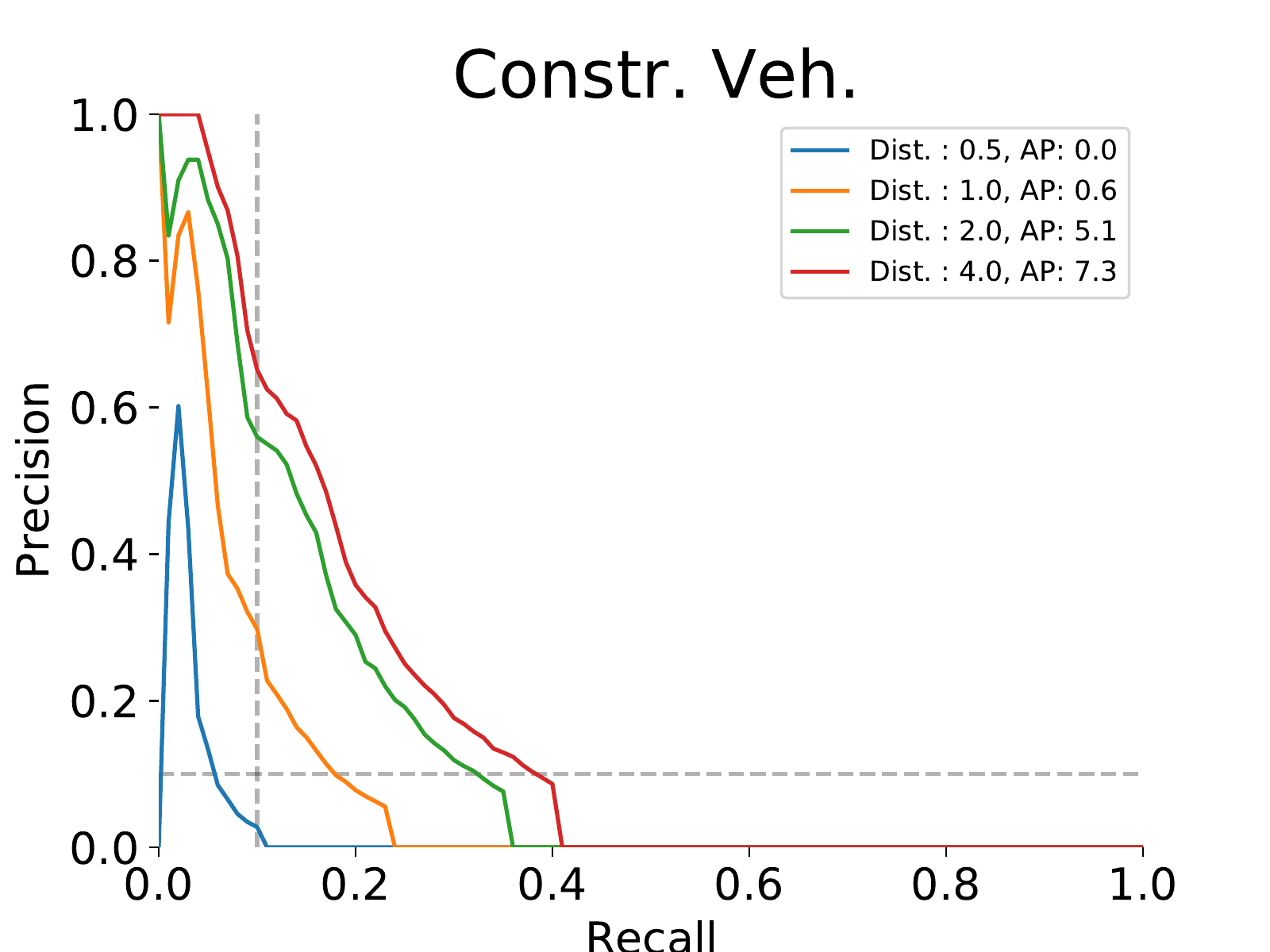}
			\end{minipage}
		}
		\subfigure[t] { 
			\begin{minipage}{0.186\textwidth}
				\centering
				\includegraphics[width=1\columnwidth]{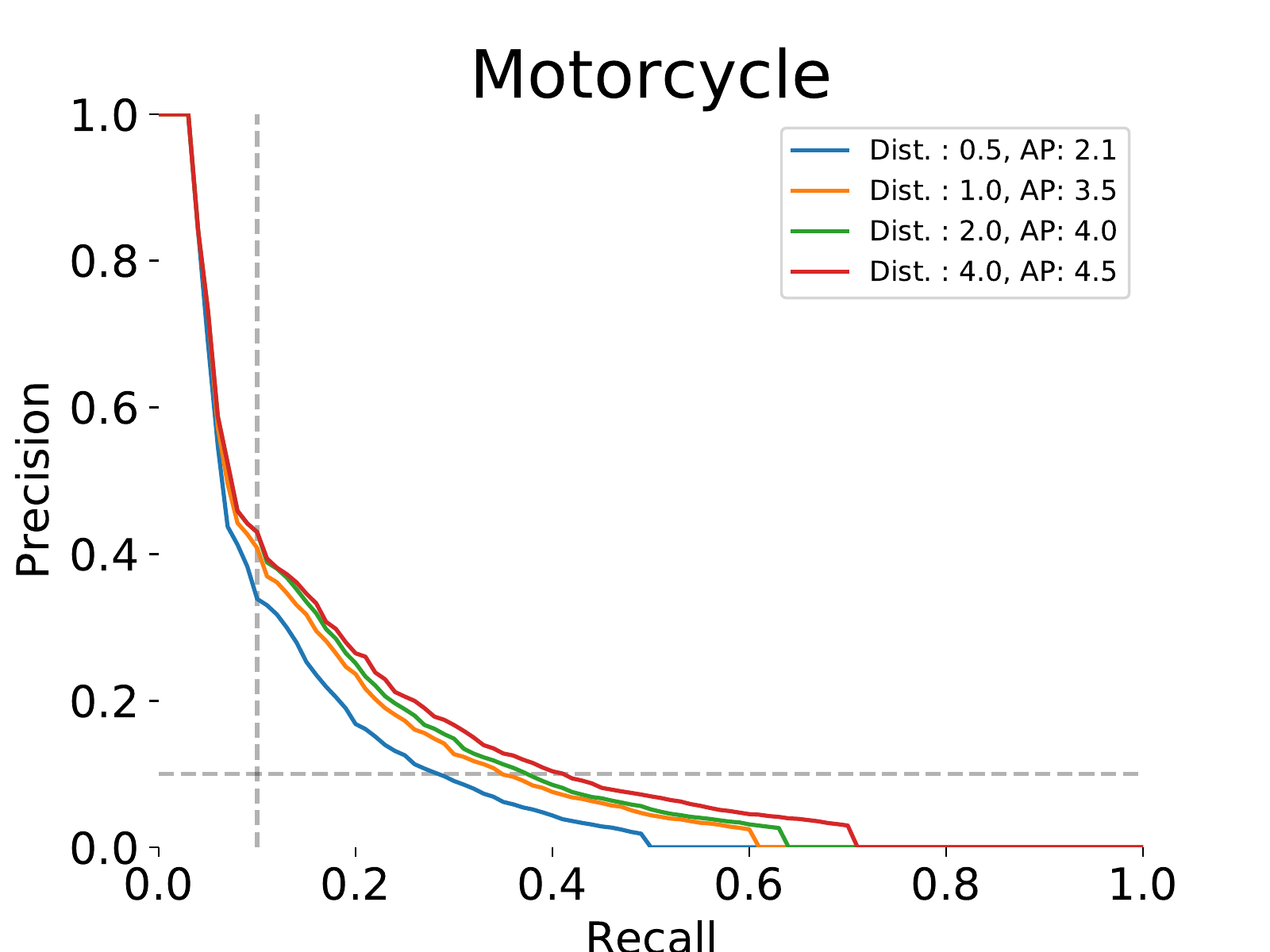}
				\\
				\includegraphics[width=1\columnwidth]{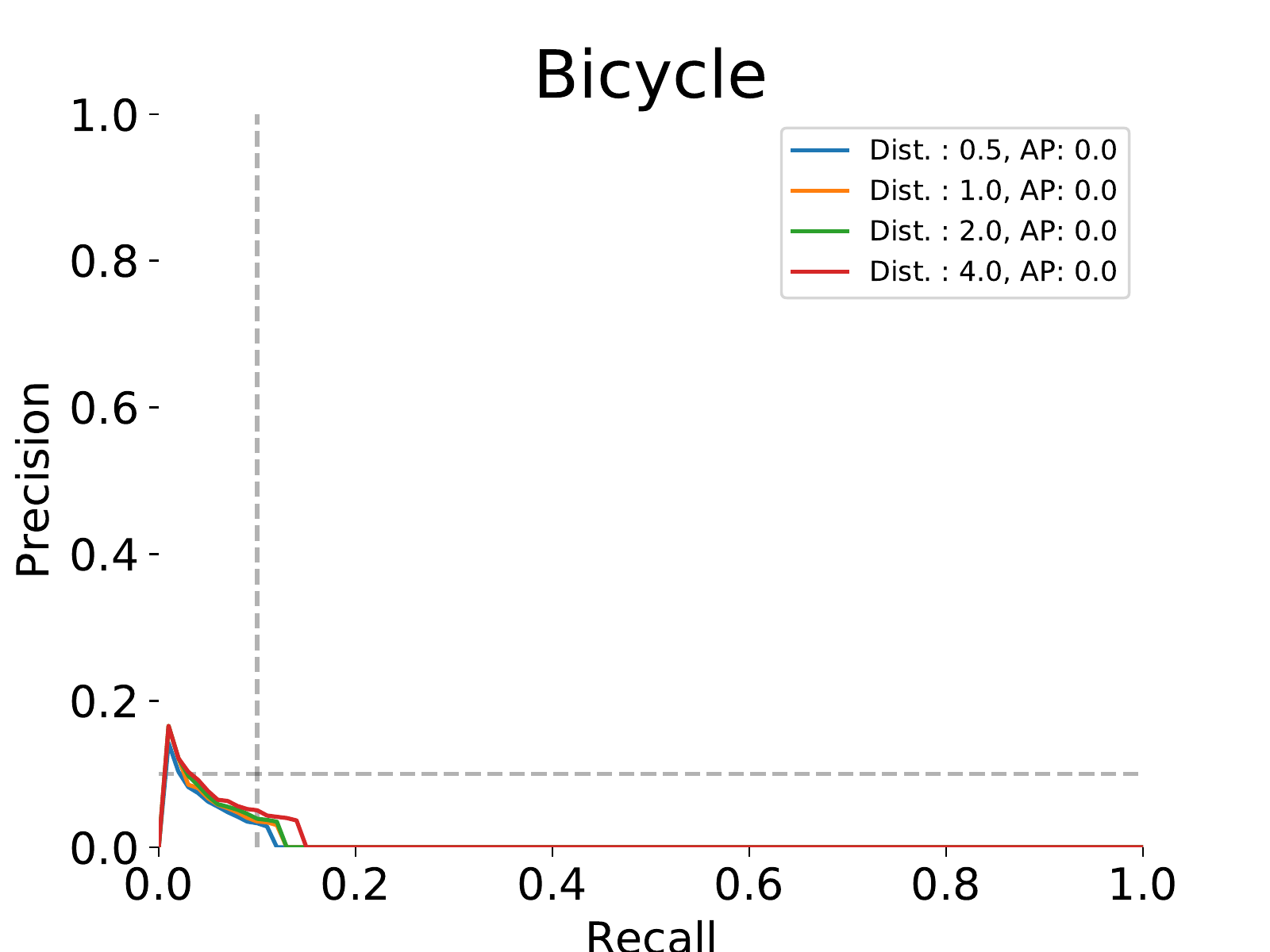}
			\end{minipage}
		}
		\subfigure[t] { 
			\begin{minipage}{0.186\textwidth}
				\centering
				\includegraphics[width=1\columnwidth]{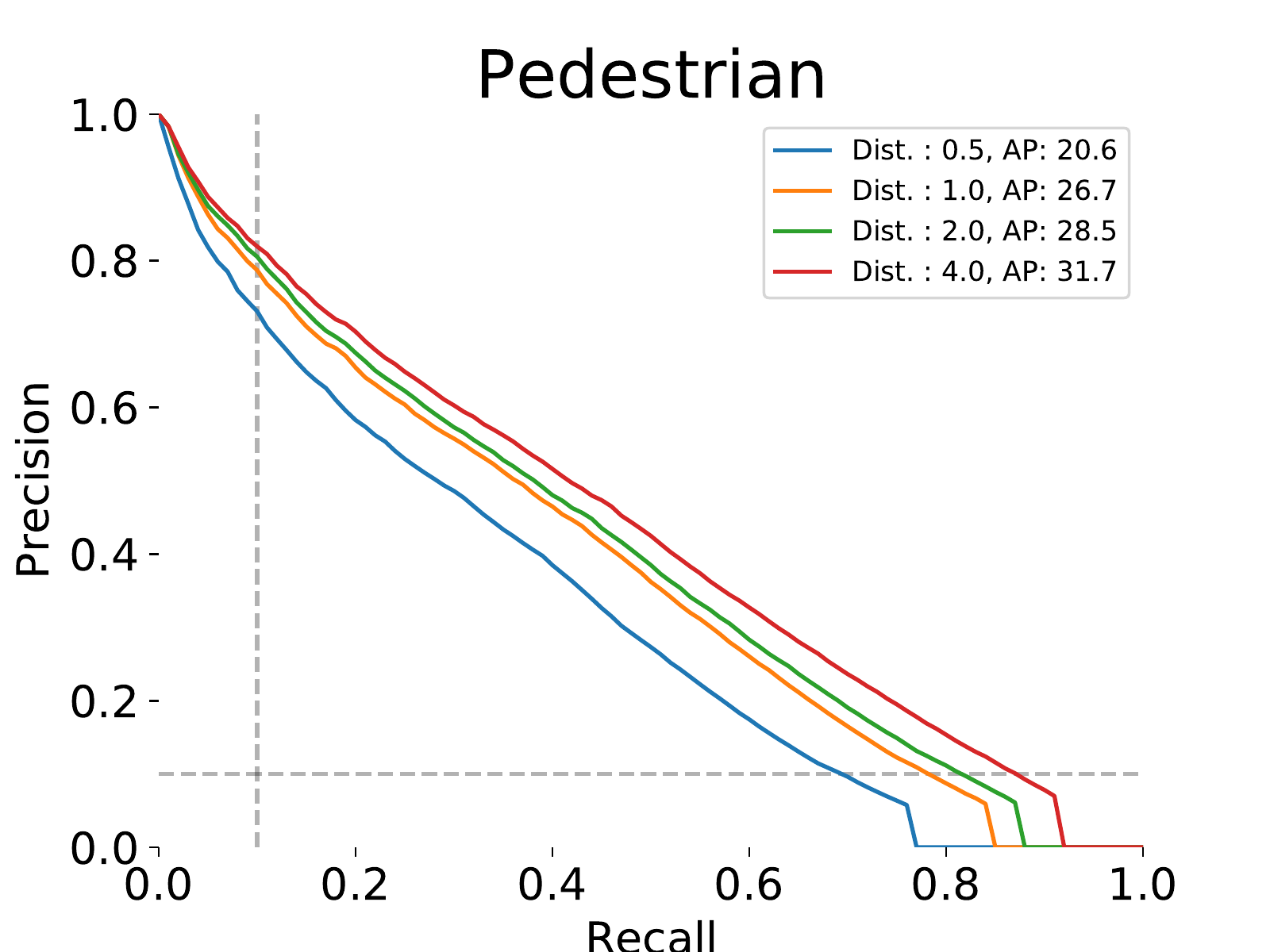}
				\\
				\includegraphics[width=1\columnwidth]{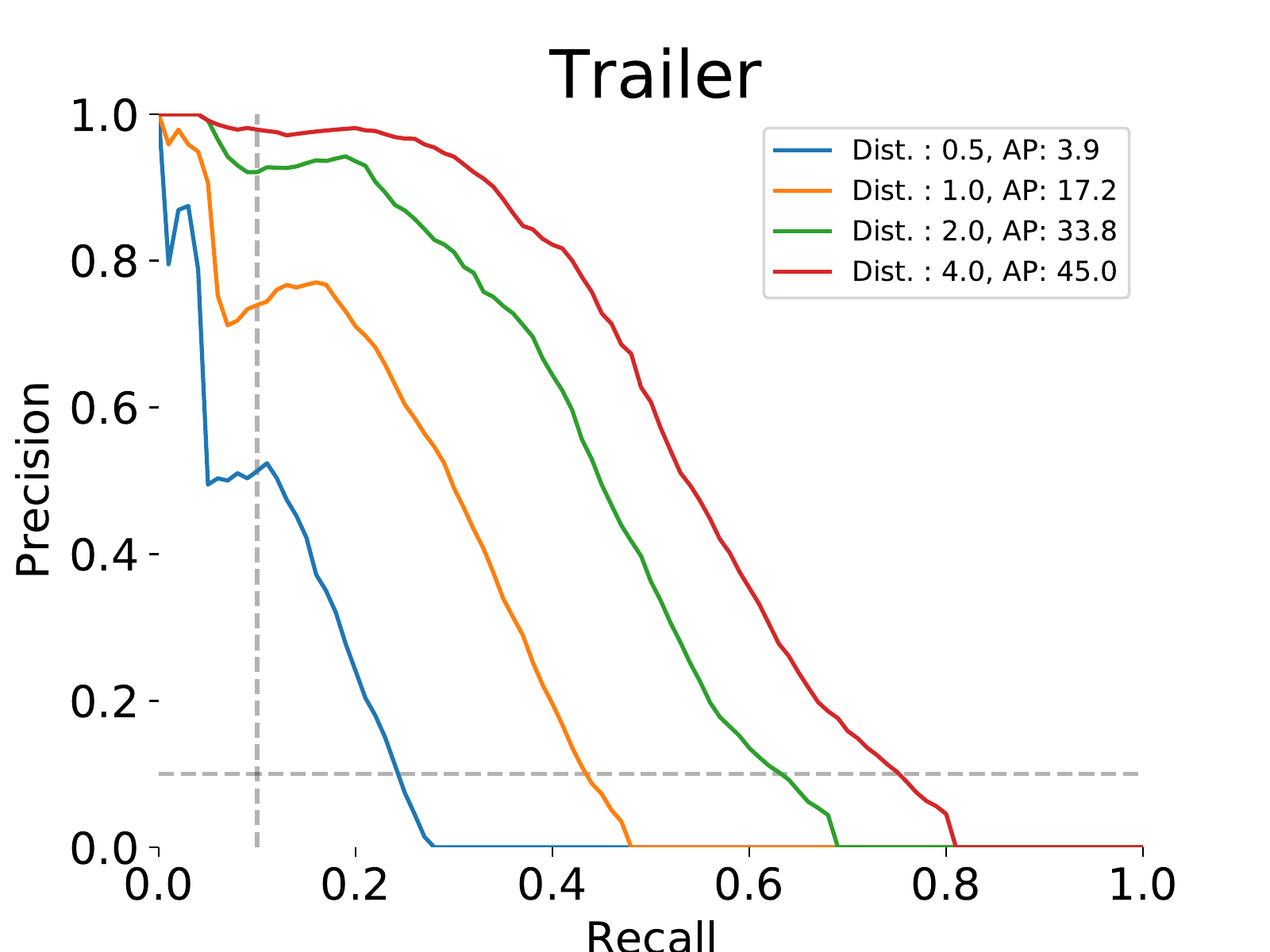}
			\end{minipage}
		}
		\subfigure[t] { 
			\begin{minipage}{0.186\textwidth}
				\centering
				\includegraphics[width=1\columnwidth]{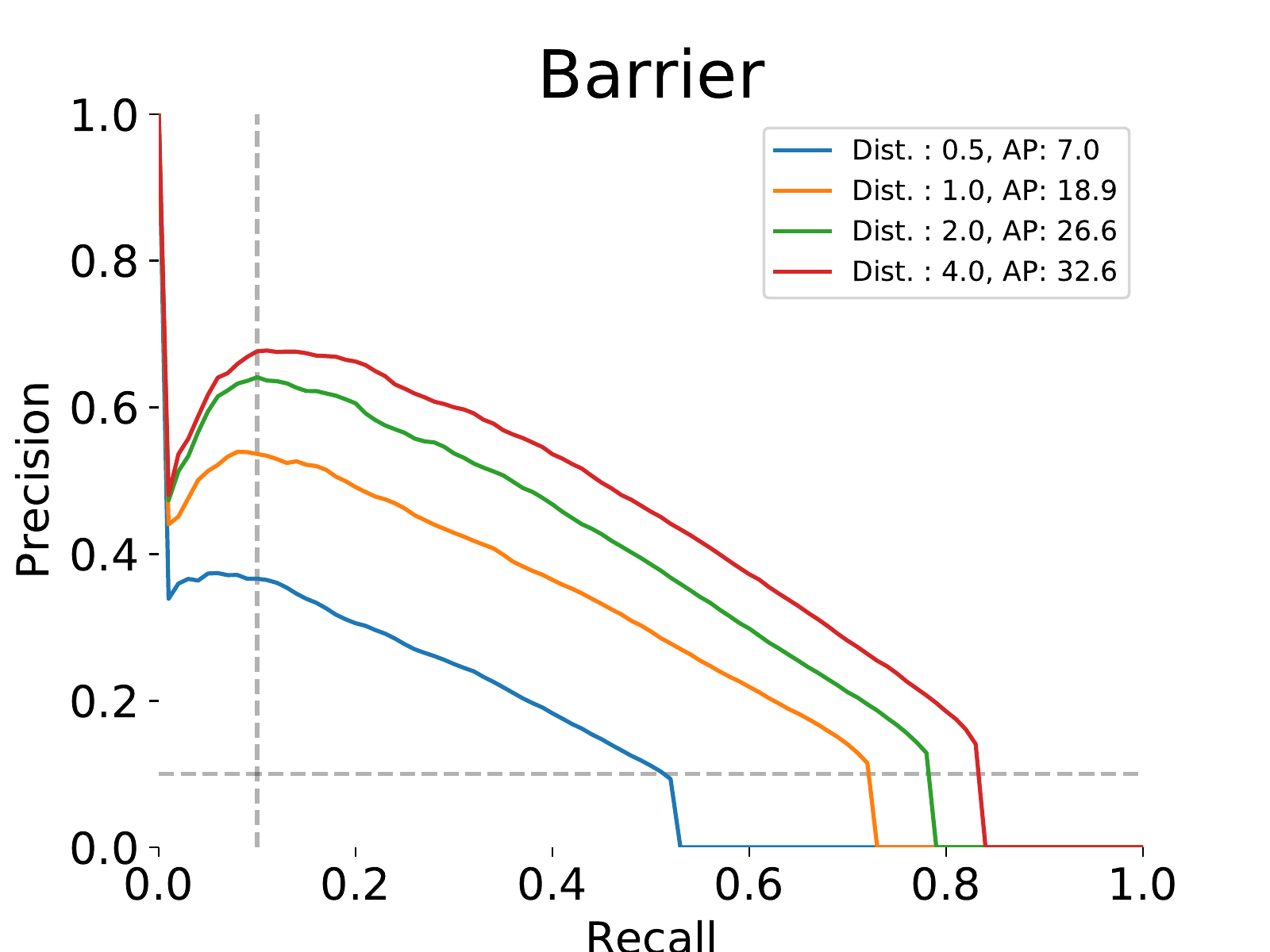}
				\\
				\includegraphics[width=1\columnwidth]{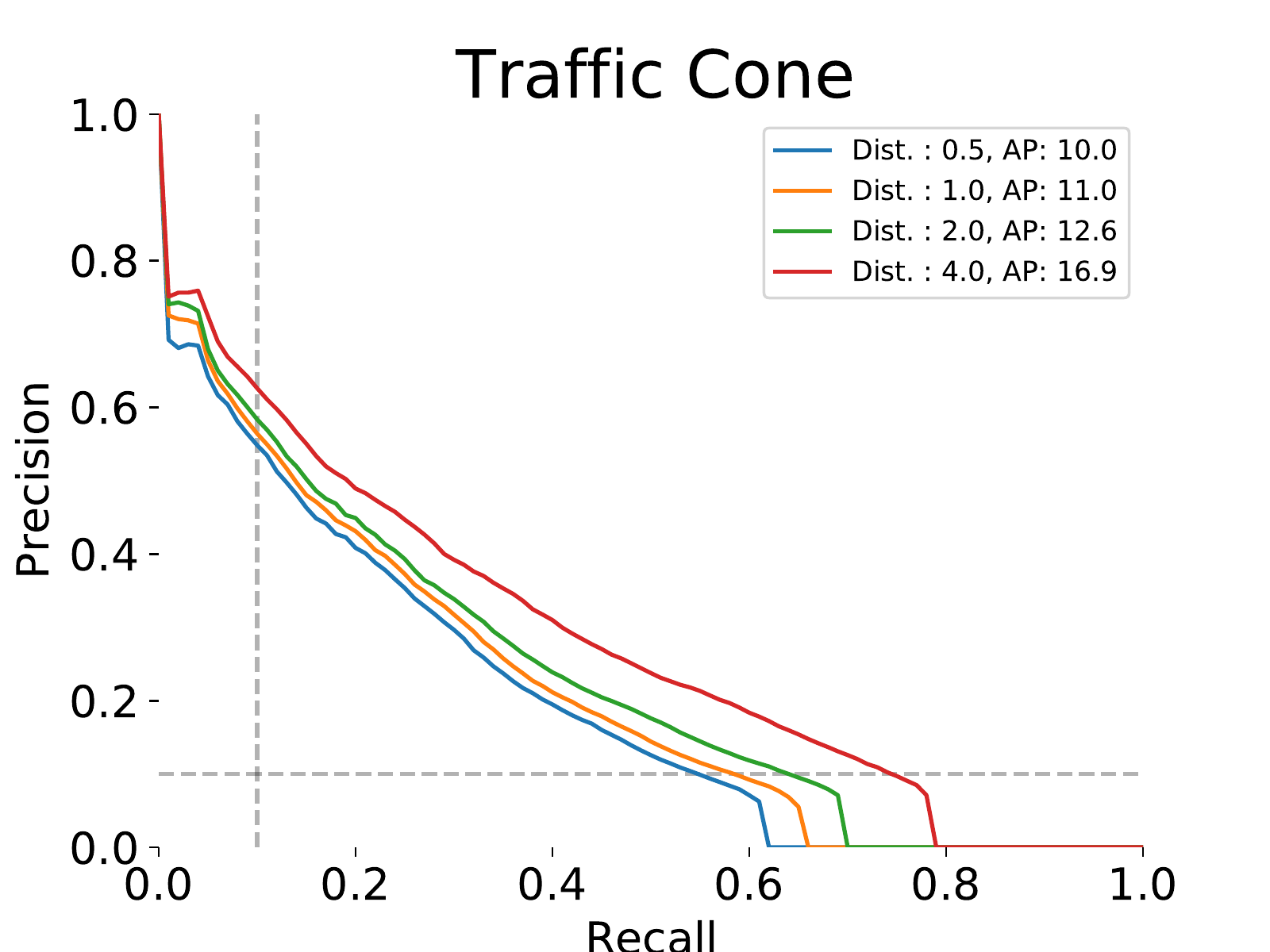}
			\end{minipage}
		}		
	\end{center}
	
	\caption{Per category precision-recall plot of attacking the \textbf{translation} only (black box) on the nuScenes validation set [2].}
	\label{fig:black_GNSS}
	%\label{fig:onecol}
\end{figure*}

\begin{figure*}[t]
	\begin{center}

		\subfigure[t] { 
			\begin{minipage}{0.186\textwidth}
				\centering
				\includegraphics[width=1\columnwidth]{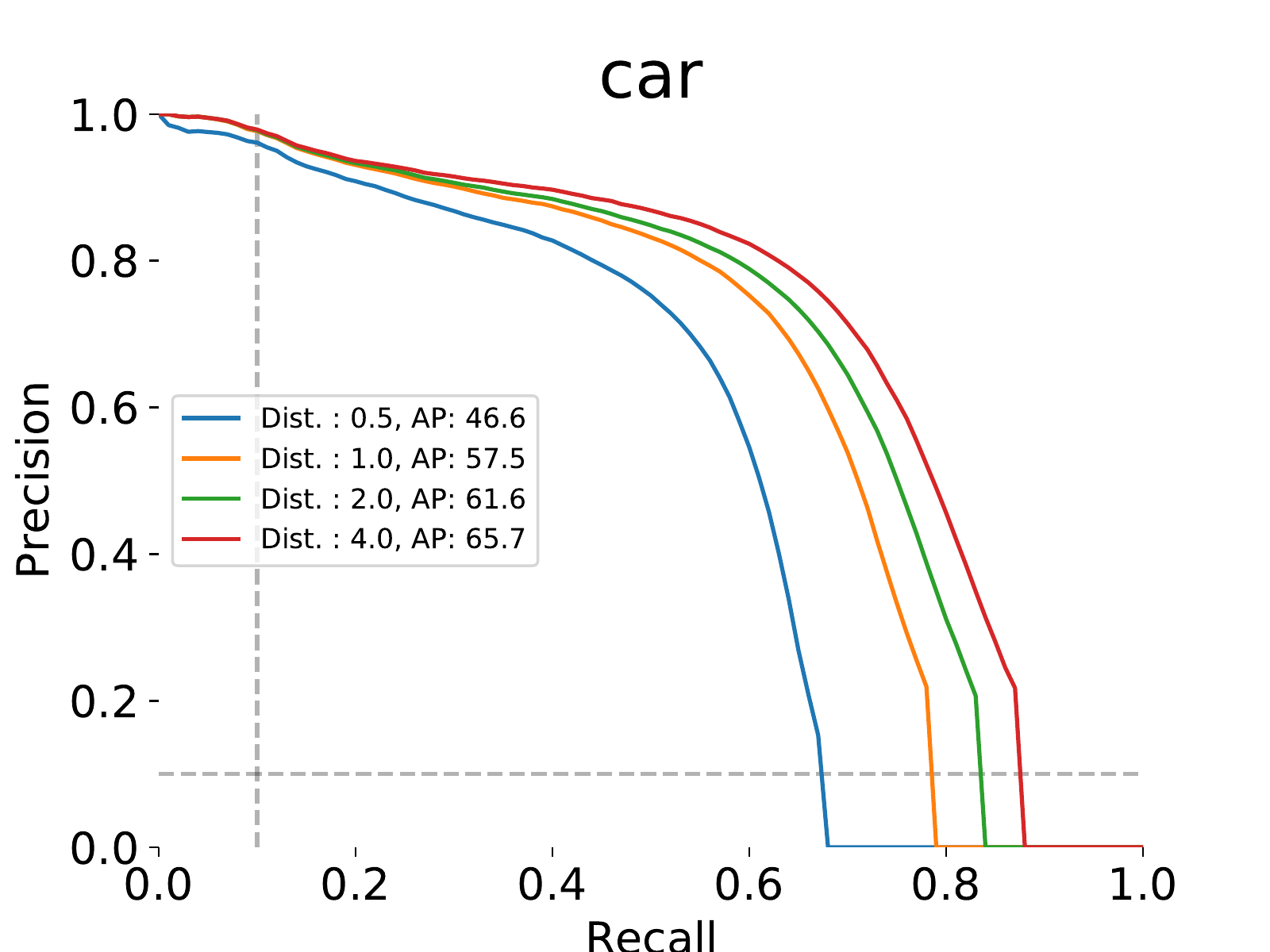}
				\\
				\includegraphics[width=1\columnwidth]{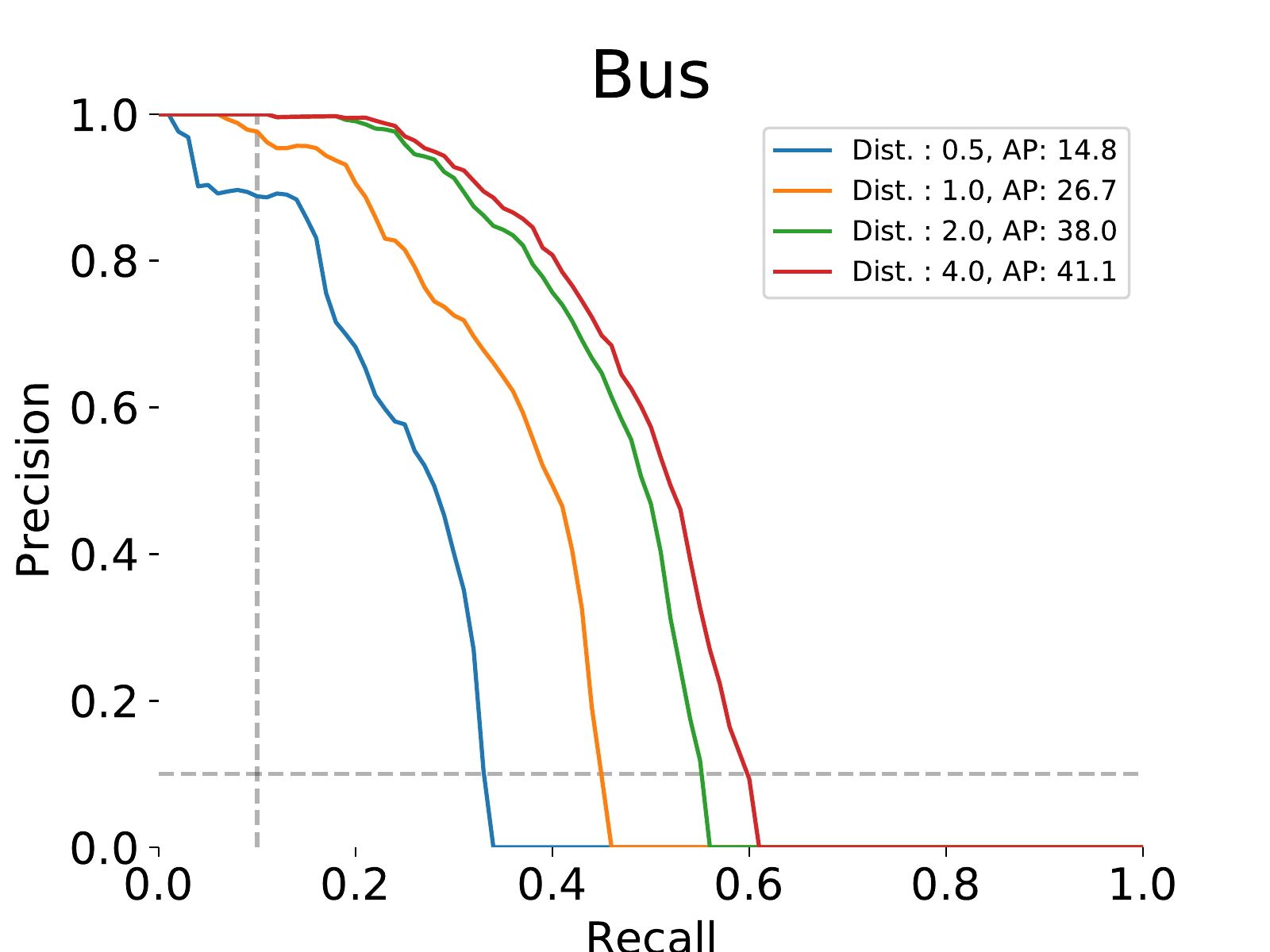}
			\end{minipage}
		}
		\subfigure[t] { 
			\begin{minipage}{0.186\textwidth}
				\centering
				\includegraphics[width=1\columnwidth]{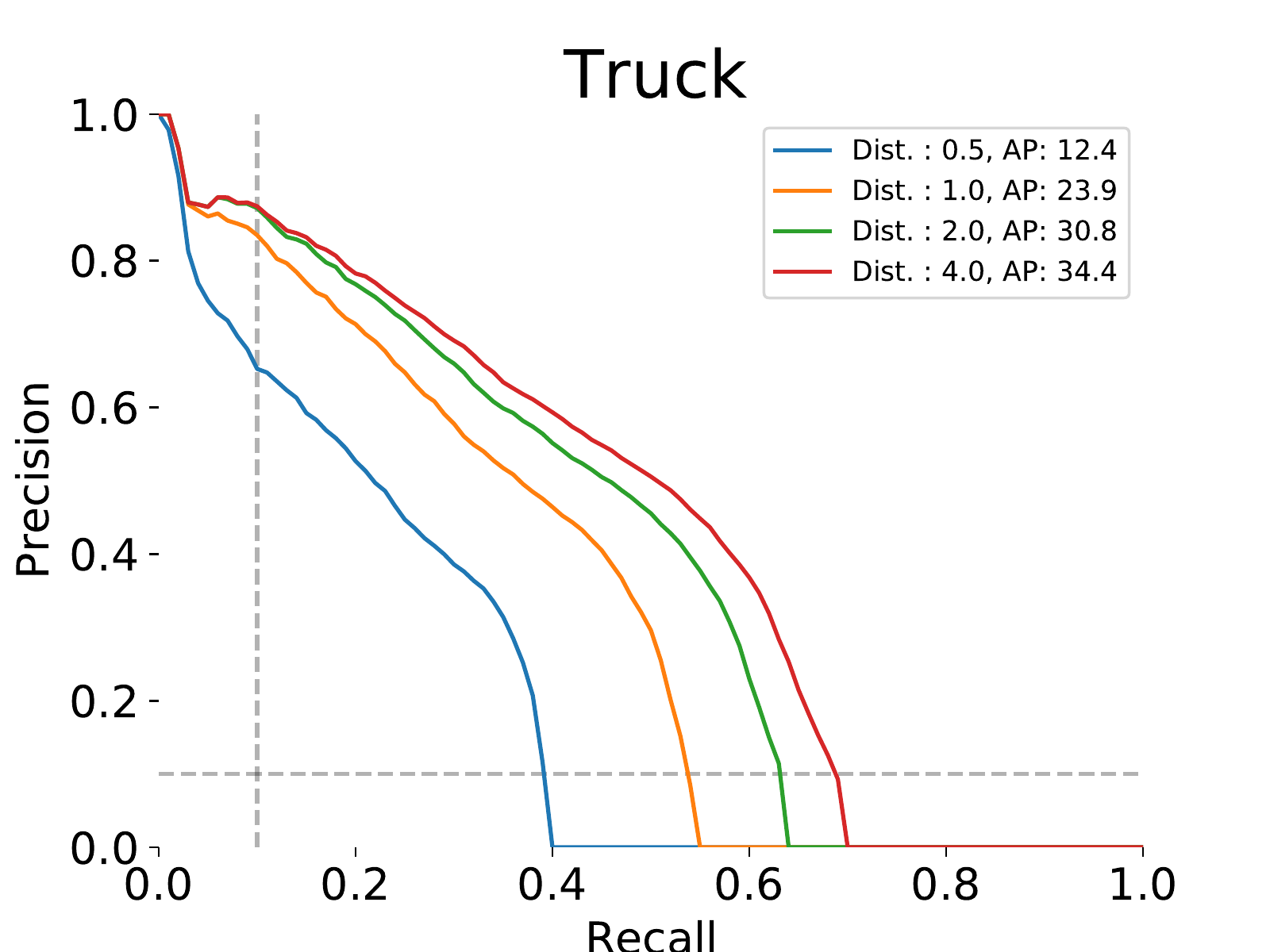}
				\\
				\includegraphics[width=1\columnwidth]{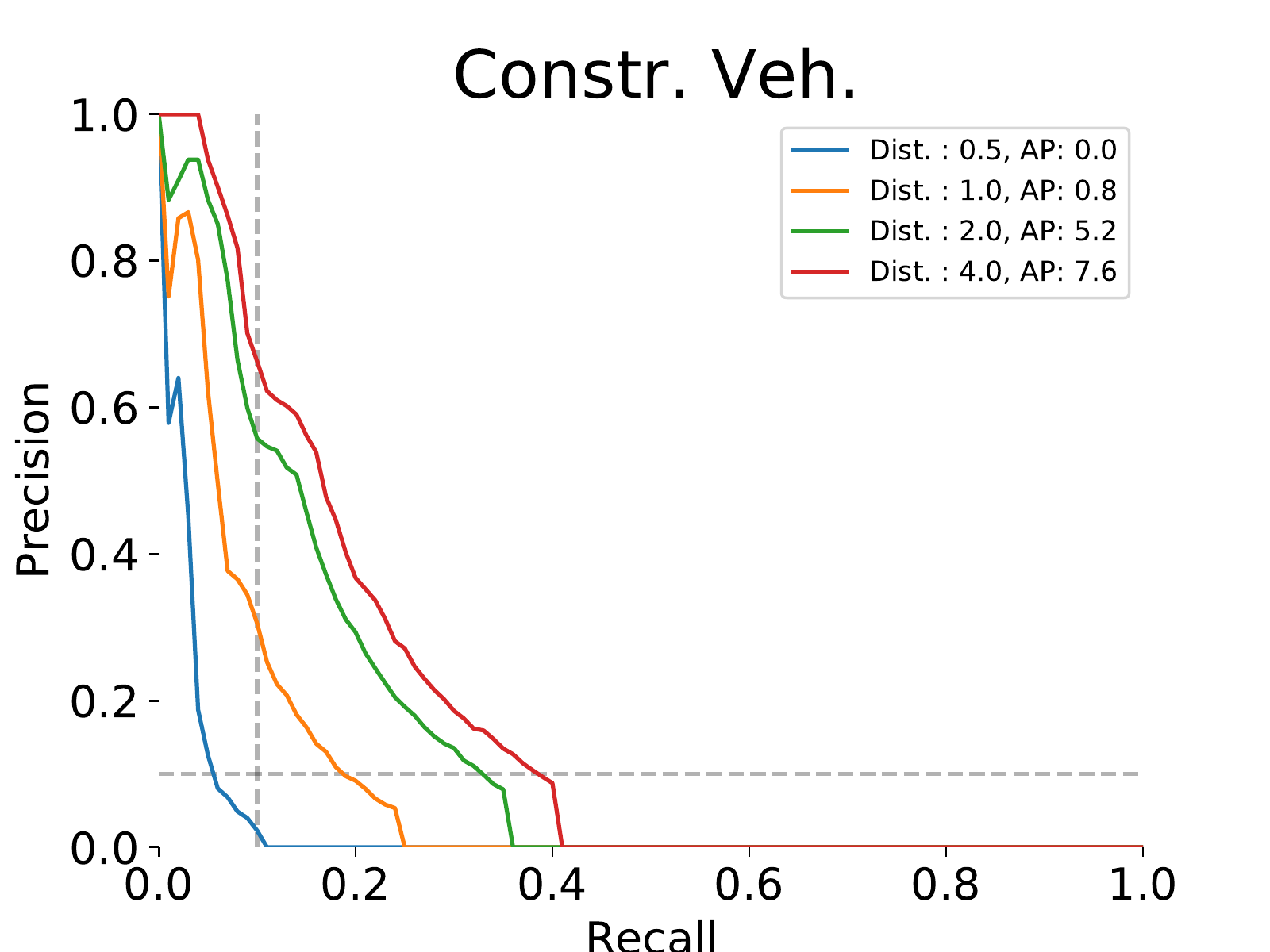}
			\end{minipage}
		}
		\subfigure[t] { 
			\begin{minipage}{0.186\textwidth}
				\centering
				\includegraphics[width=1\columnwidth]{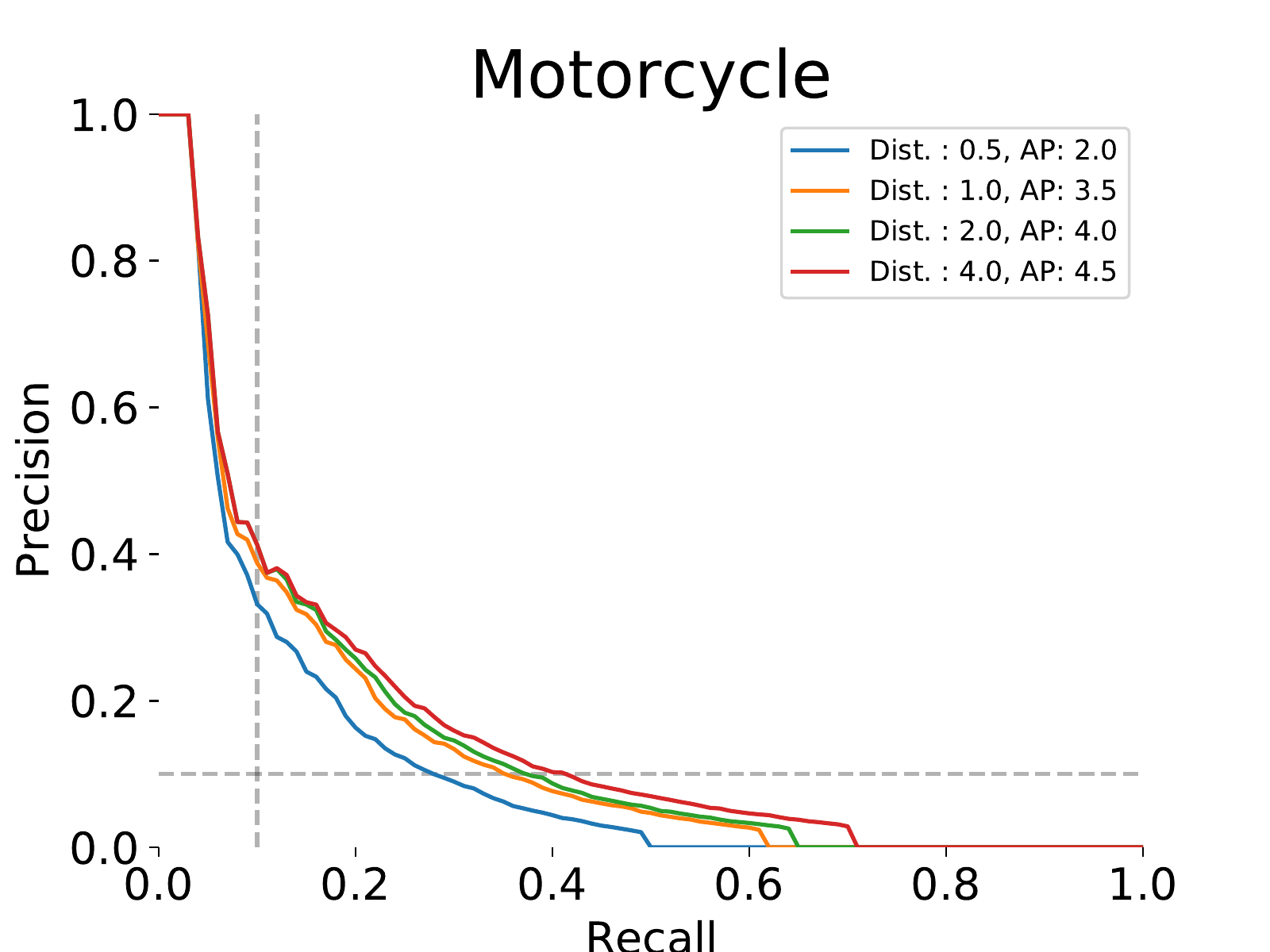}
				\\
				\includegraphics[width=1\columnwidth]{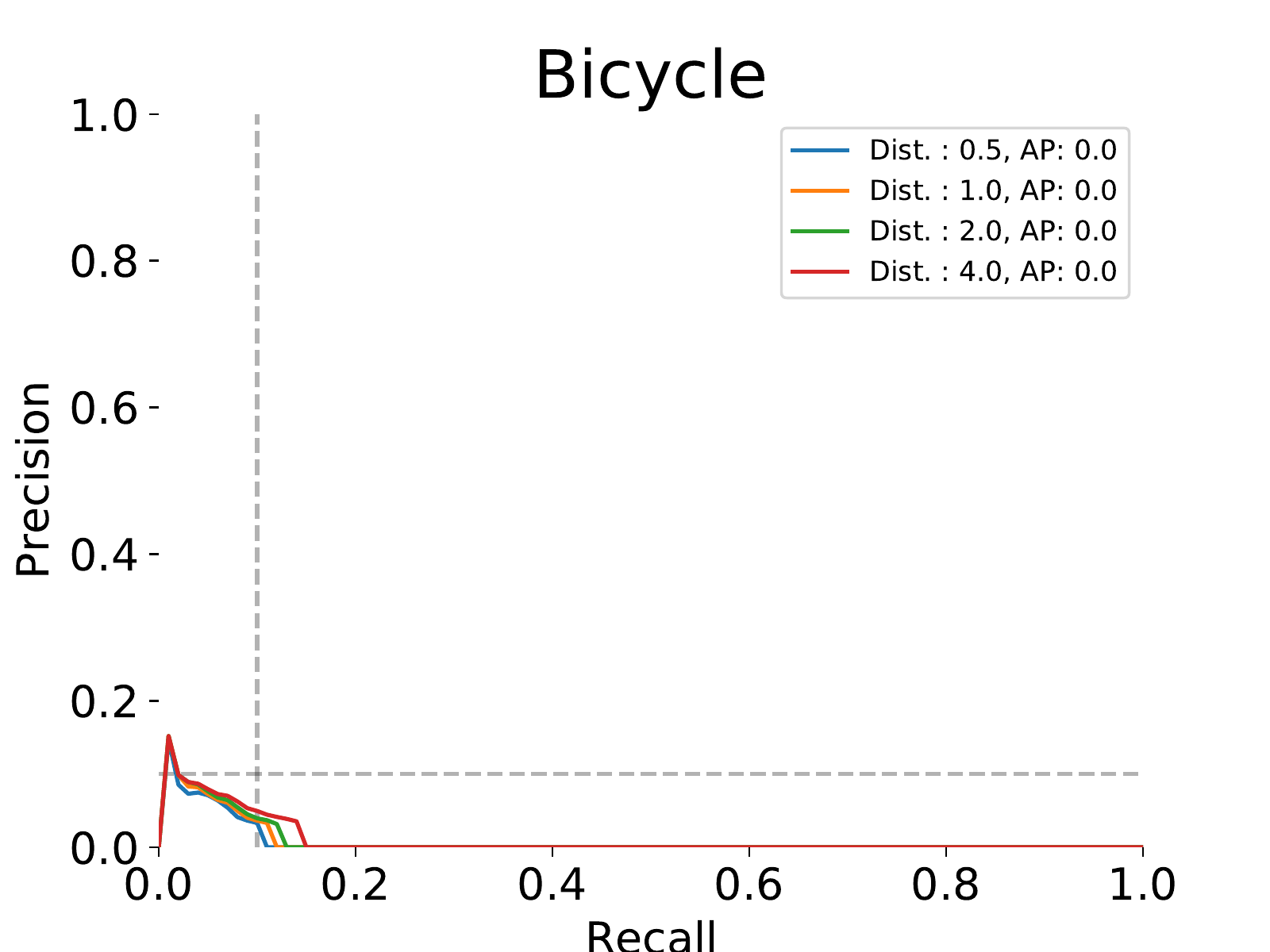}
			\end{minipage}
		}
		\subfigure[t] { 
			\begin{minipage}{0.186\textwidth}
				\centering
				\includegraphics[width=1\columnwidth]{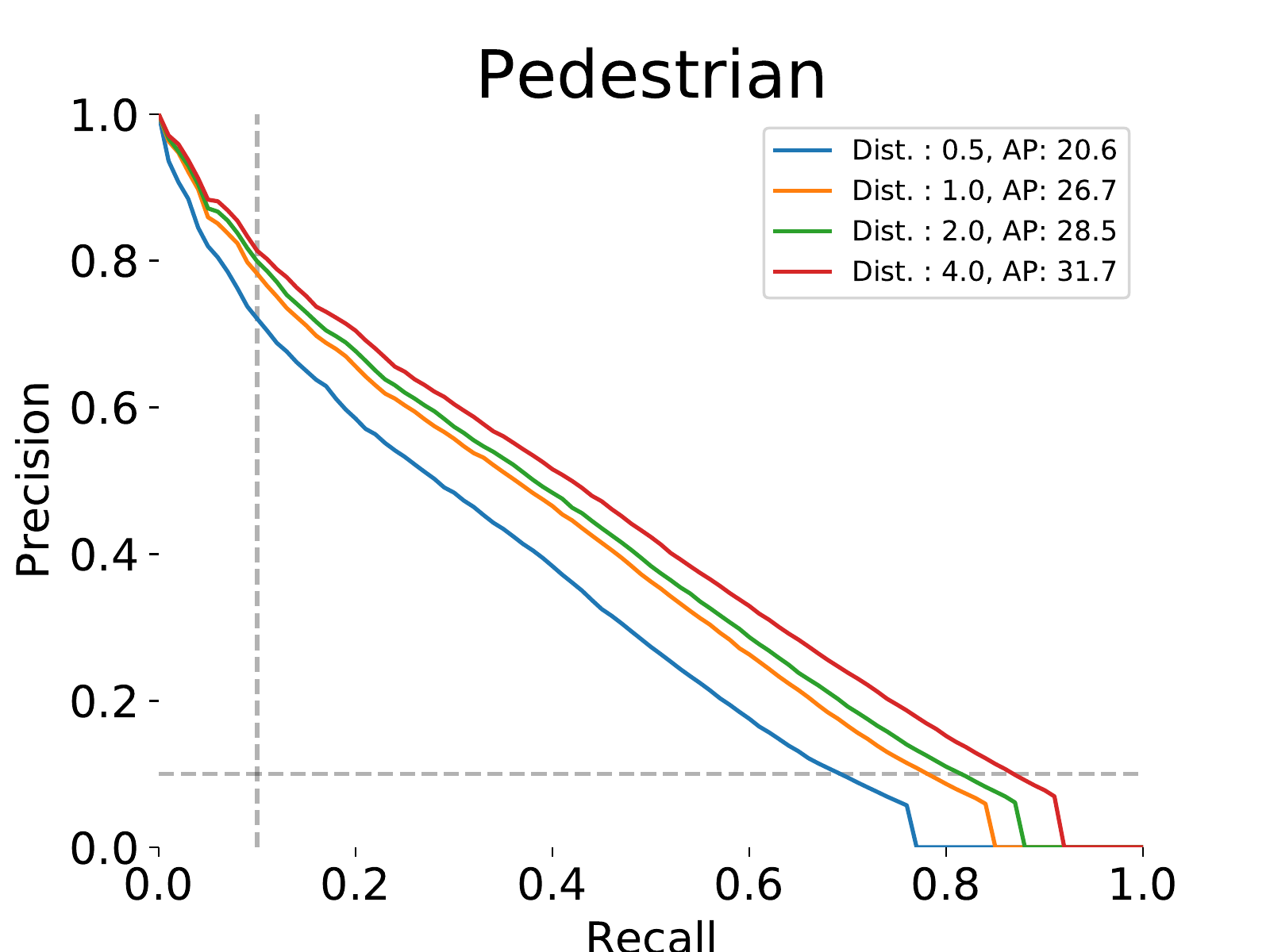}
				\\
				\includegraphics[width=1\columnwidth]{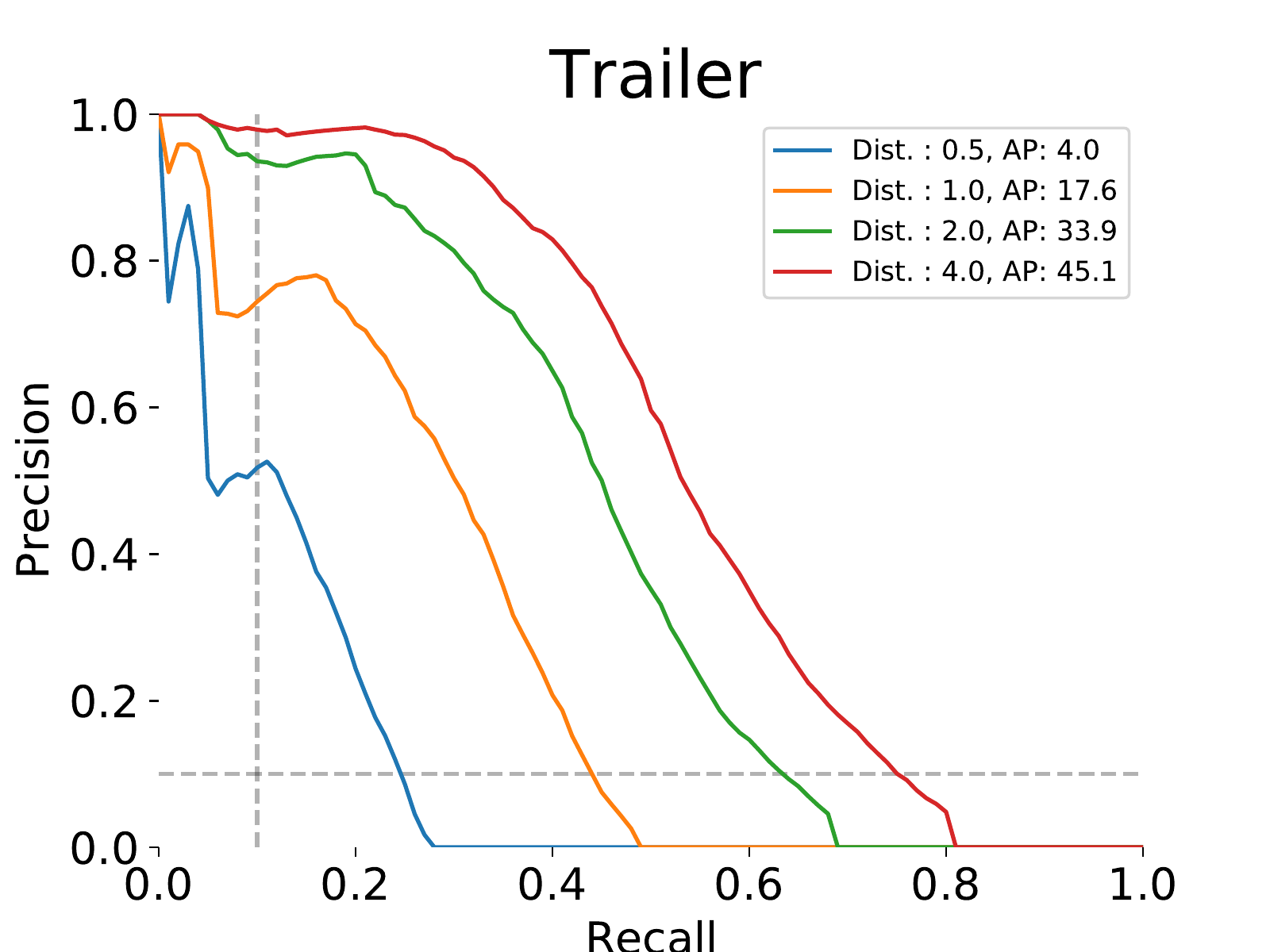}
			\end{minipage}
		}
		\subfigure[t] { 
			\begin{minipage}{0.186\textwidth}
				\centering
				\includegraphics[width=1\columnwidth]{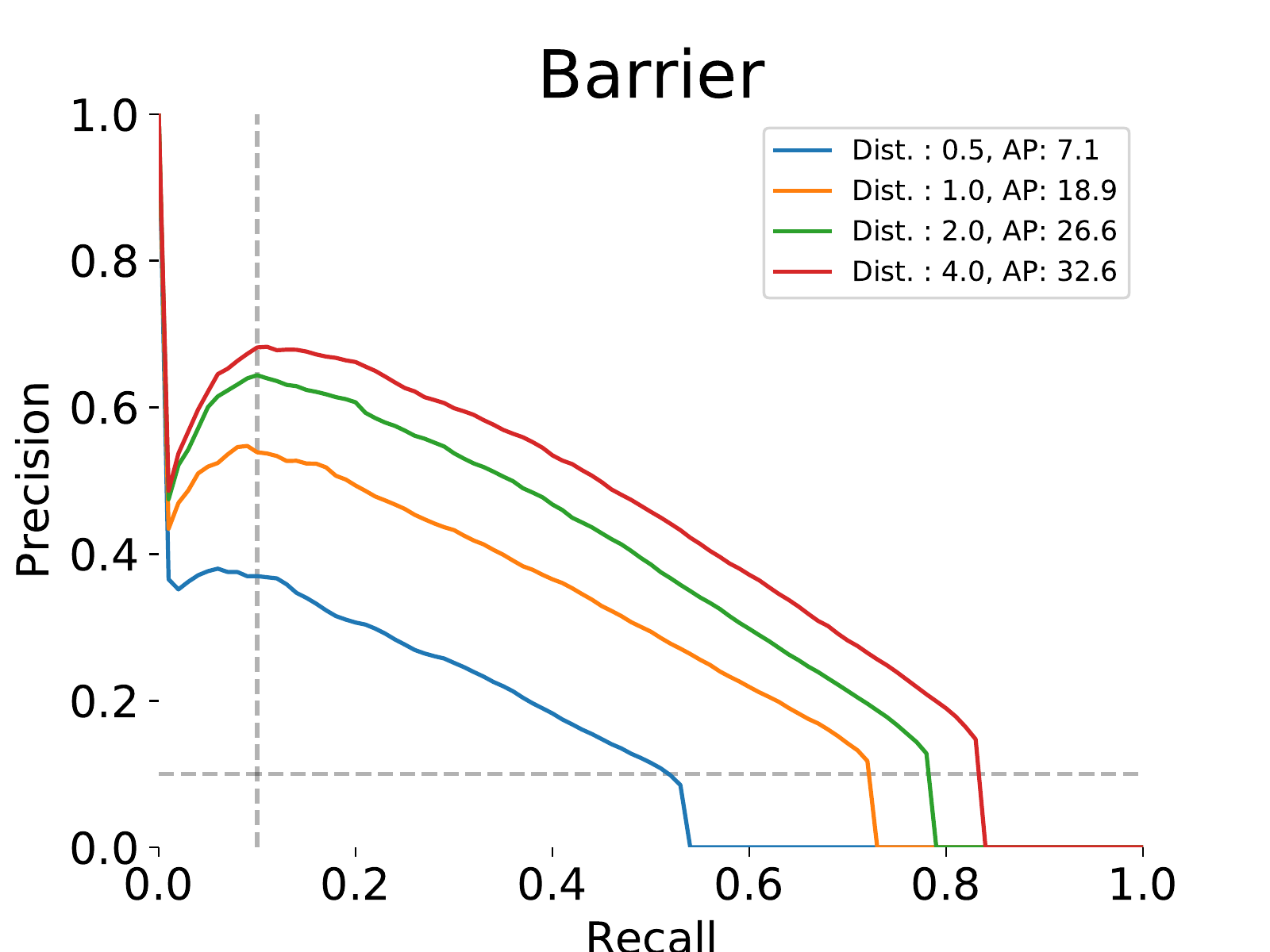}
				\\
				\includegraphics[width=1\columnwidth]{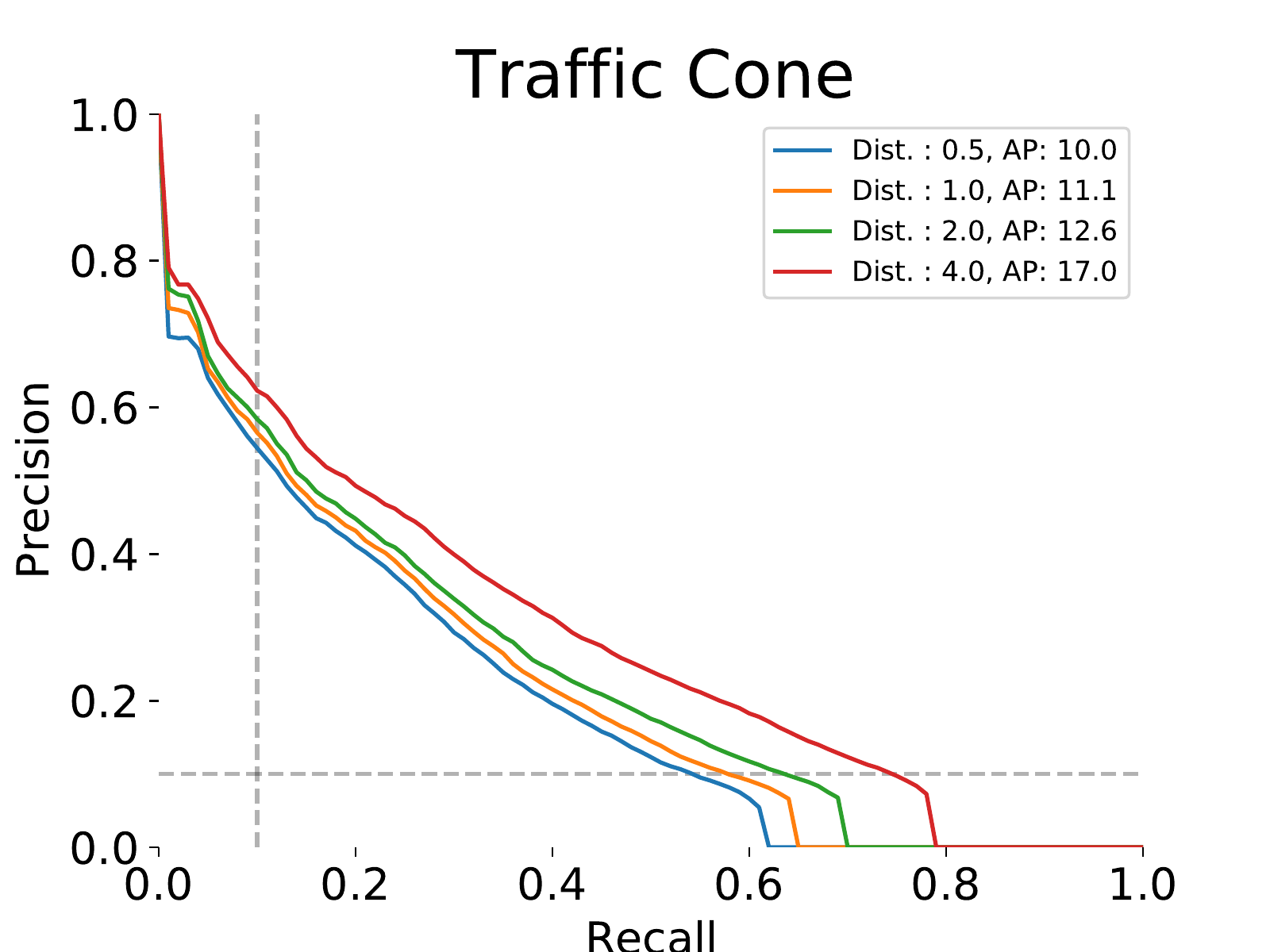}
			\end{minipage}
		}		
	\end{center}
	
	\caption{Per category precision-recall plot of attacking the \textbf{polynomial coefficients} (black box) on the nuScenes validation set [2].}
	\label{fig:black_poly}
	%\label{fig:onecol}
\end{figure*}

\begin{figure*}[t]
	\begin{center}

		\subfigure[t] { 
			\begin{minipage}{0.186\textwidth}
				\centering
				\includegraphics[width=1\columnwidth]{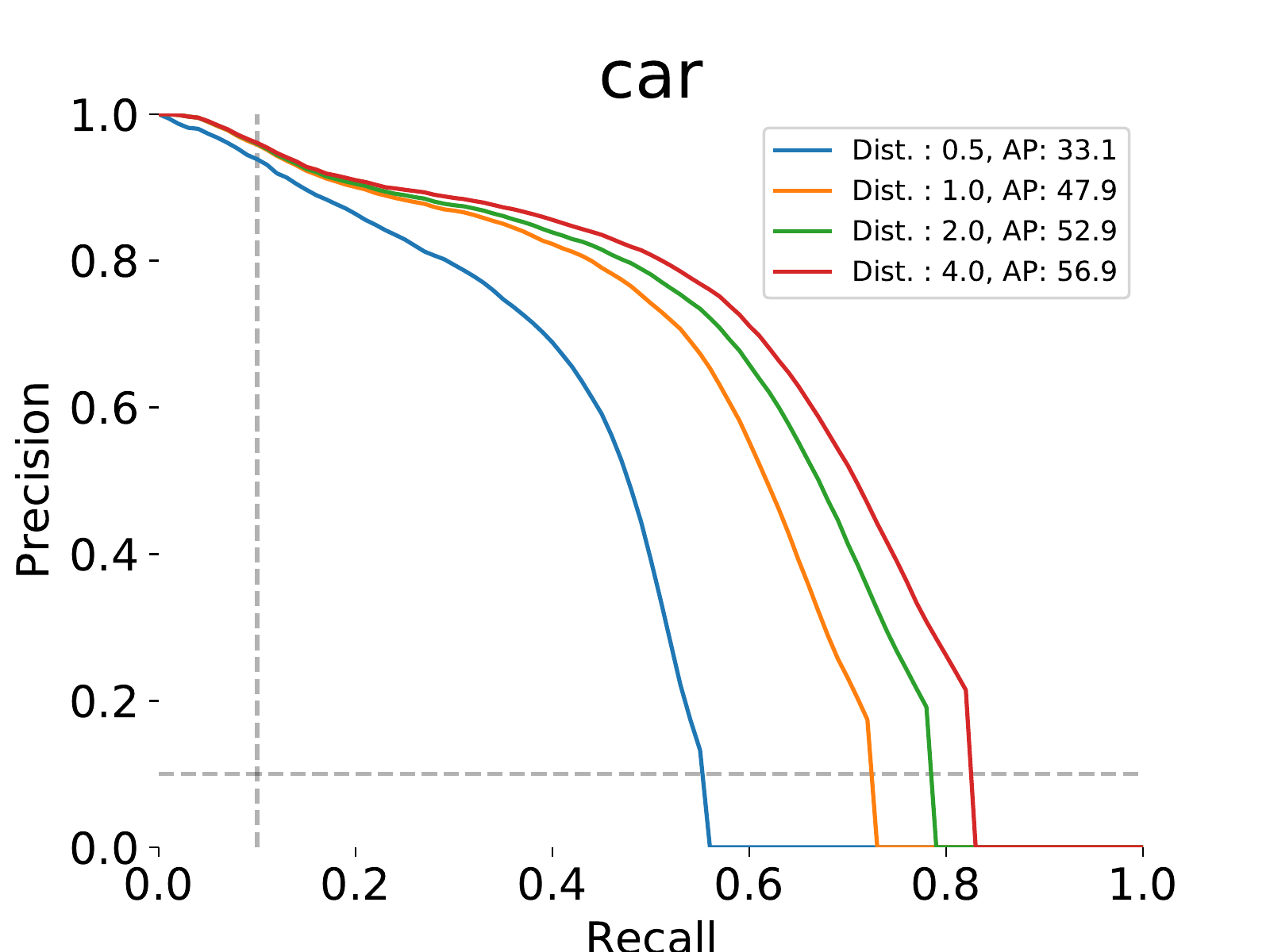}
				\\
				\includegraphics[width=1\columnwidth]{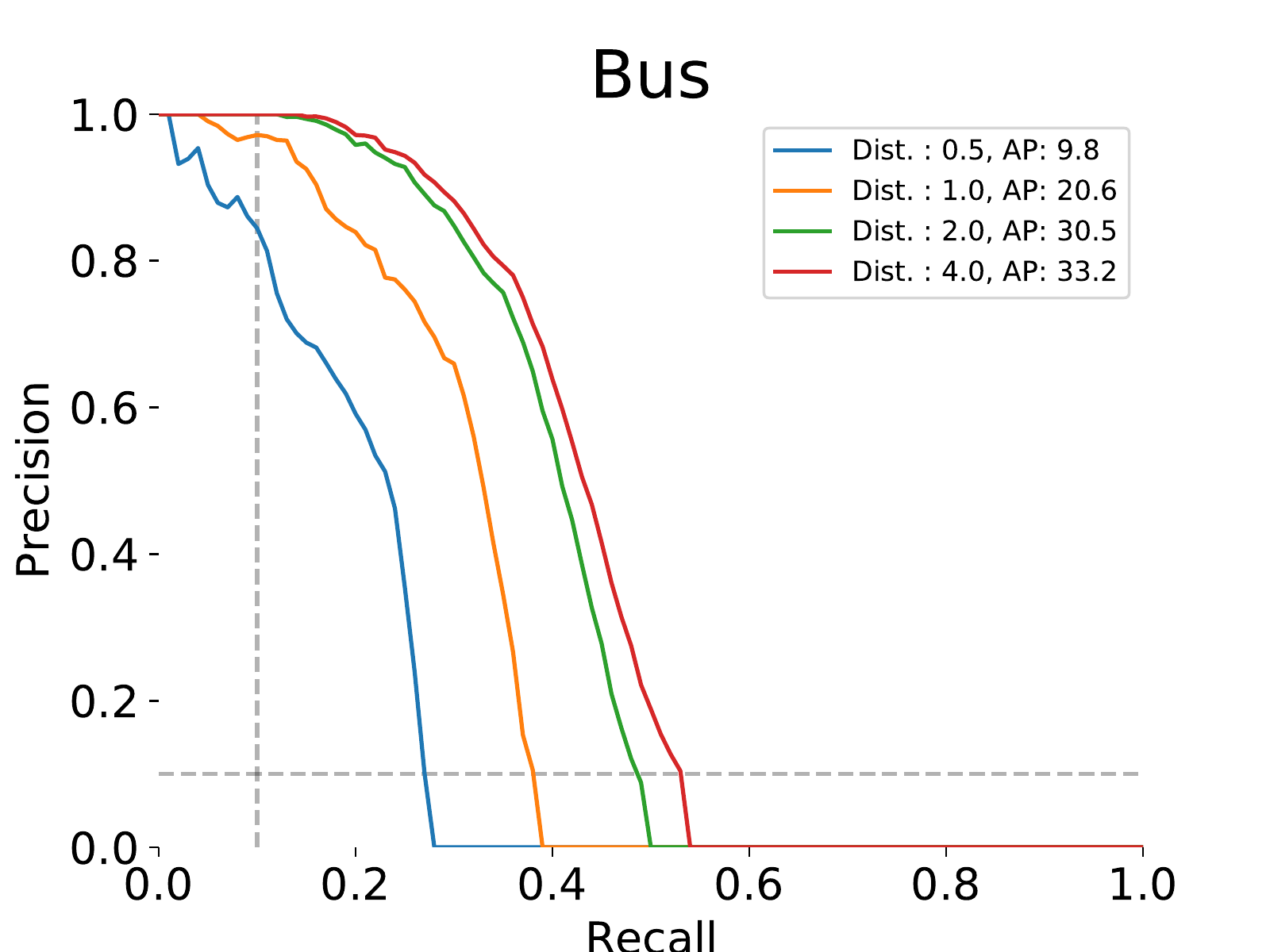}
			\end{minipage}
		}
		\subfigure[t] { 
			\begin{minipage}{0.186\textwidth}
				\centering
				\includegraphics[width=1\columnwidth]{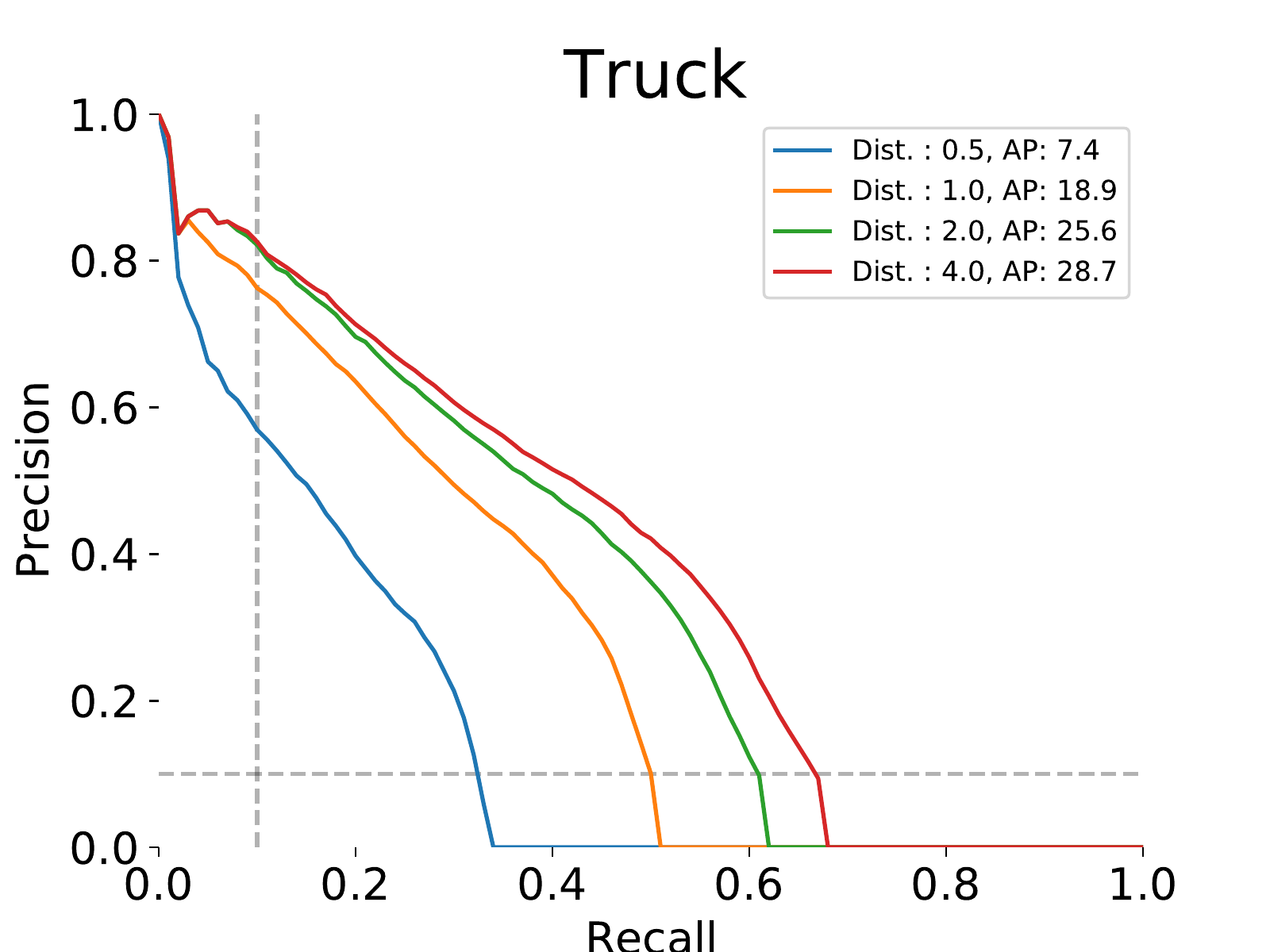}
				\\
				\includegraphics[width=1\columnwidth]{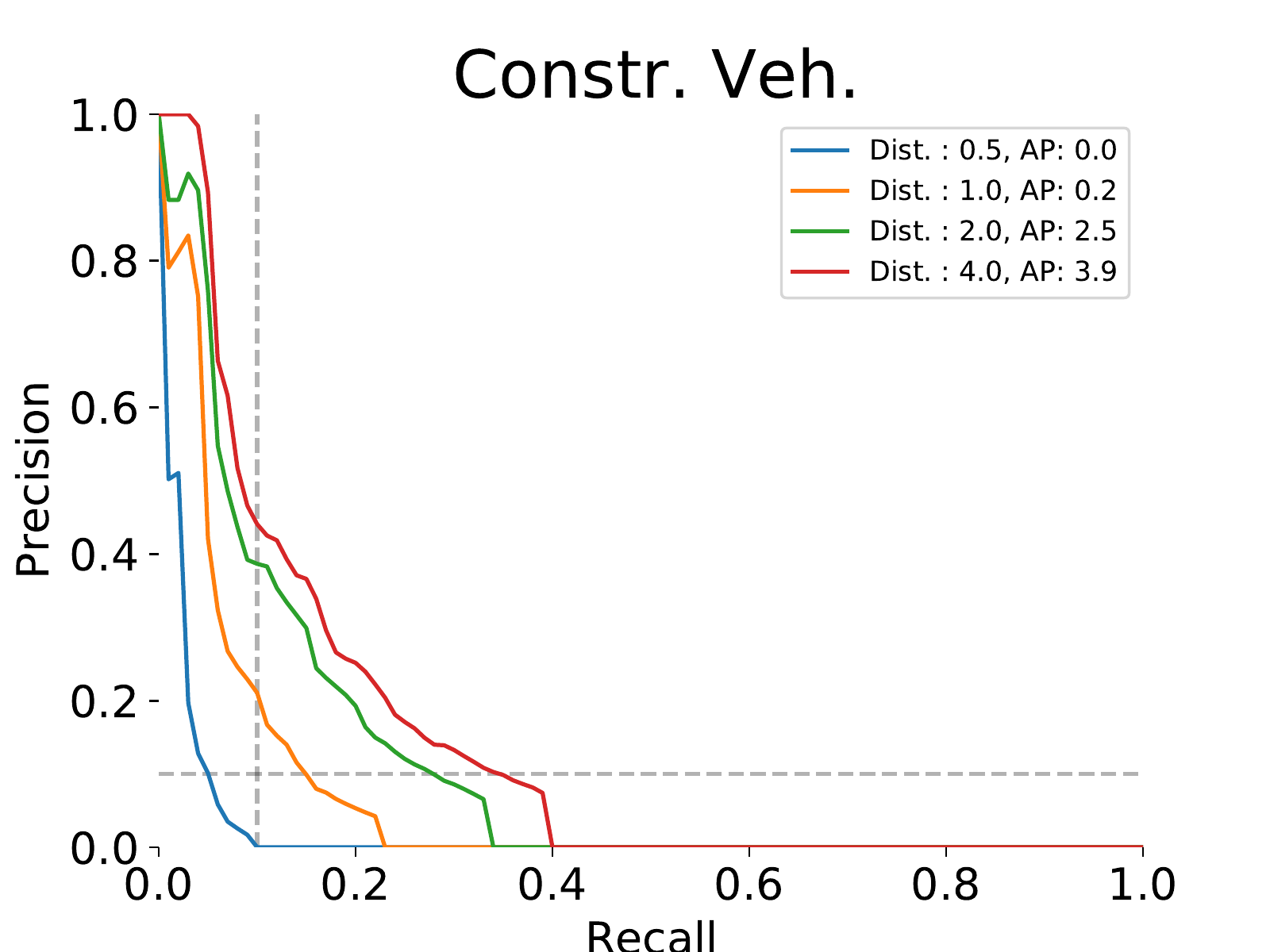}
			\end{minipage}
		}
		\subfigure[t] { 
			\begin{minipage}{0.186\textwidth}
				\centering
				\includegraphics[width=1\columnwidth]{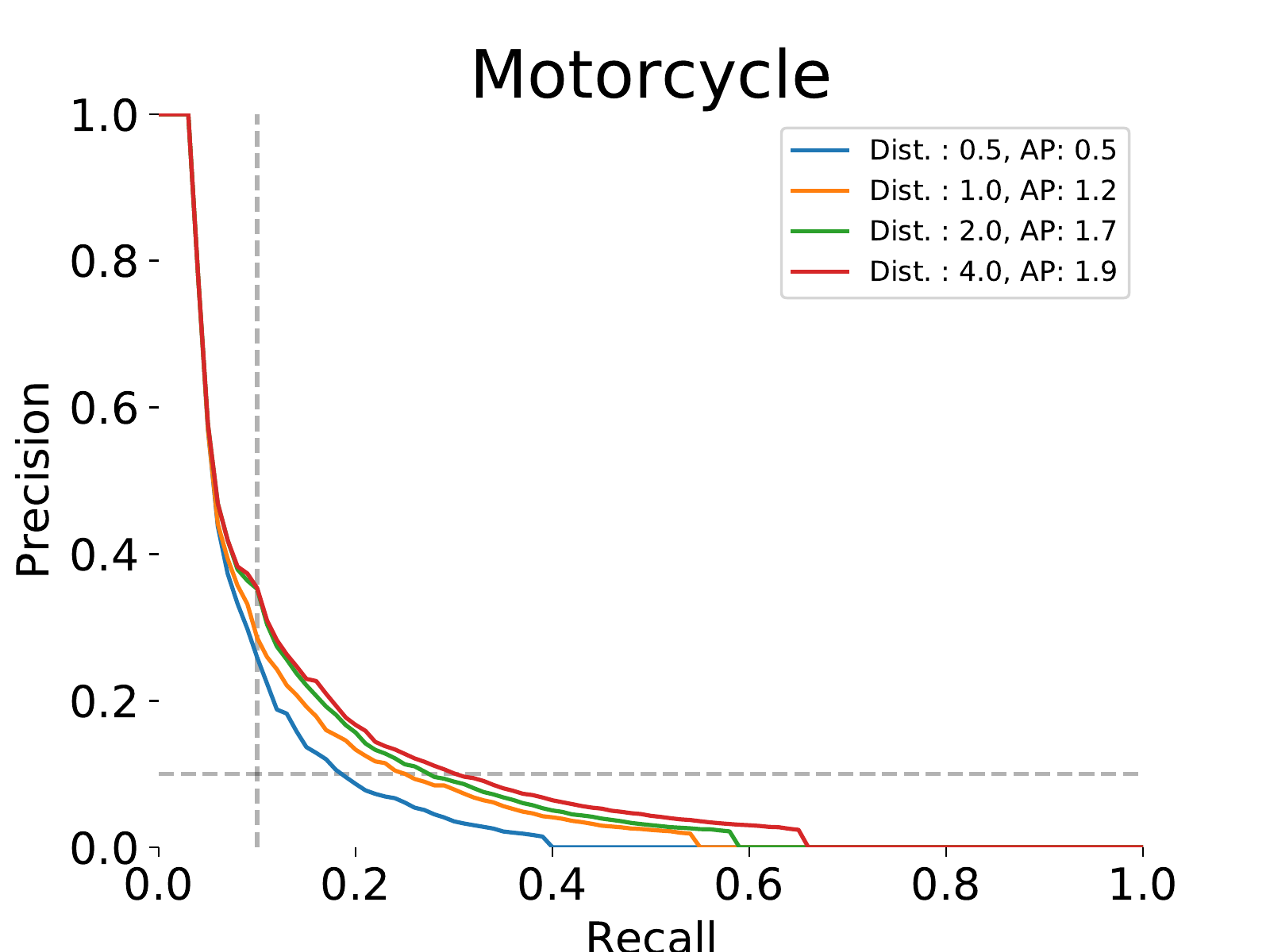}
				\\
				\includegraphics[width=1\columnwidth]{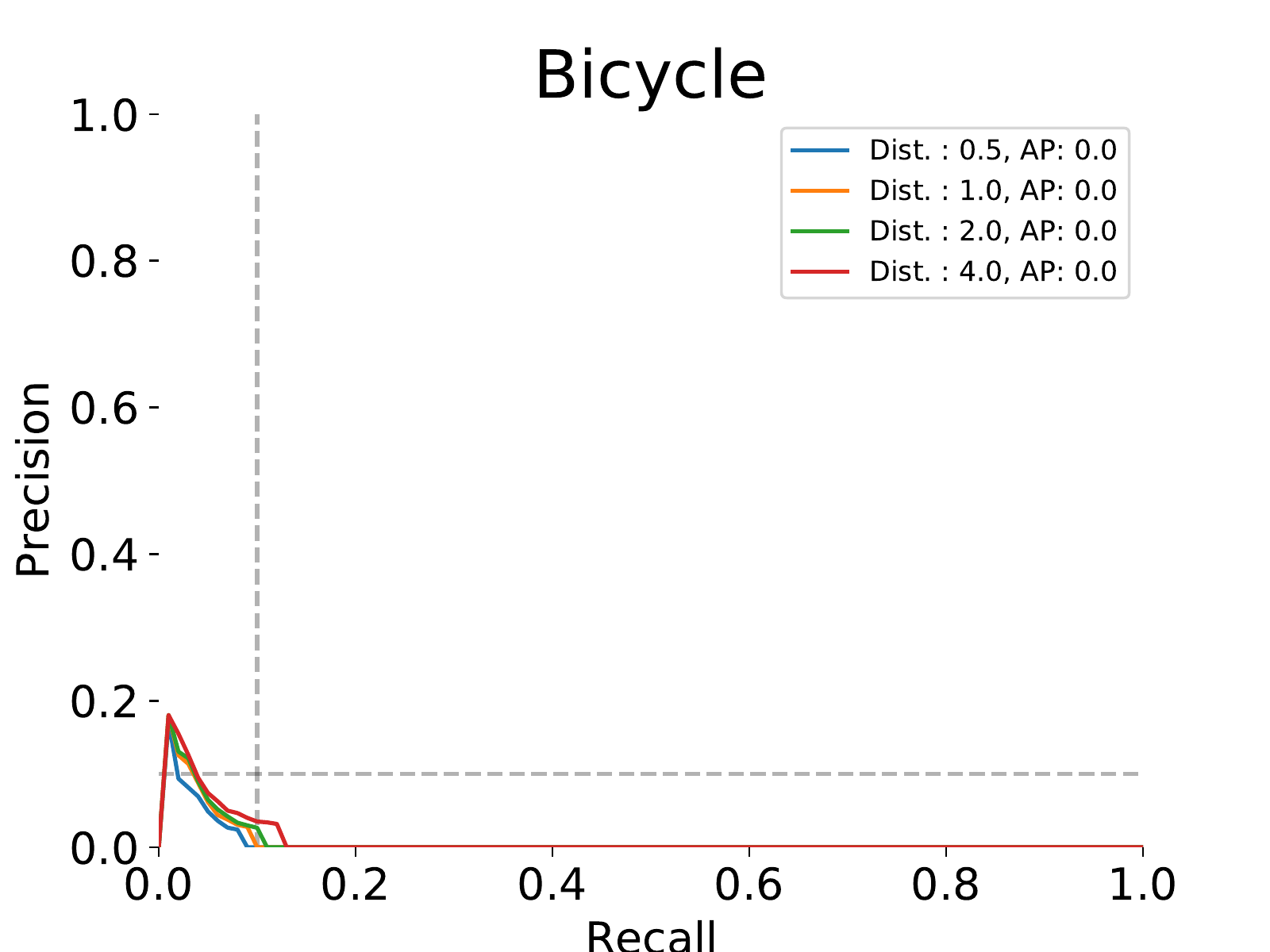}
			\end{minipage}
		}
		\subfigure[t] { 
			\begin{minipage}{0.186\textwidth}
				\centering
				\includegraphics[width=1\columnwidth]{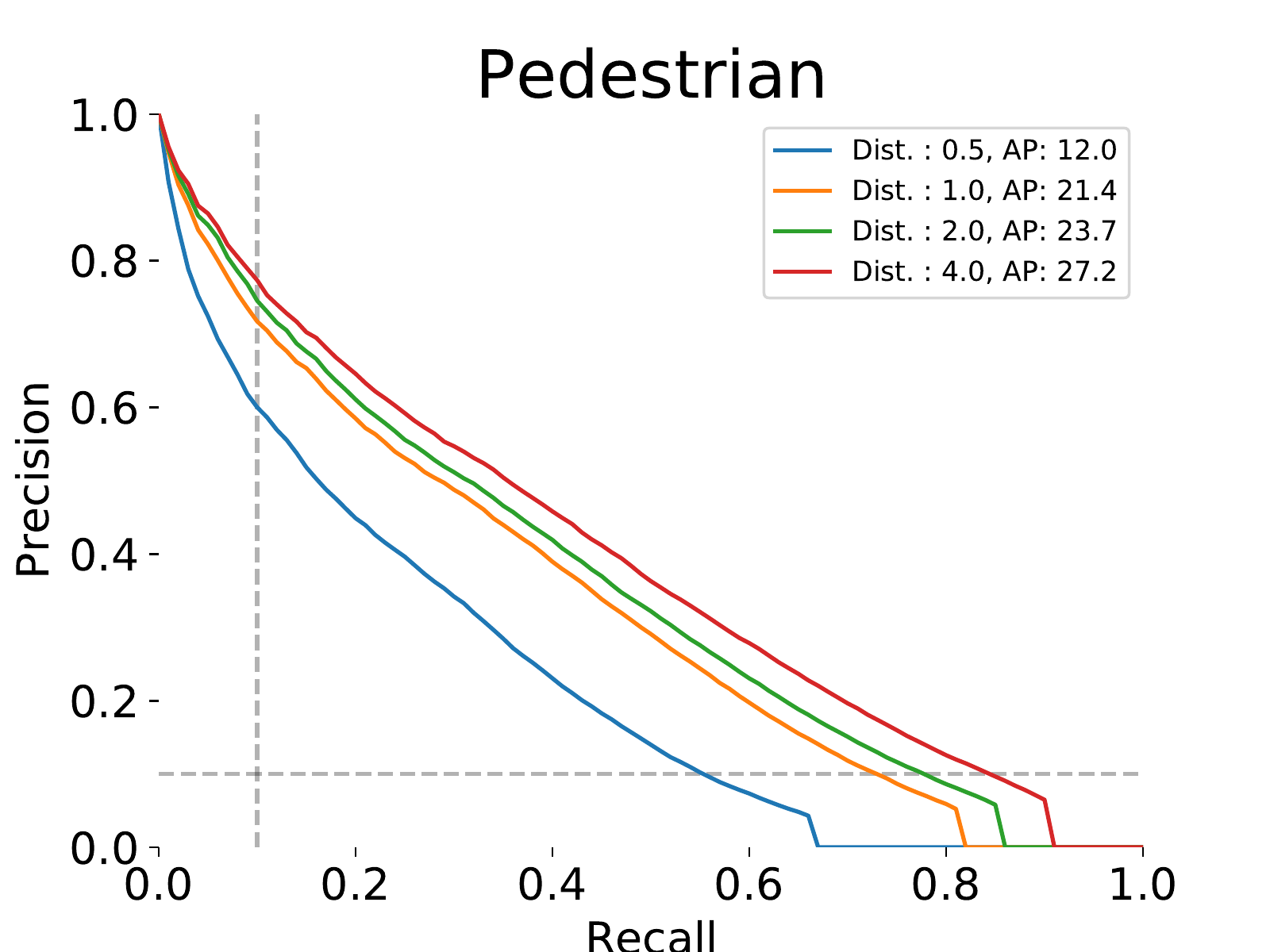}
				\\
				\includegraphics[width=1\columnwidth]{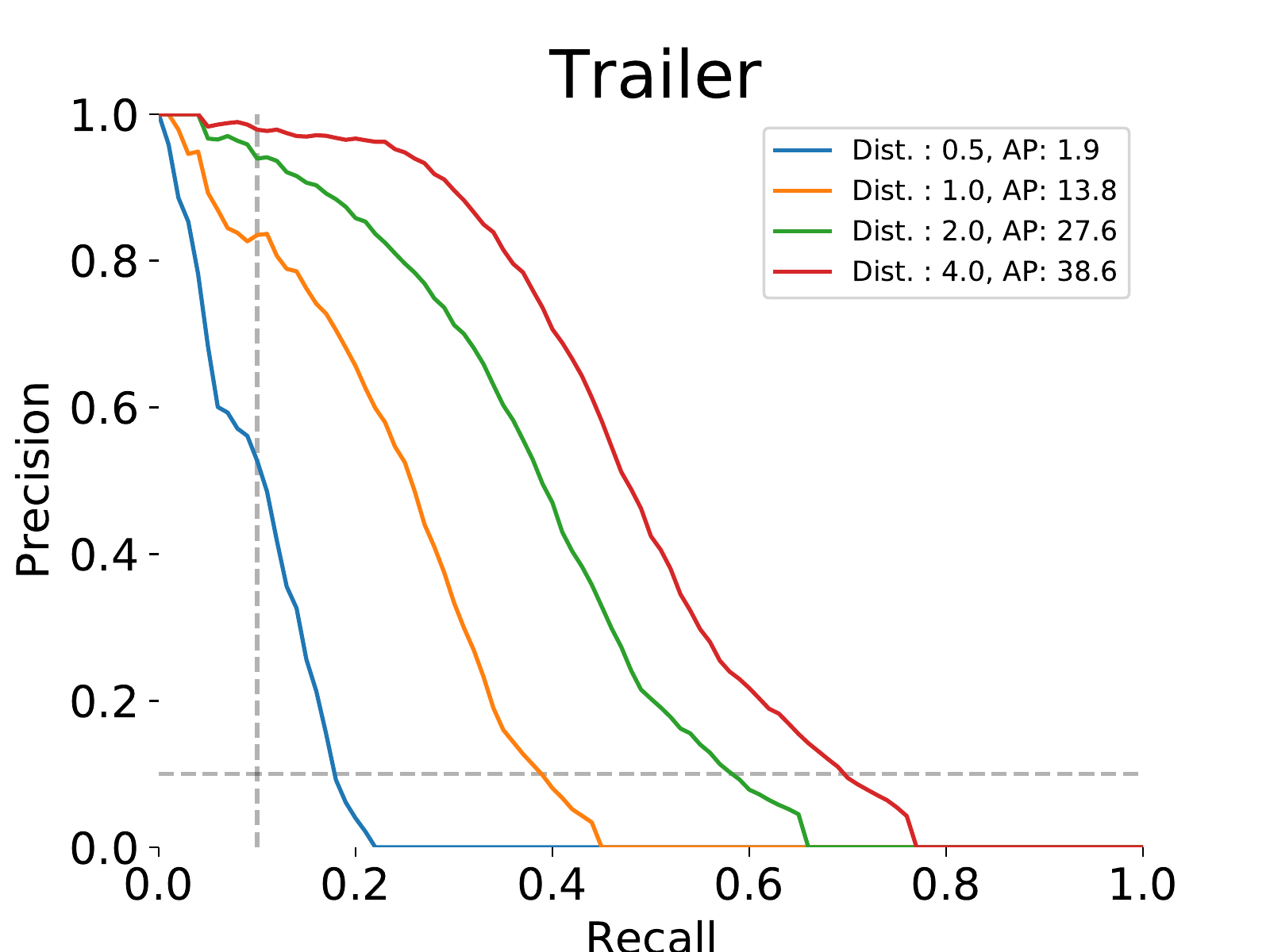}
			\end{minipage}
		}
		\subfigure[t] { 
			\begin{minipage}{0.186\textwidth}
				\centering
				\includegraphics[width=1\columnwidth]{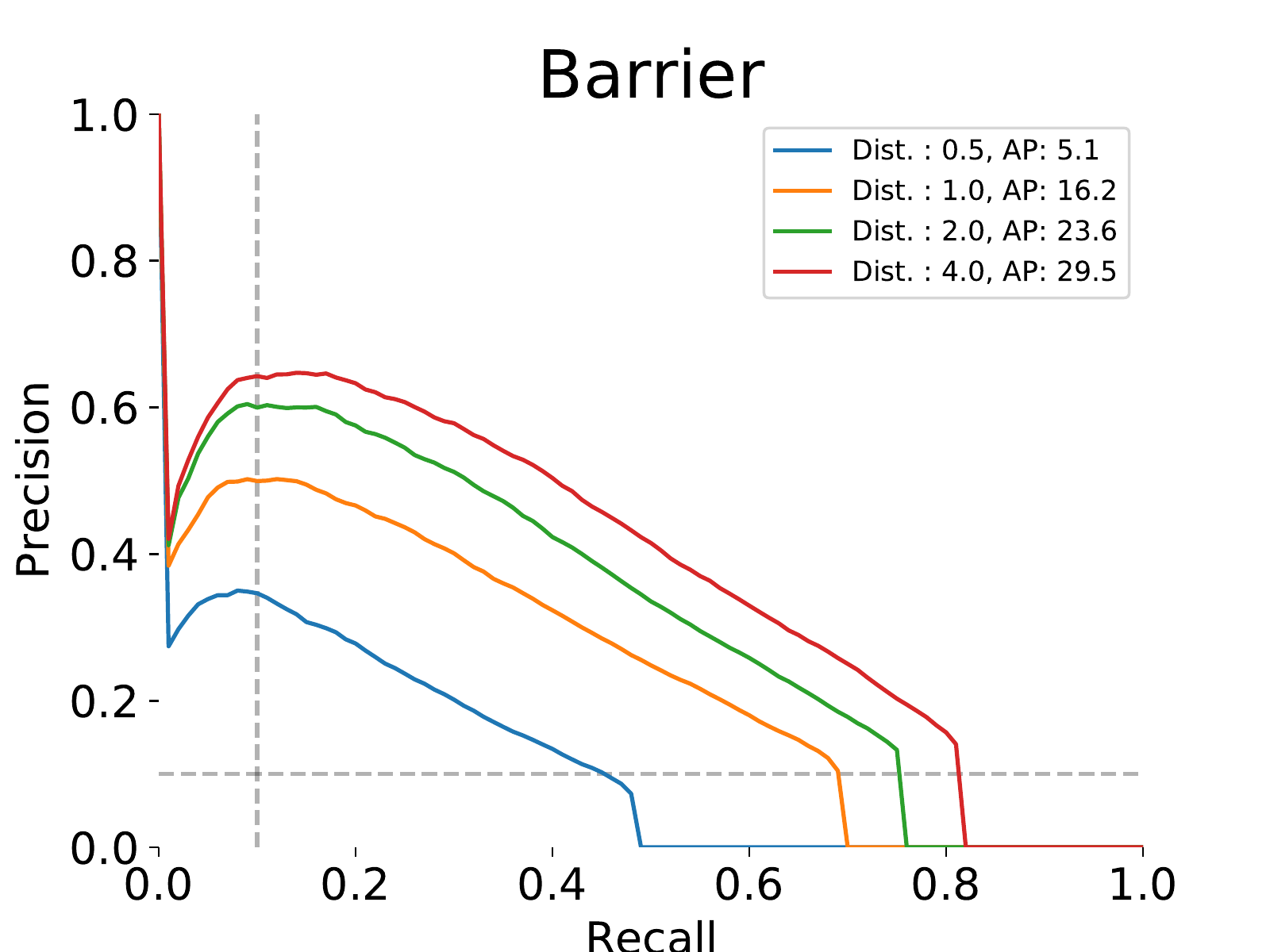}
				\\
				\includegraphics[width=1\columnwidth]{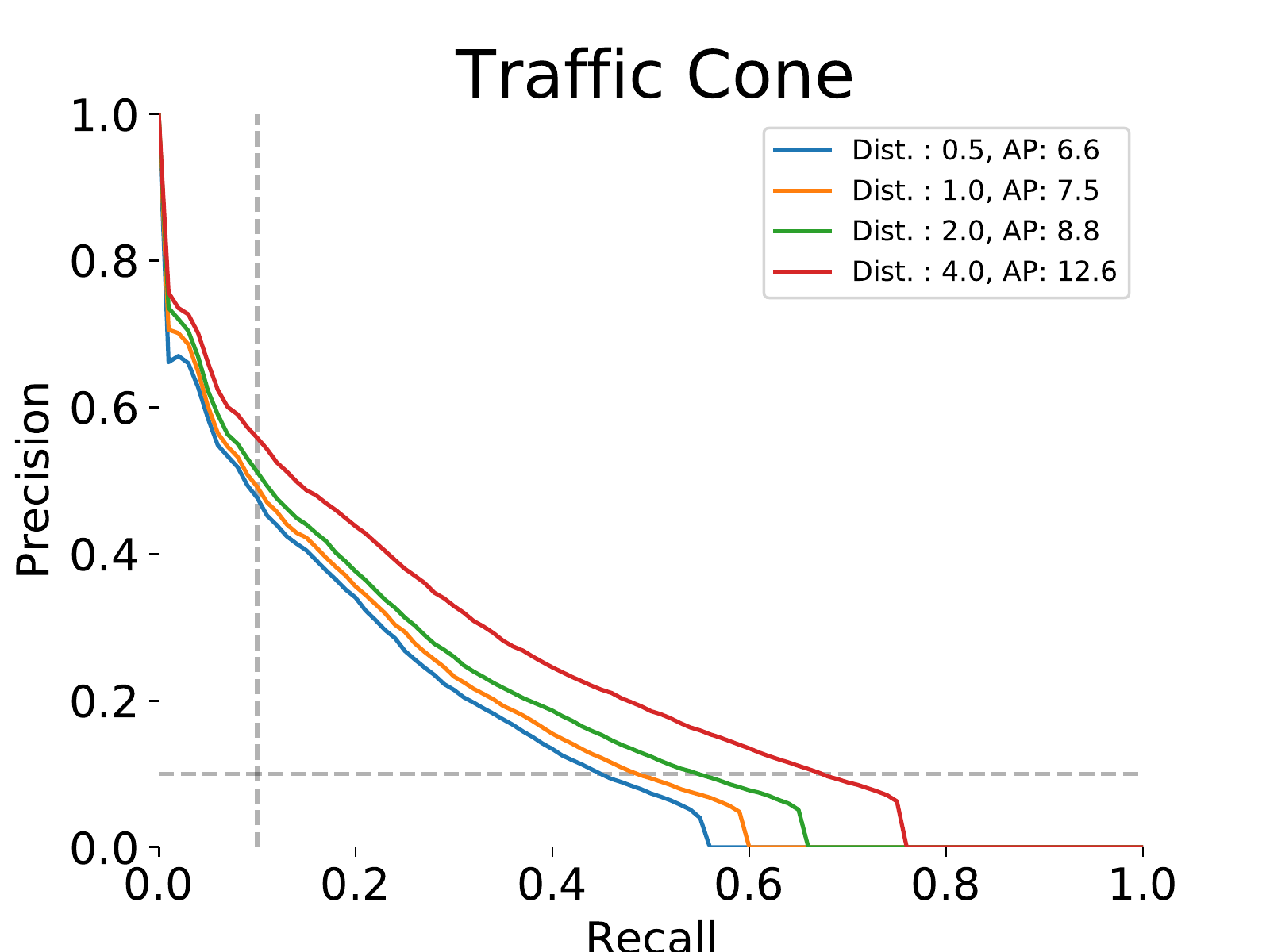}
			\end{minipage}
		}		
	\end{center}
	
	\caption{Per category precision-recall plot of attacking the \textbf{rotation} only (black box) on the nuScenes validation set [2].}
	\label{fig:black_INS}
	%\label{fig:onecol}
\end{figure*}

\begin{figure*}[t]
	\begin{center}

		\subfigure[t] { 
			\begin{minipage}{0.186\textwidth}
				\centering
				\includegraphics[width=1\columnwidth]{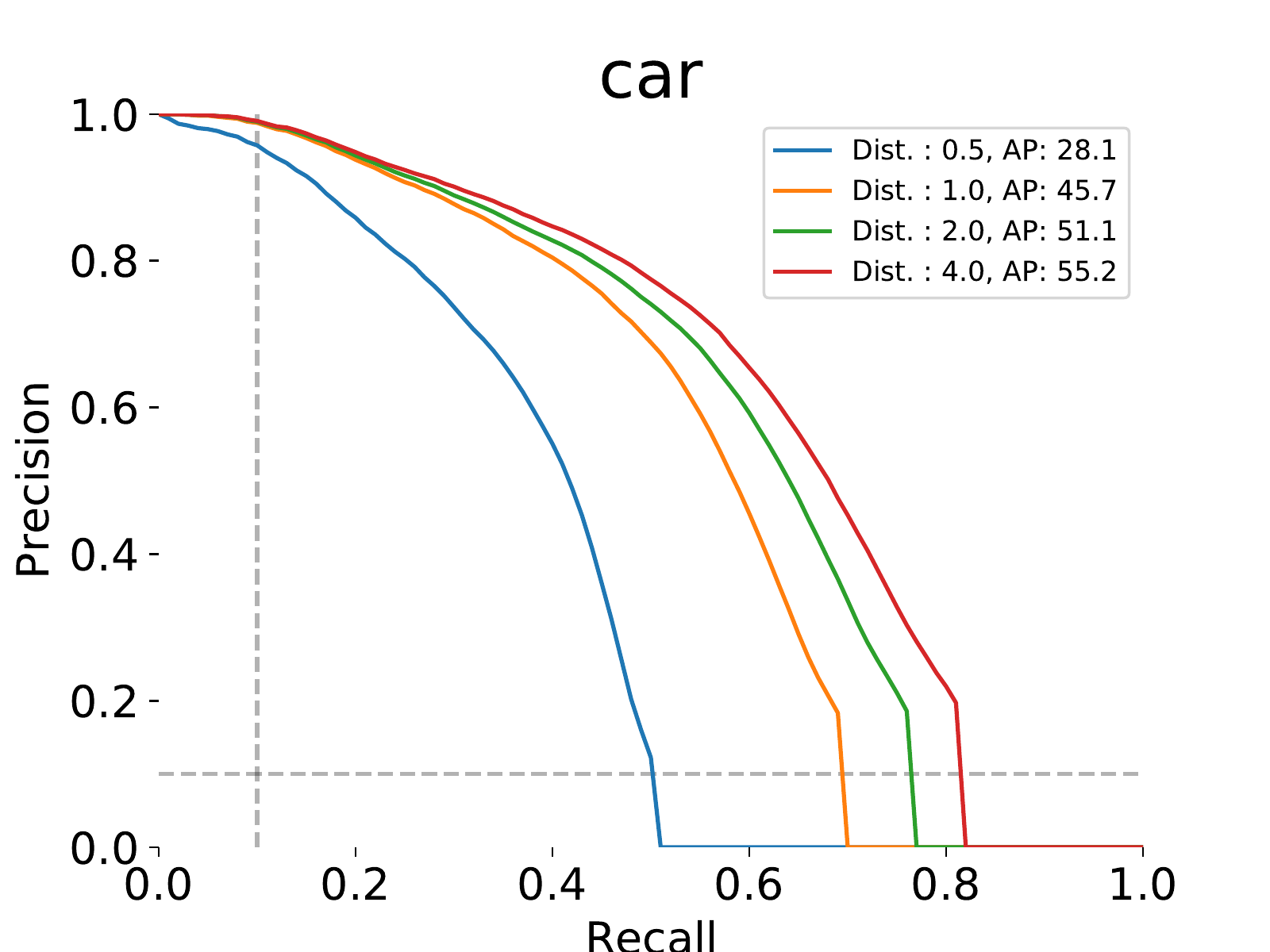}
				\\
				\includegraphics[width=1\columnwidth]{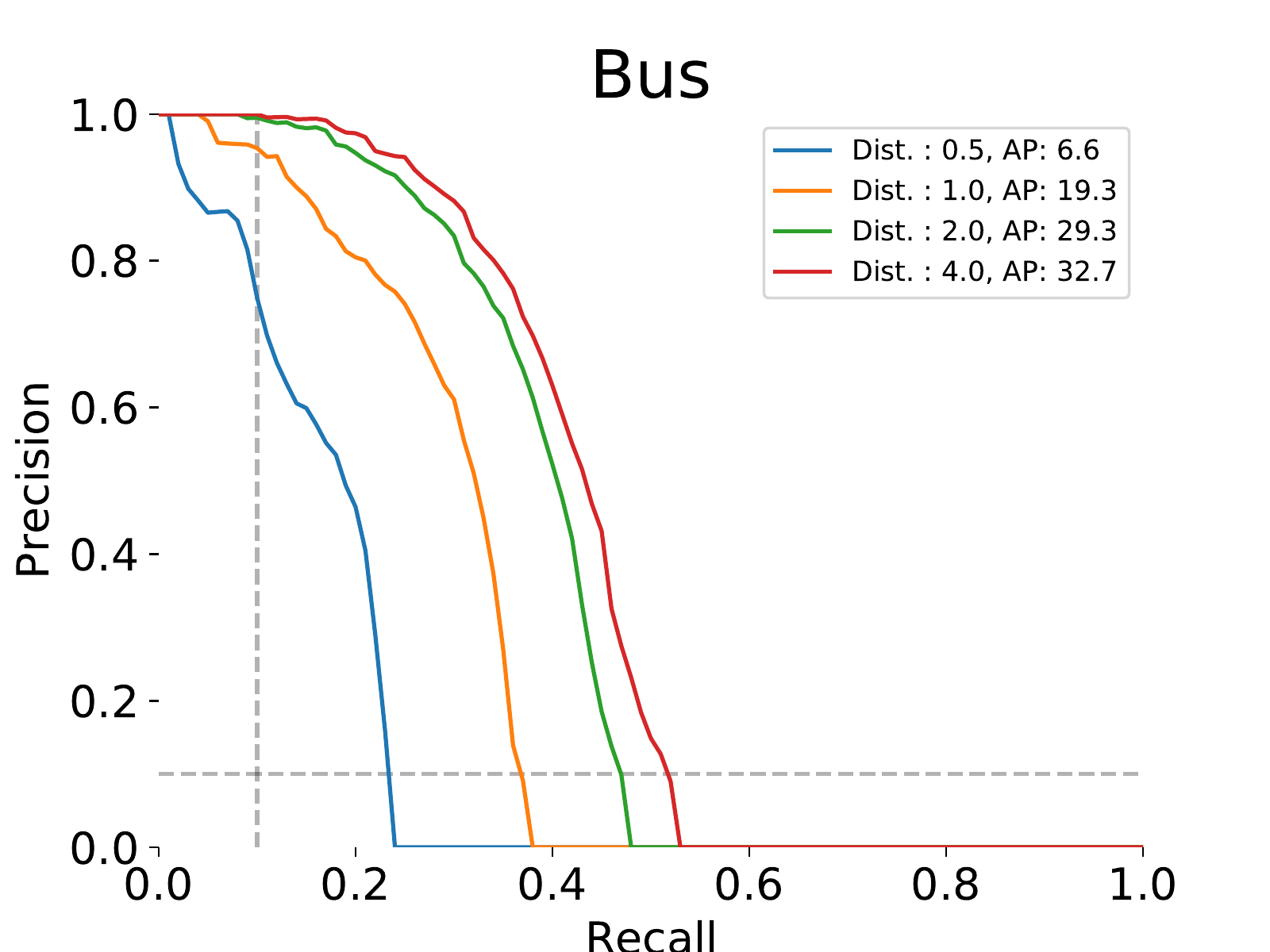}
			\end{minipage}
		}
		\subfigure[t] {  
			\begin{minipage}{0.186\textwidth}
				\centering
				\includegraphics[width=1\columnwidth]{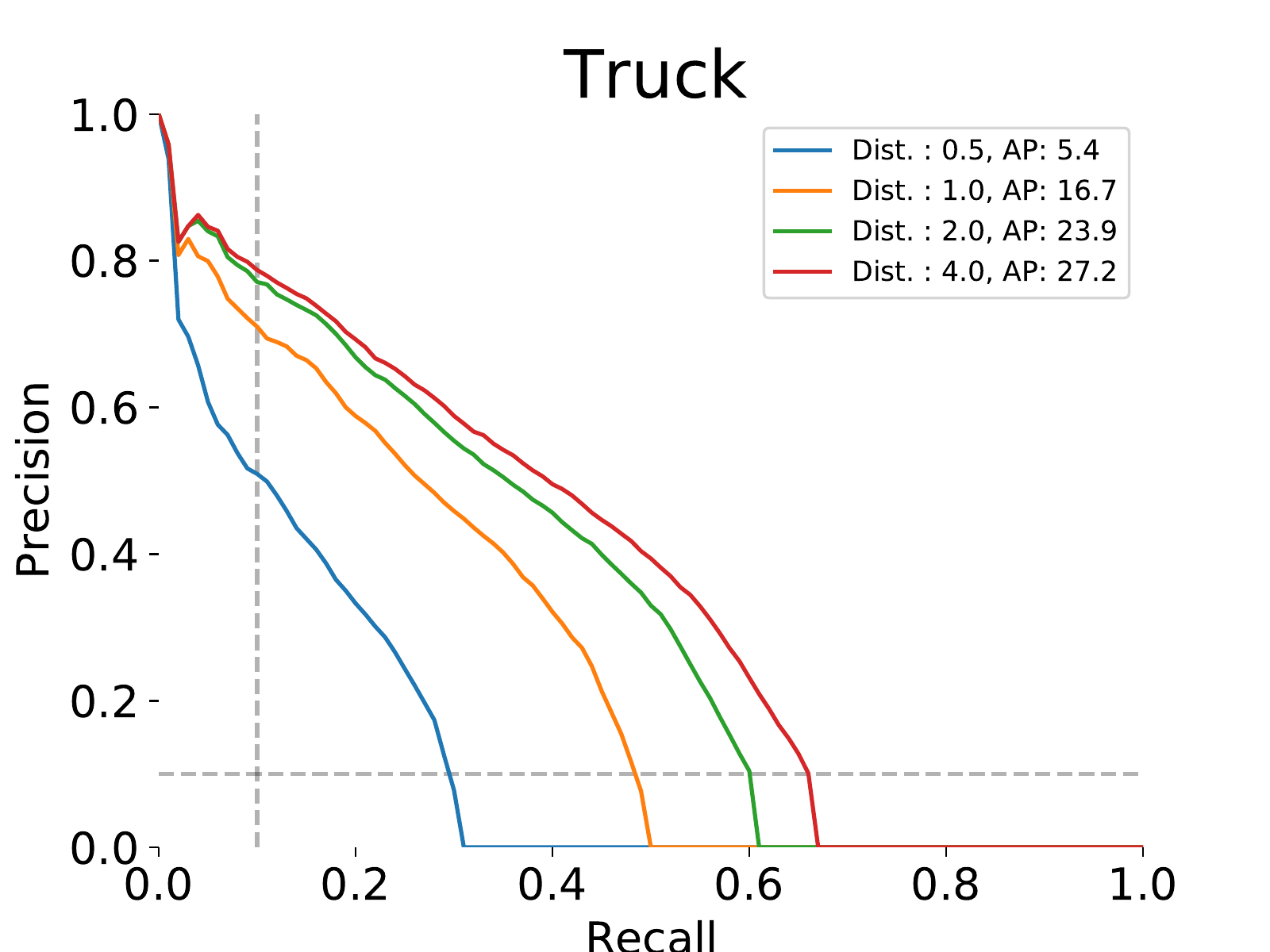}
				\\
				\includegraphics[width=1\columnwidth]{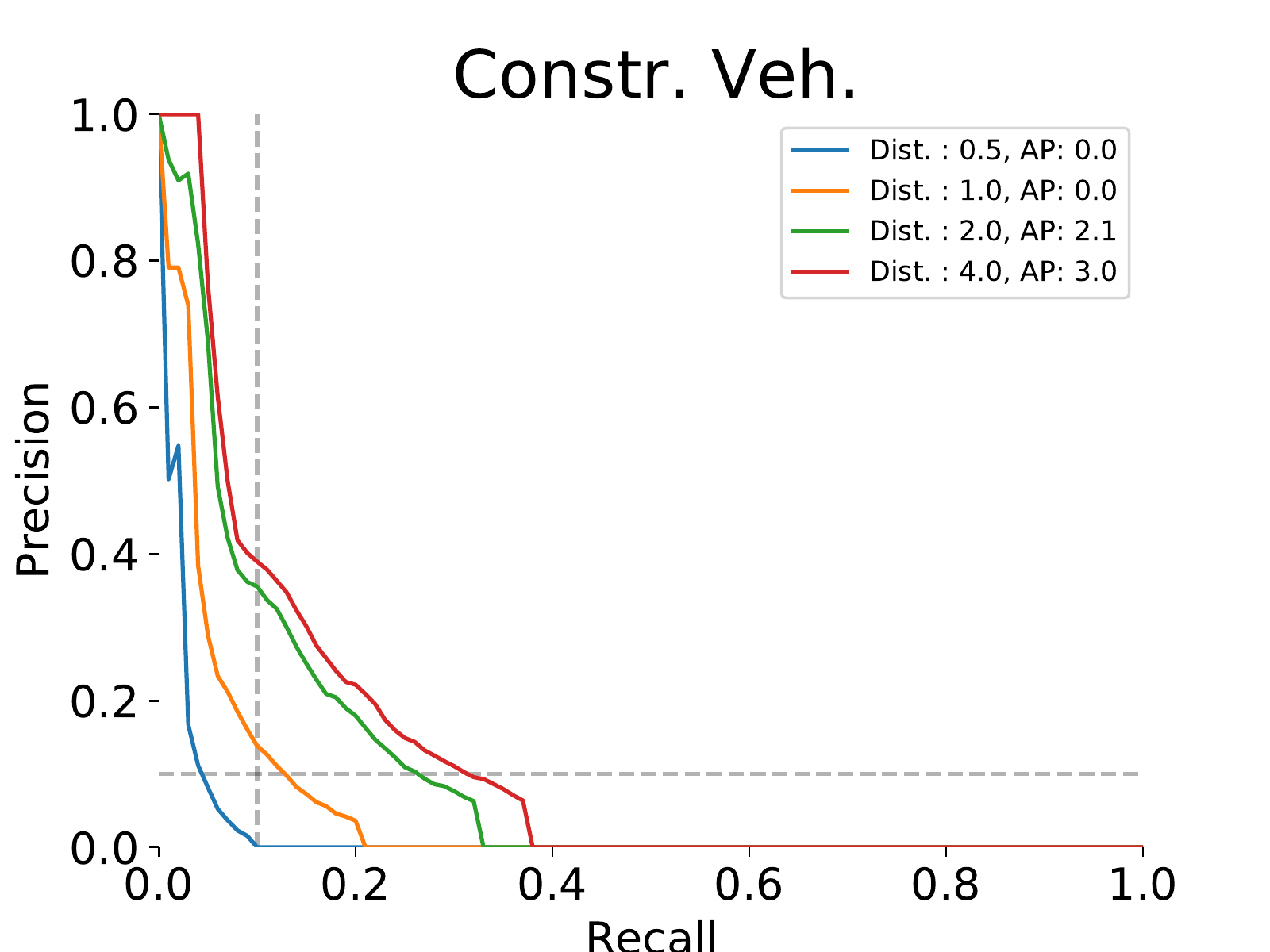}
			\end{minipage}
		}
		\subfigure[t] { 
			\begin{minipage}{0.186\textwidth}
				\centering
				\includegraphics[width=1\columnwidth]{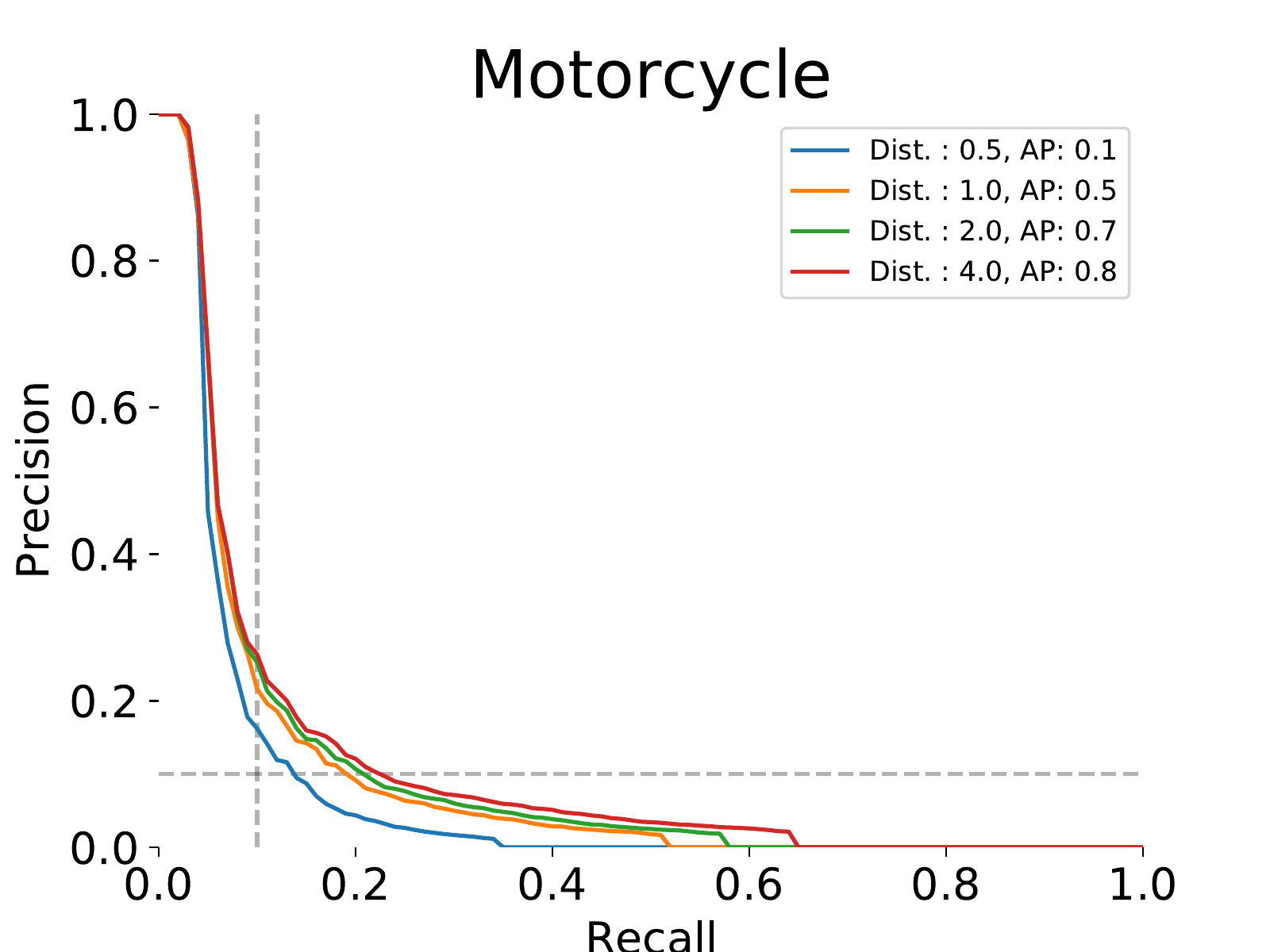}
				\\
				\includegraphics[width=1\columnwidth]{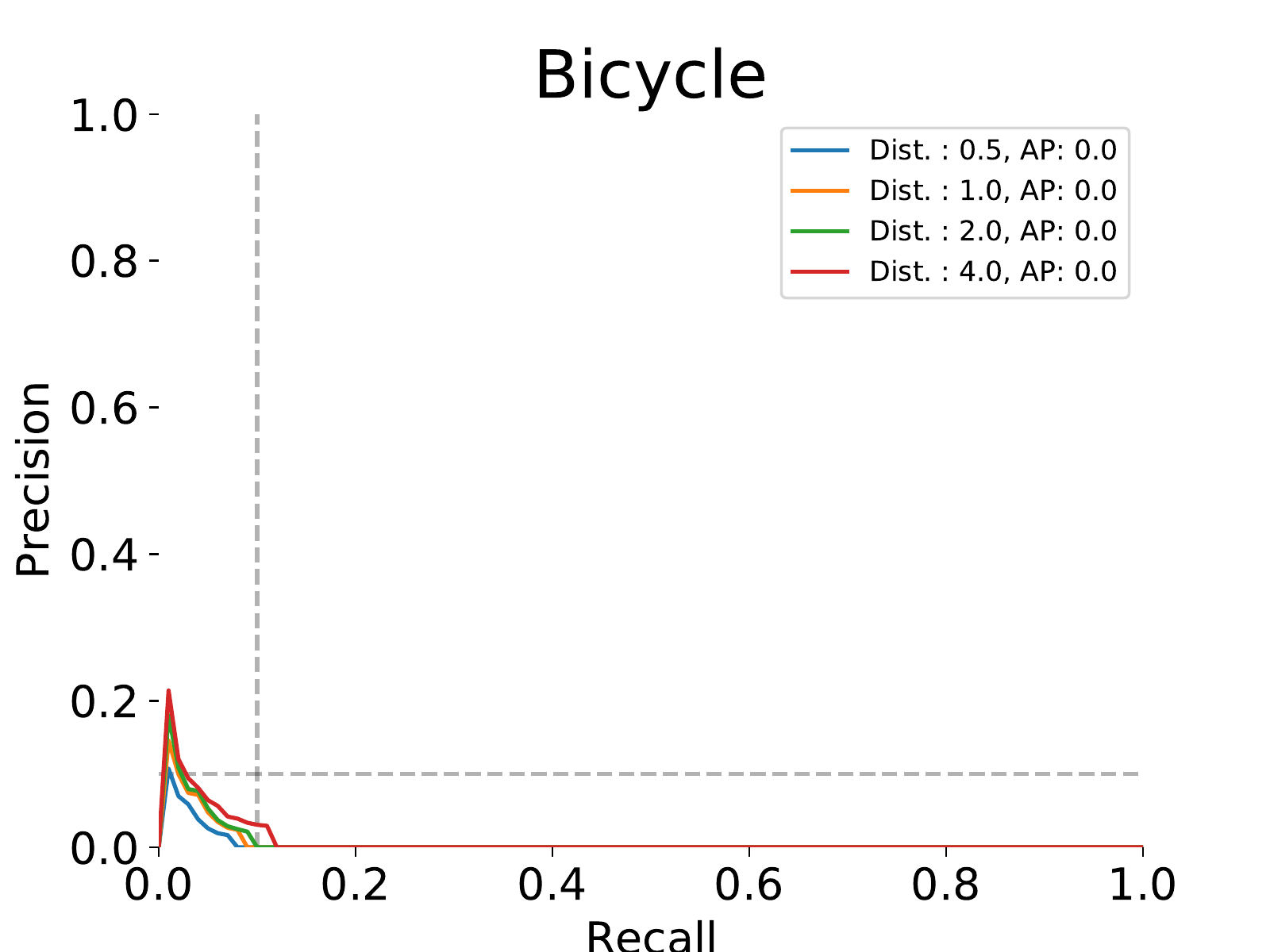}
			\end{minipage}
		}
		\subfigure[t] { 
			\begin{minipage}{0.186\textwidth}
				\centering
				\includegraphics[width=1\columnwidth]{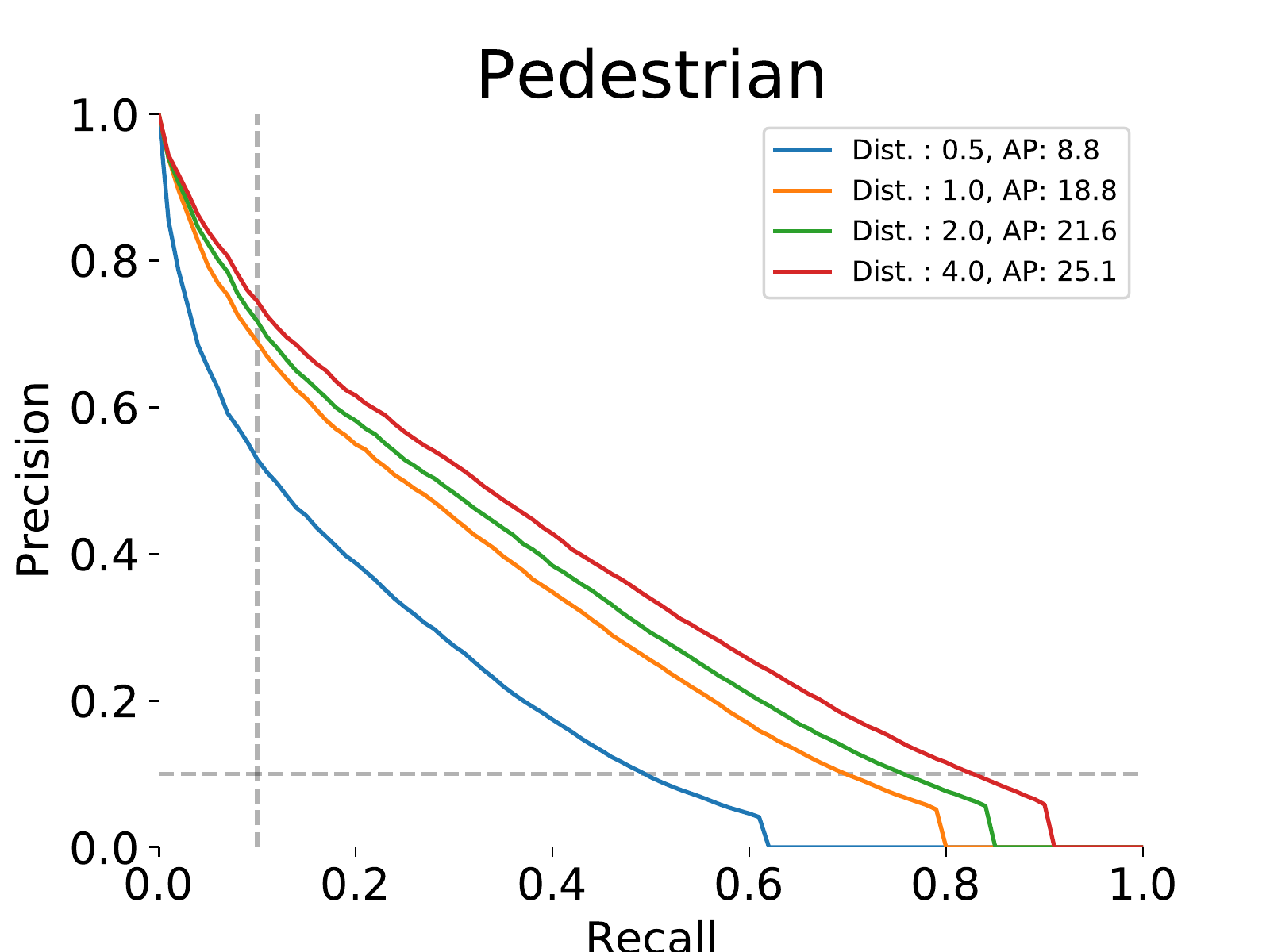}
				\\
				\includegraphics[width=1\columnwidth]{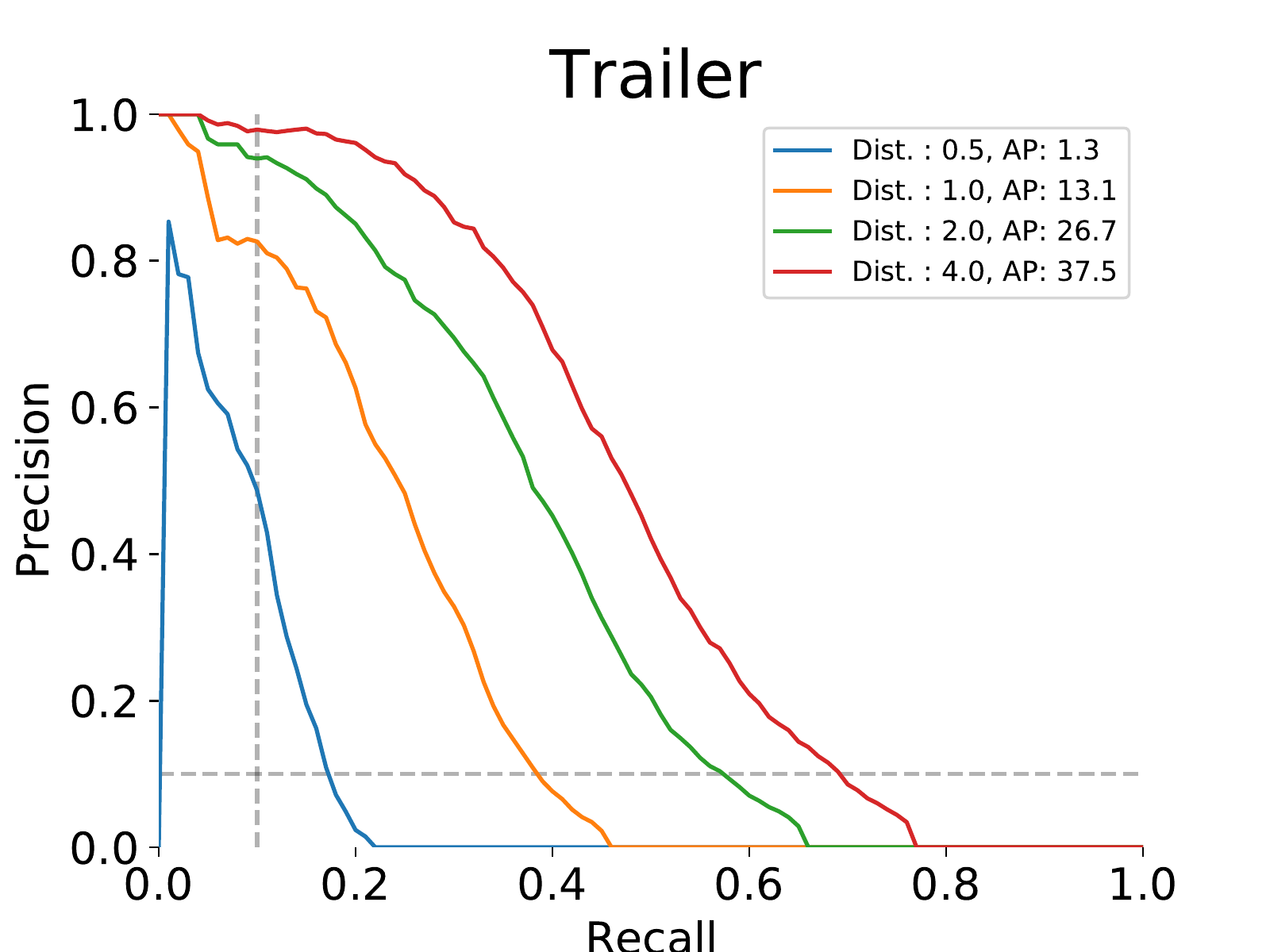}
			\end{minipage}
		}
		\subfigure[t] { 
			\begin{minipage}{0.186\textwidth}
				\centering
				\includegraphics[width=1\columnwidth]{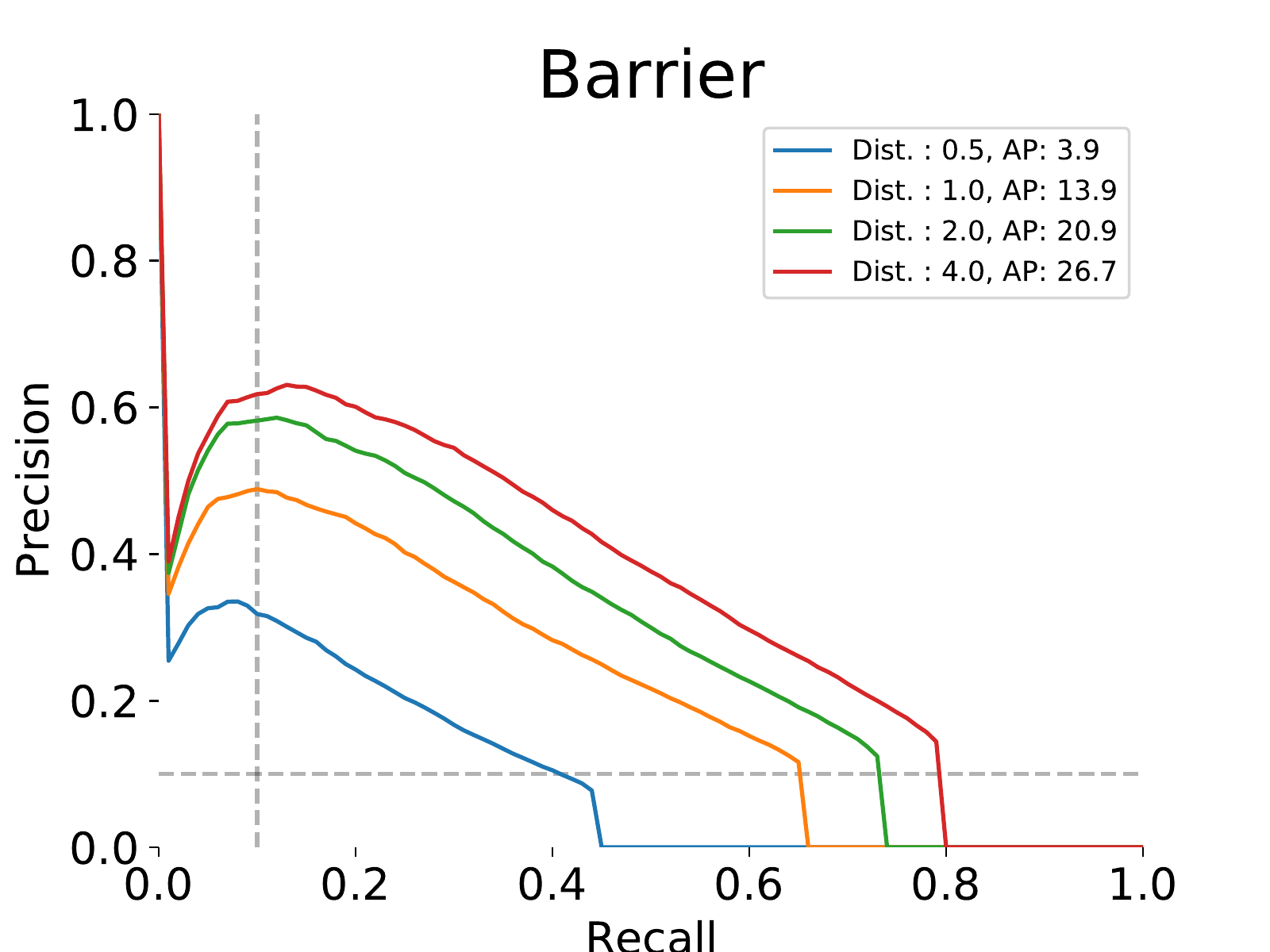}
				\\
				\includegraphics[width=1\columnwidth]{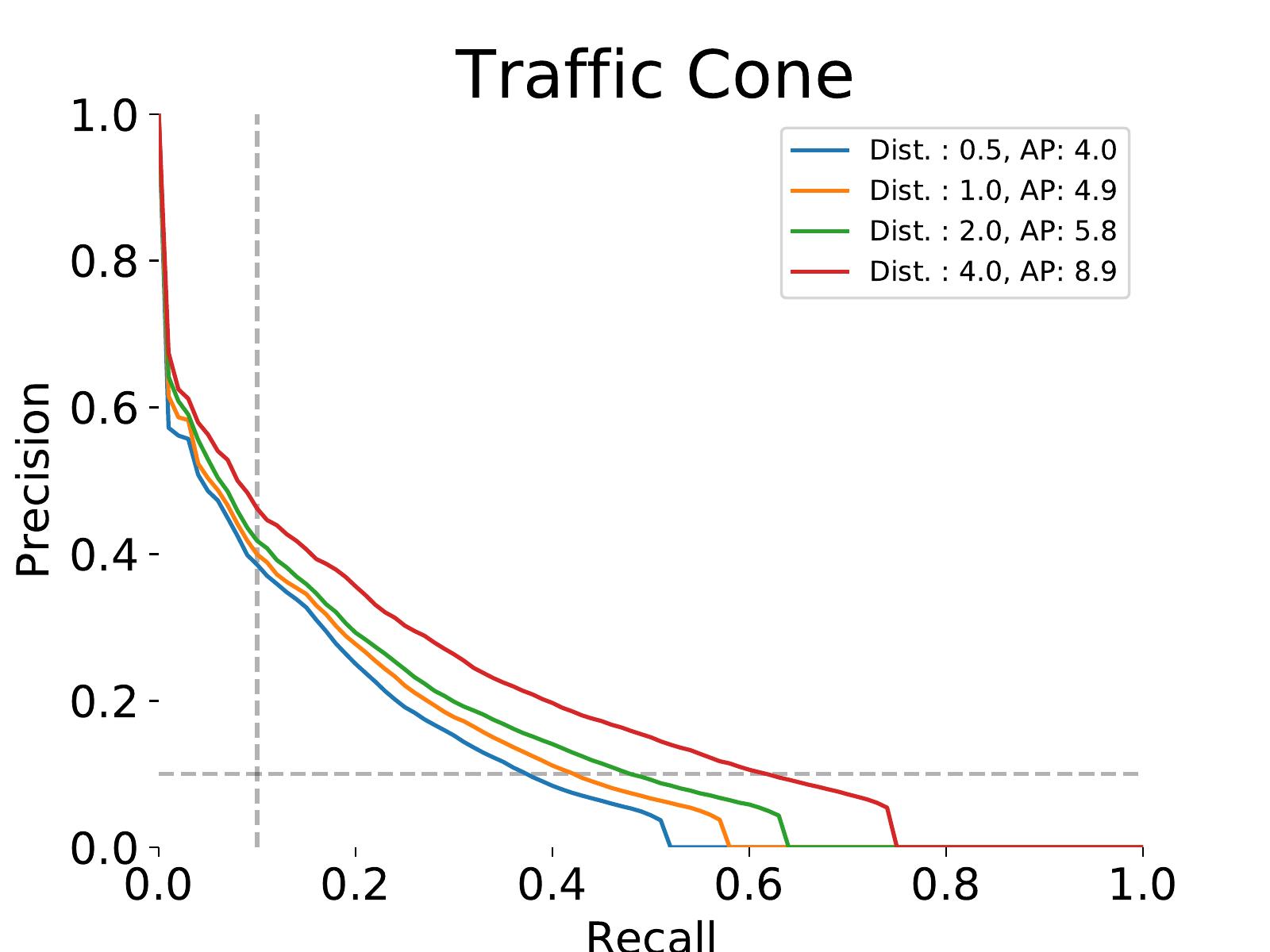}
			\end{minipage}
		}		
	\end{center}
	
	\caption{Per category precision-recall plot of attacking the \textbf{full trajectory} (black box) on the nuScenes validation set [2].}
	\label{fig:black_GNSS_INS}
	%\label{fig:onecol}
\end{figure*}

\begin{figure*}[t]
	\begin{center}

		\subfigure[t] {  
			\begin{minipage}{0.236\textwidth}
				\centering
				\includegraphics[width=1\columnwidth]{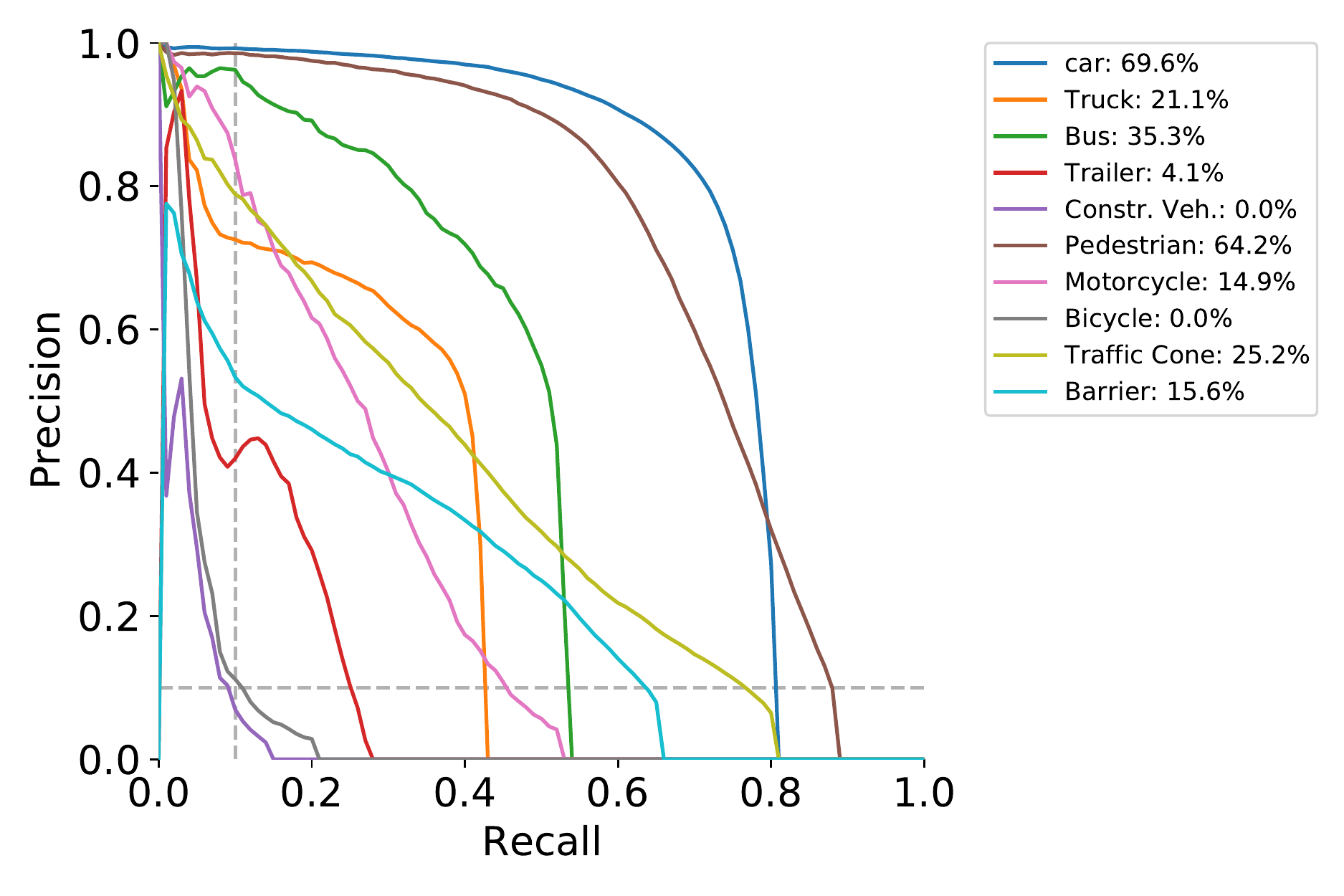}
				\\
				\includegraphics[width=1\columnwidth]{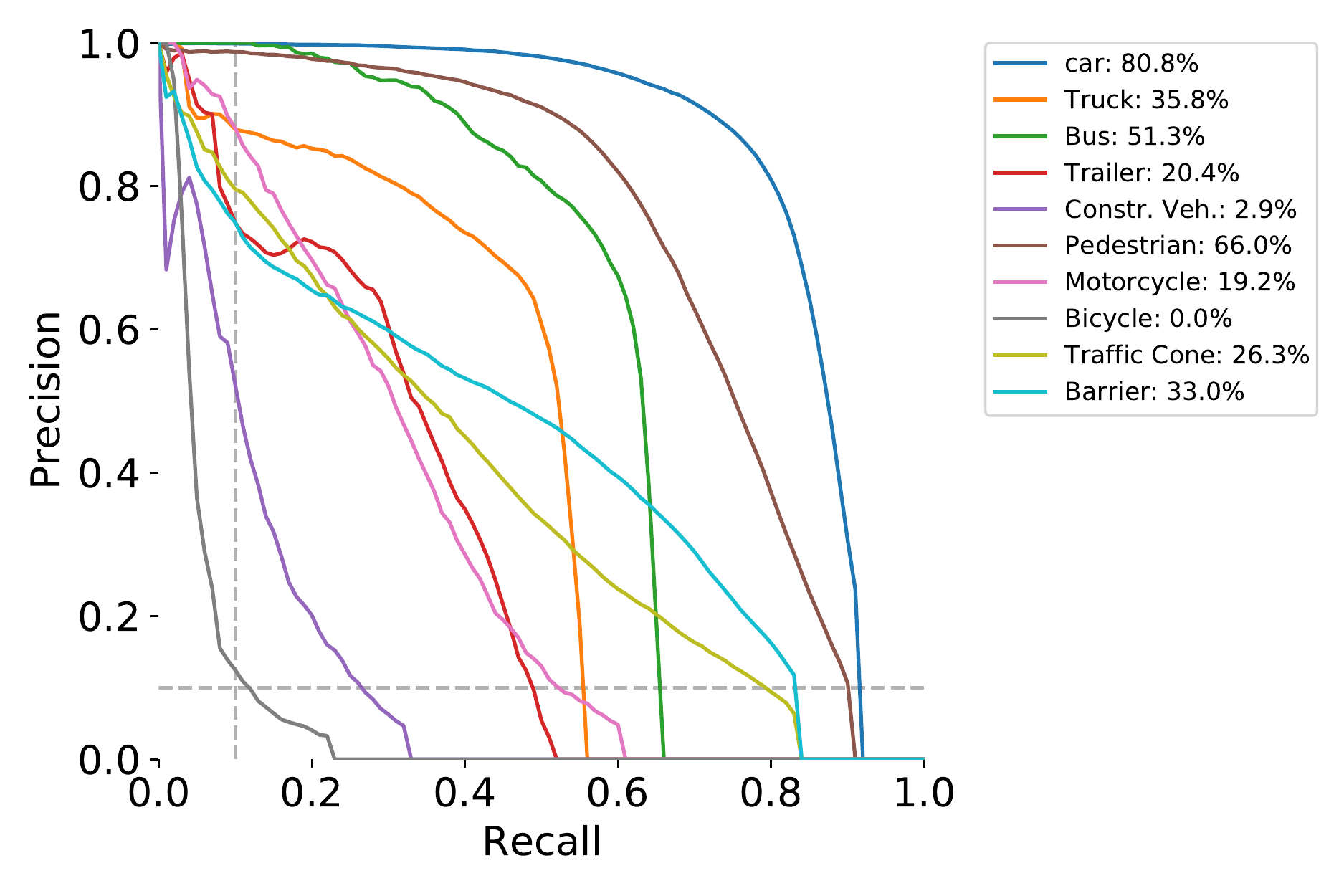}
				\\
				\includegraphics[width=1\columnwidth]{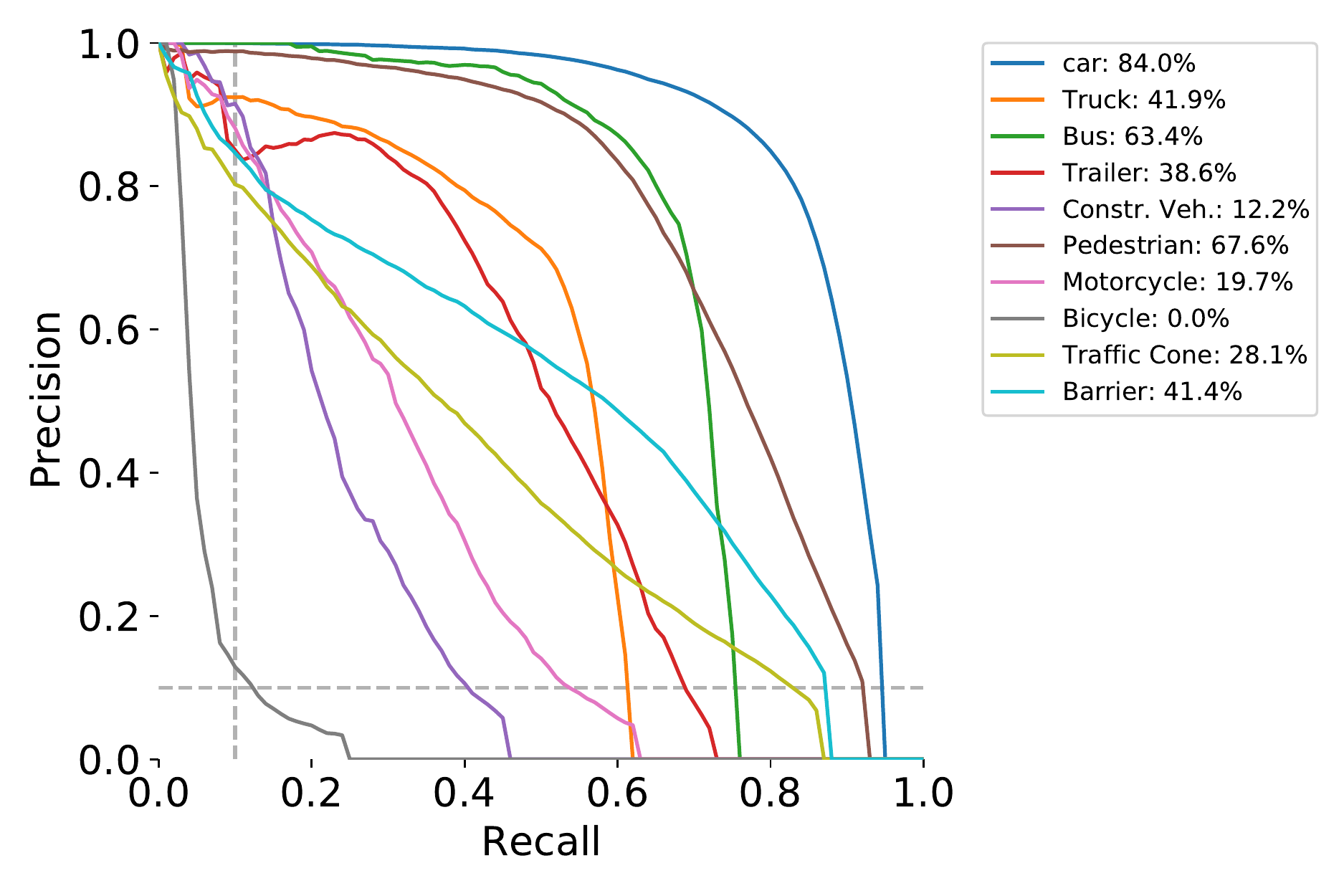}
				\\
				\includegraphics[width=1\columnwidth]{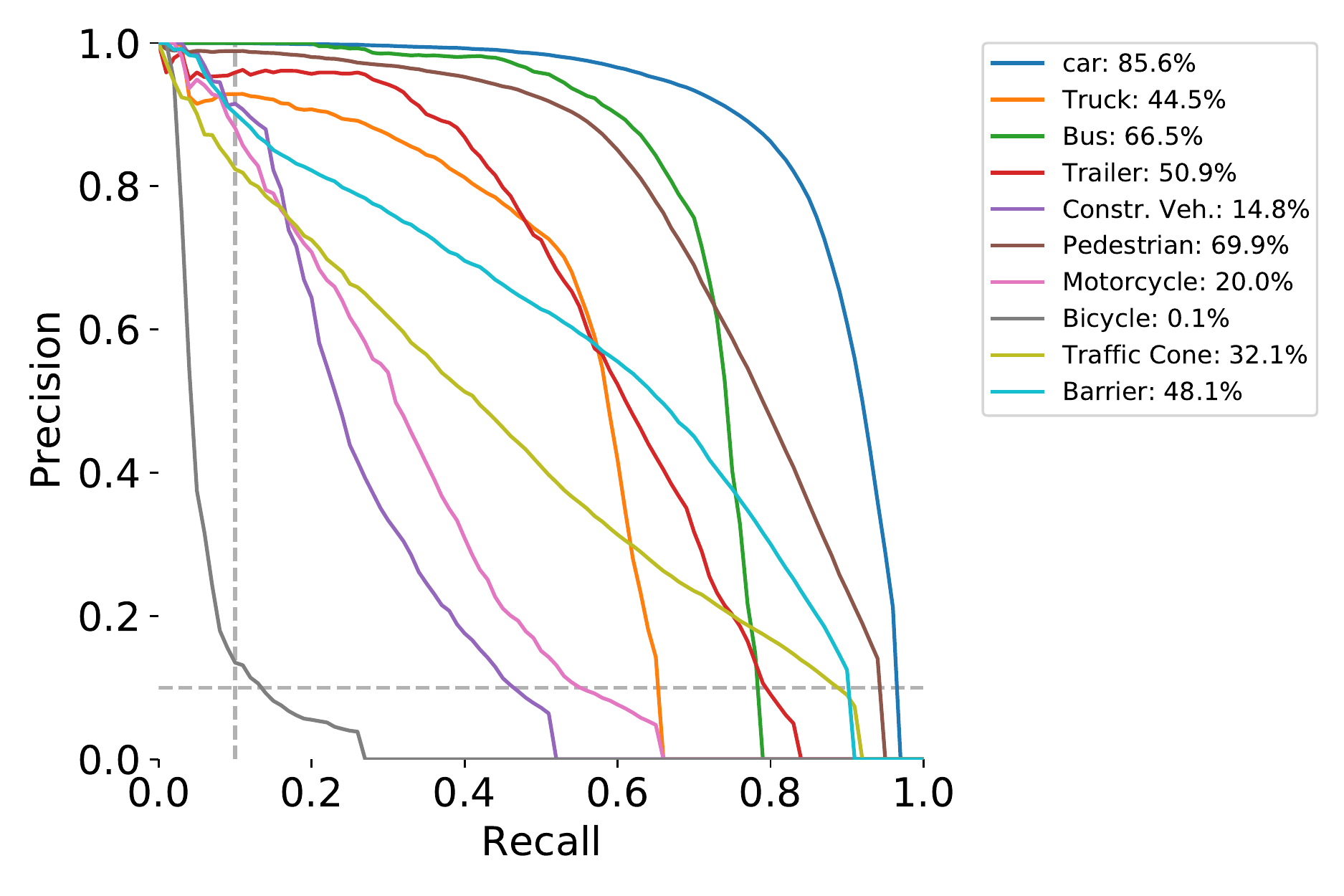}				
			\end{minipage}
		}
		\subfigure[t] {  
			\begin{minipage}{0.236\textwidth}
				\centering
				\includegraphics[width=1\columnwidth]{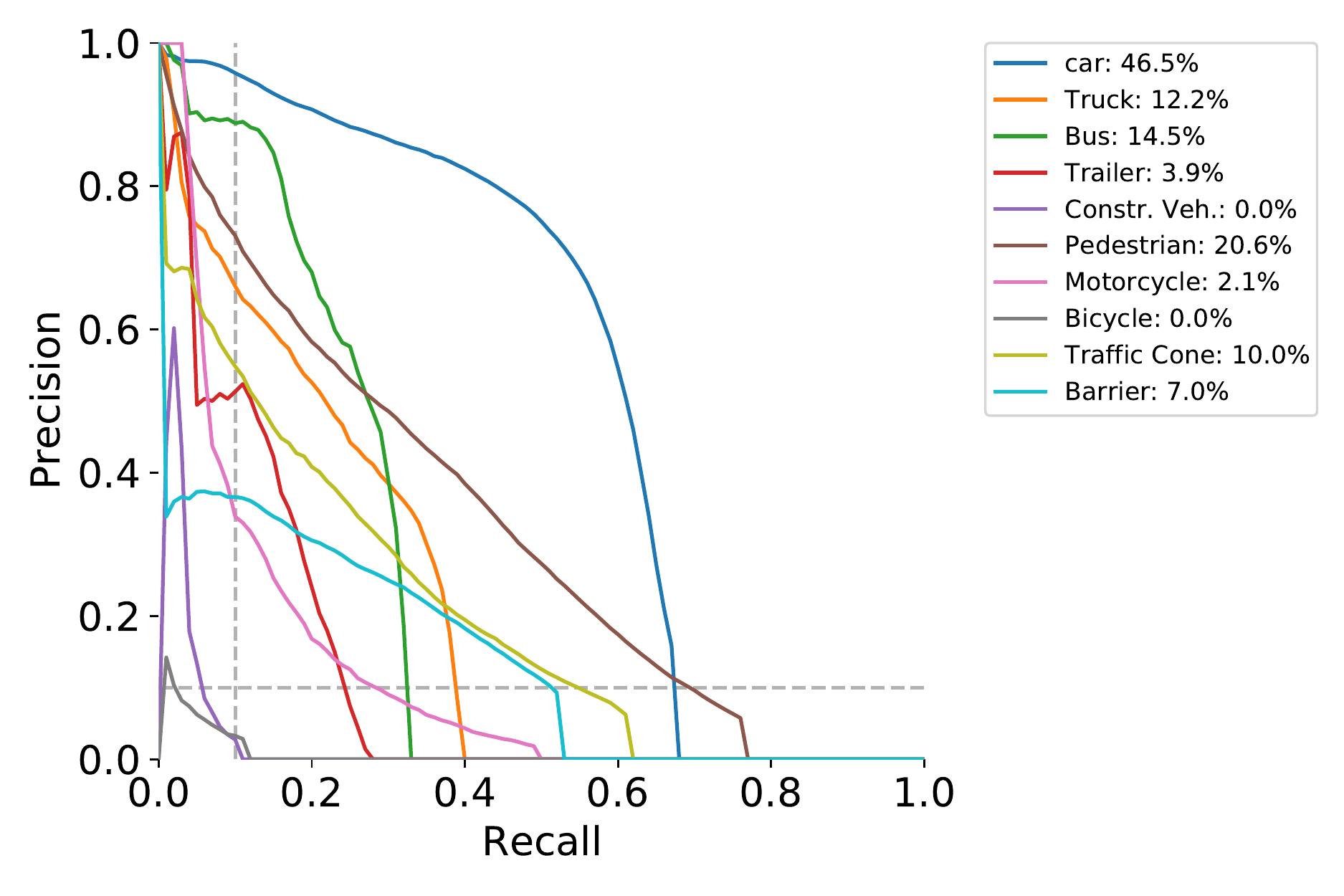}
				\\
				\includegraphics[width=1\columnwidth]{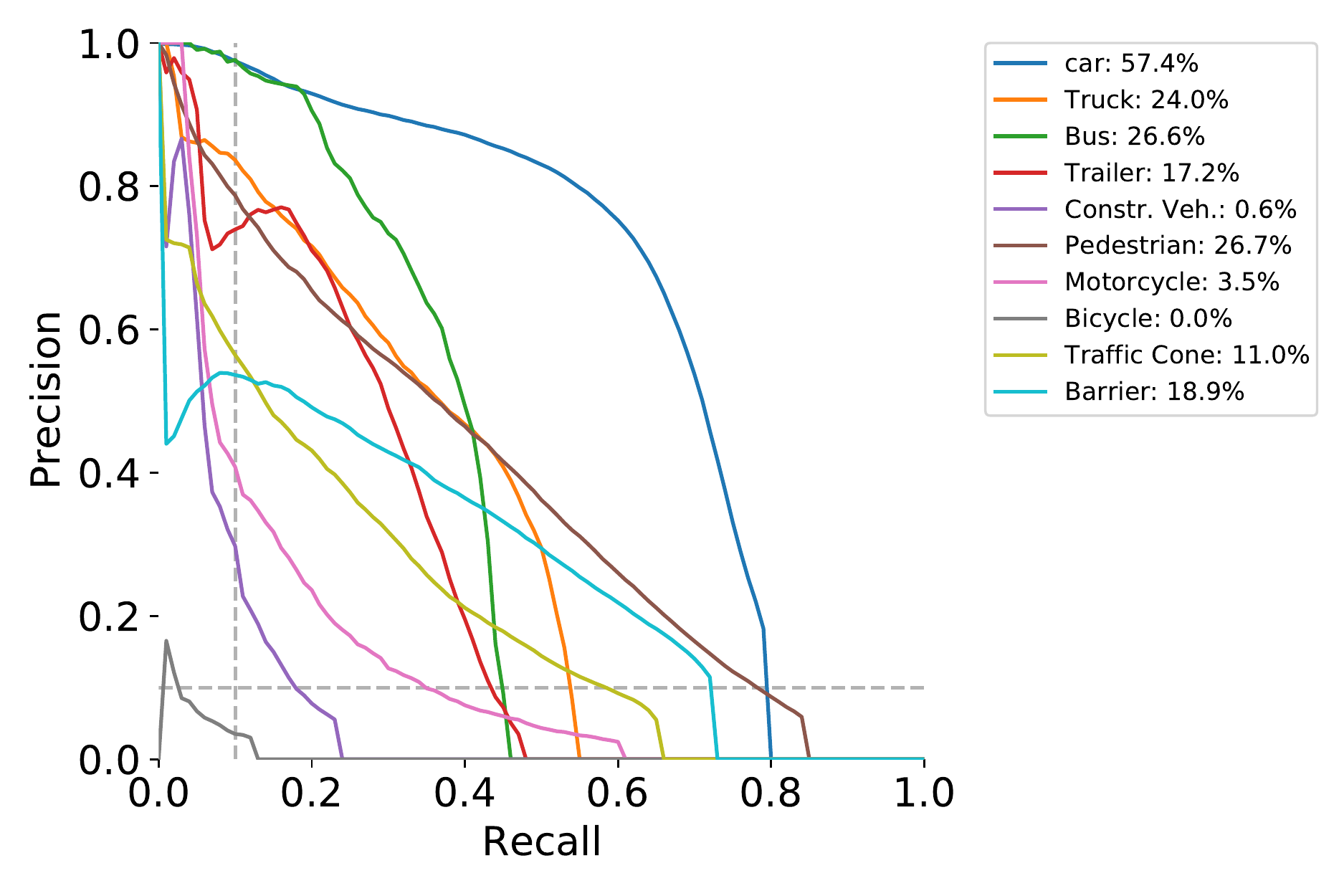}
				\\
				\includegraphics[width=1\columnwidth]{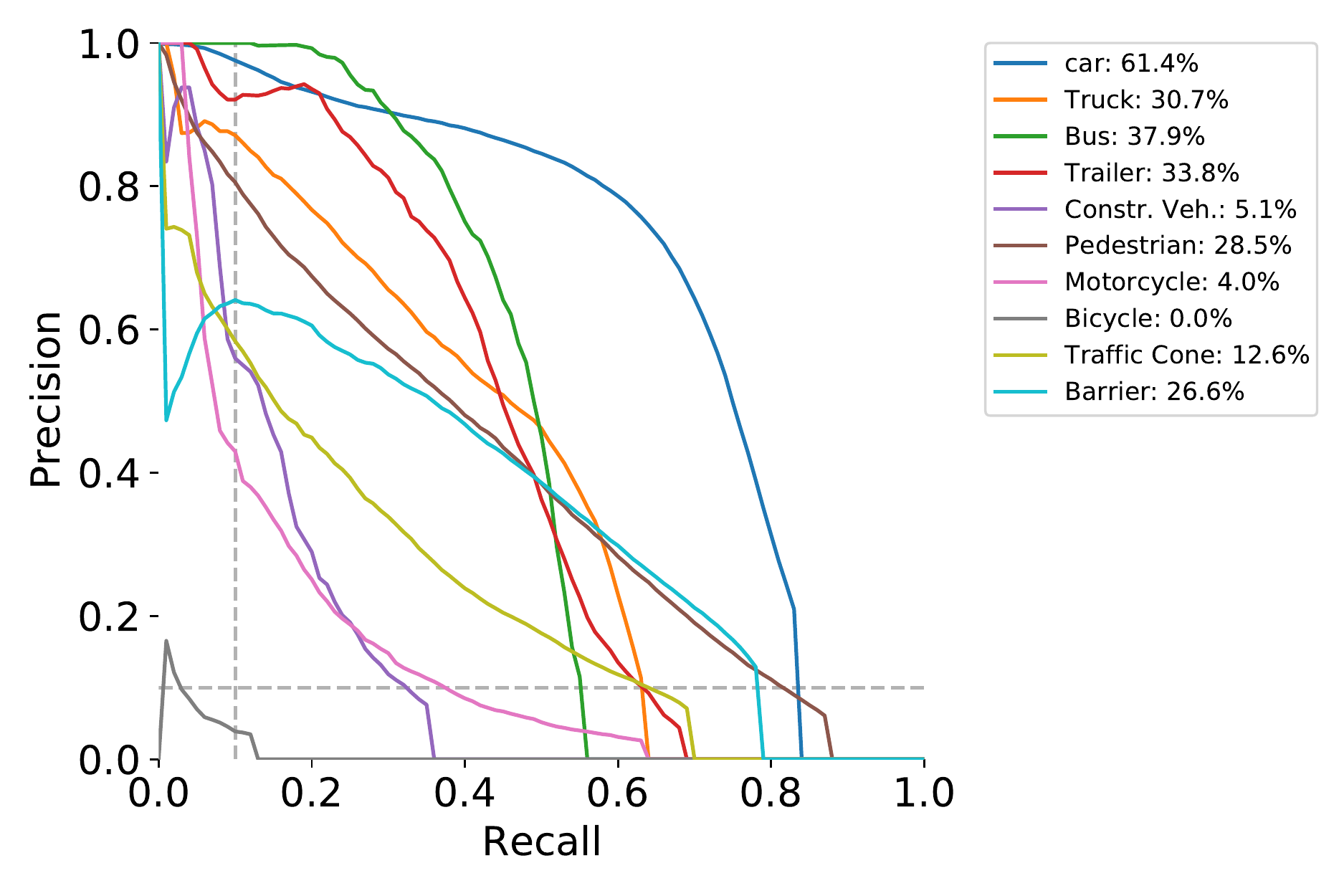}
				\\
				\includegraphics[width=1\columnwidth]{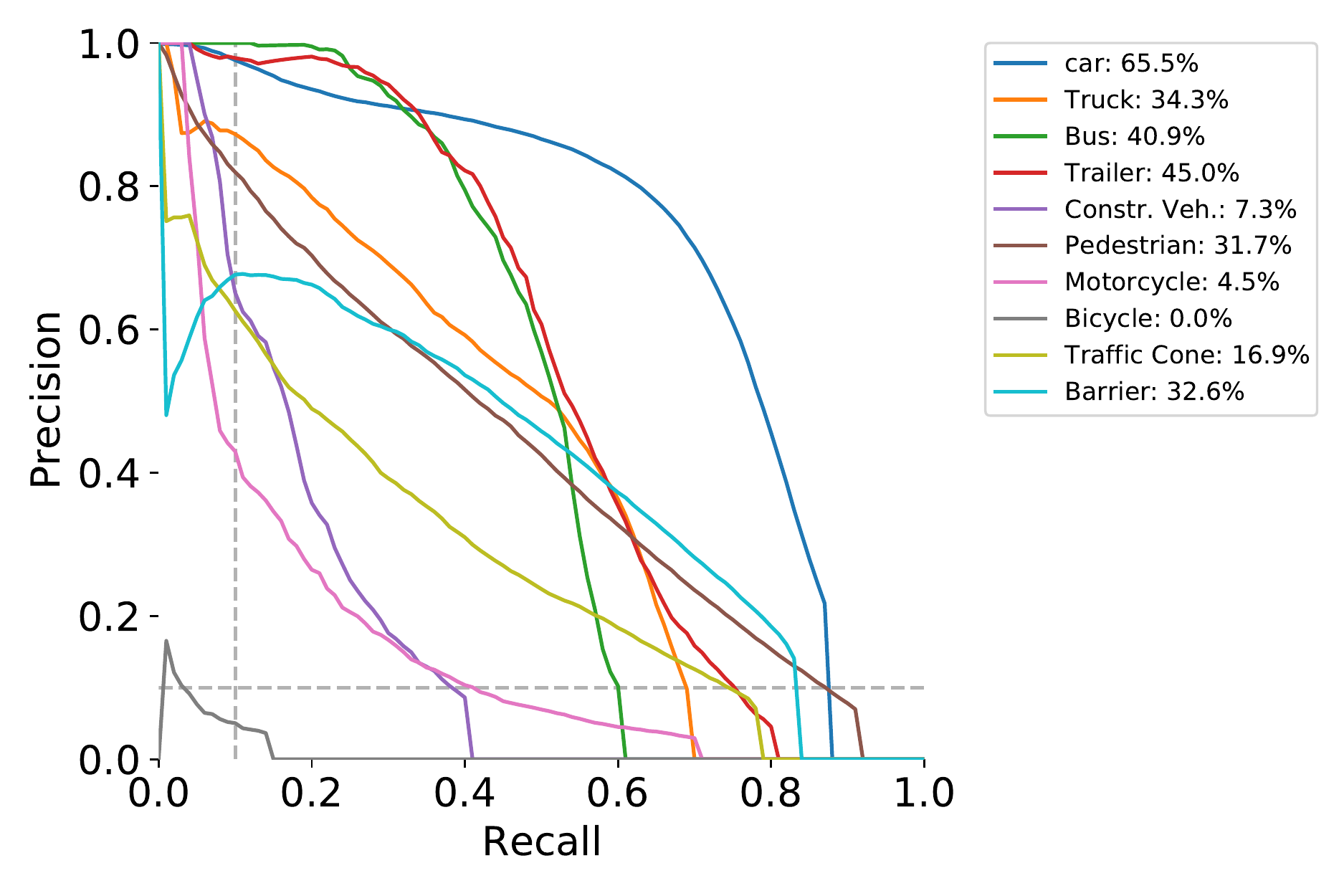}	
			\end{minipage}
		}
		\subfigure[t] {  
			\begin{minipage}{0.236\textwidth}
				\centering
				\includegraphics[width=1\columnwidth]{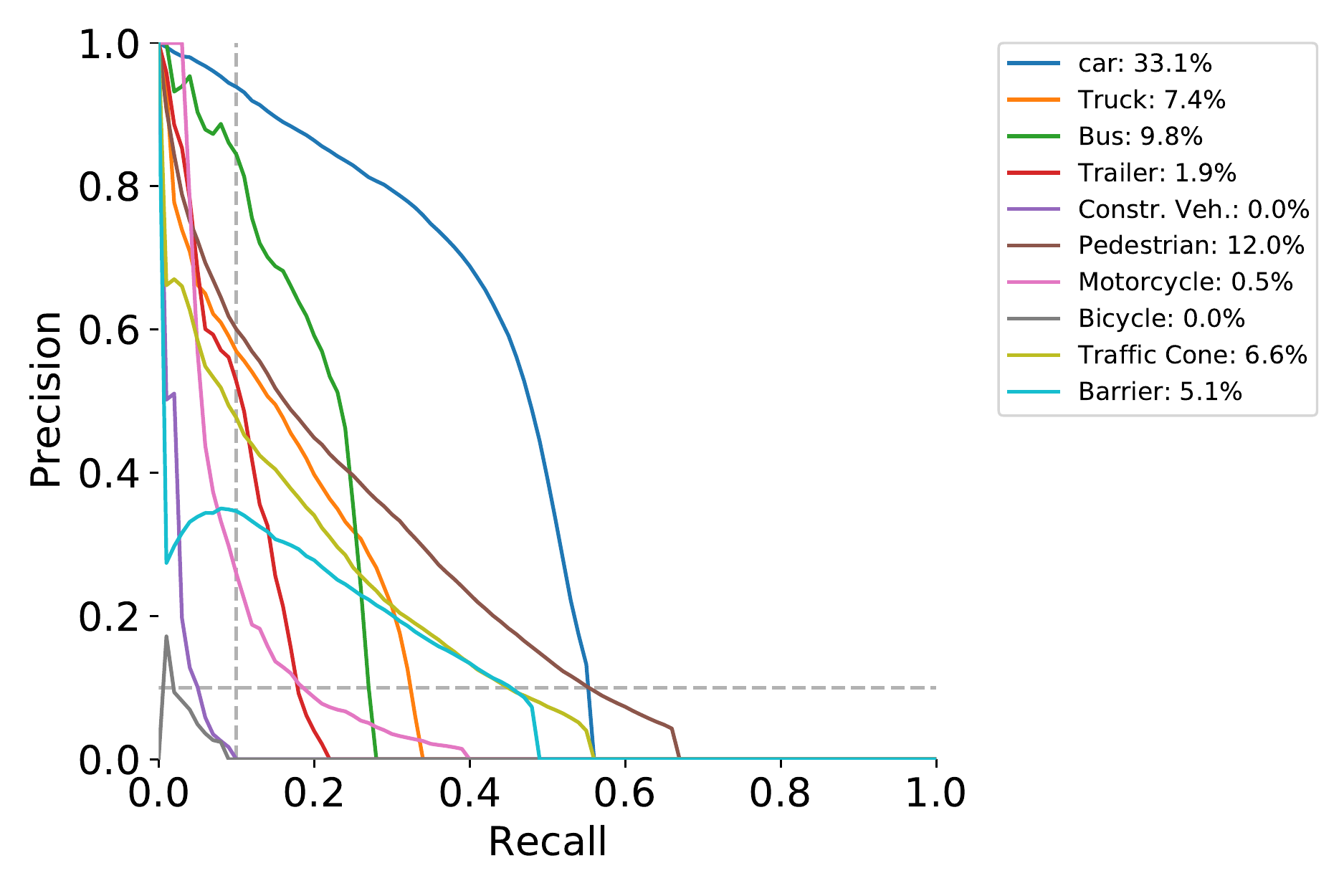}
				\\
				\includegraphics[width=1\columnwidth]{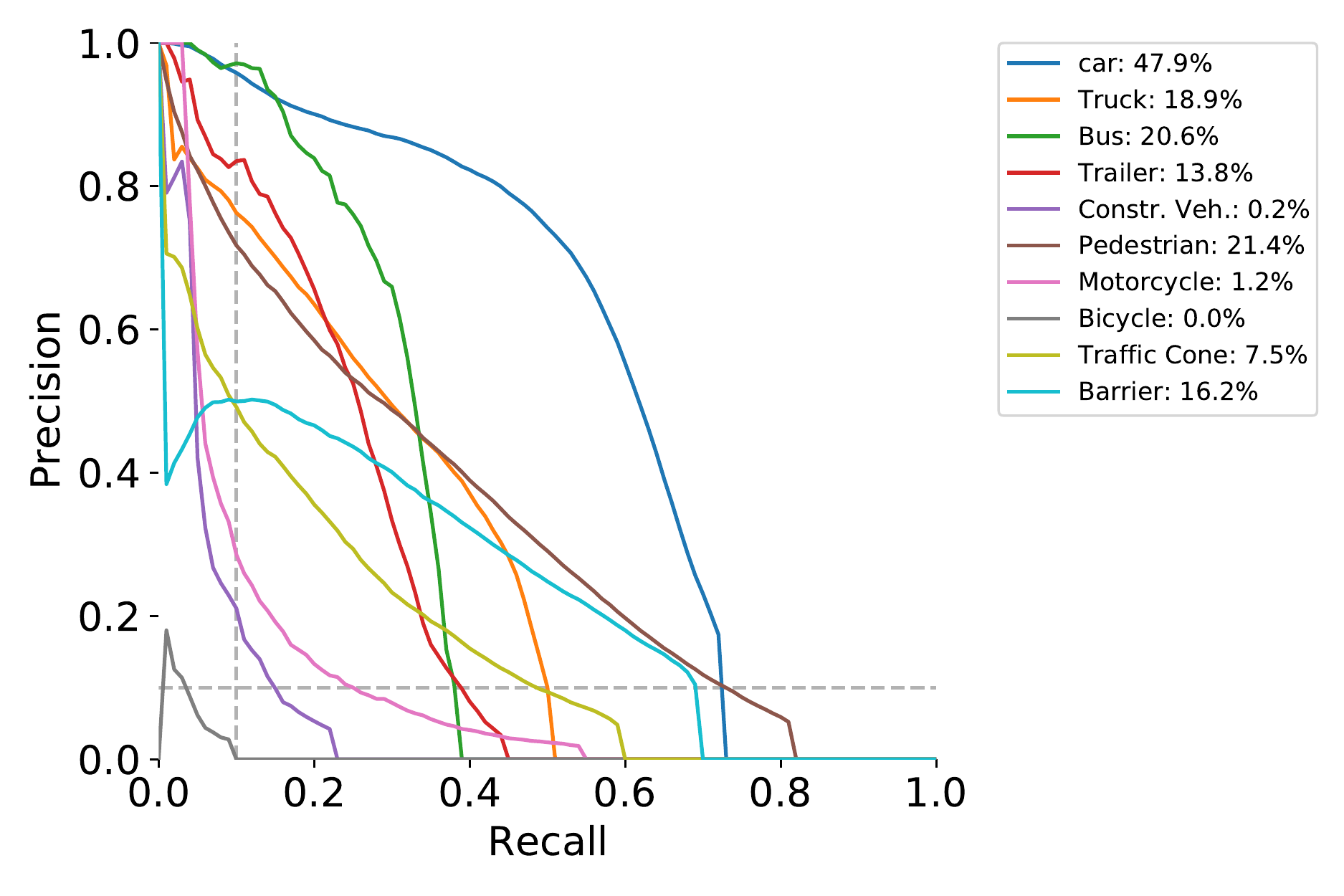}
				\\
				\includegraphics[width=1\columnwidth]{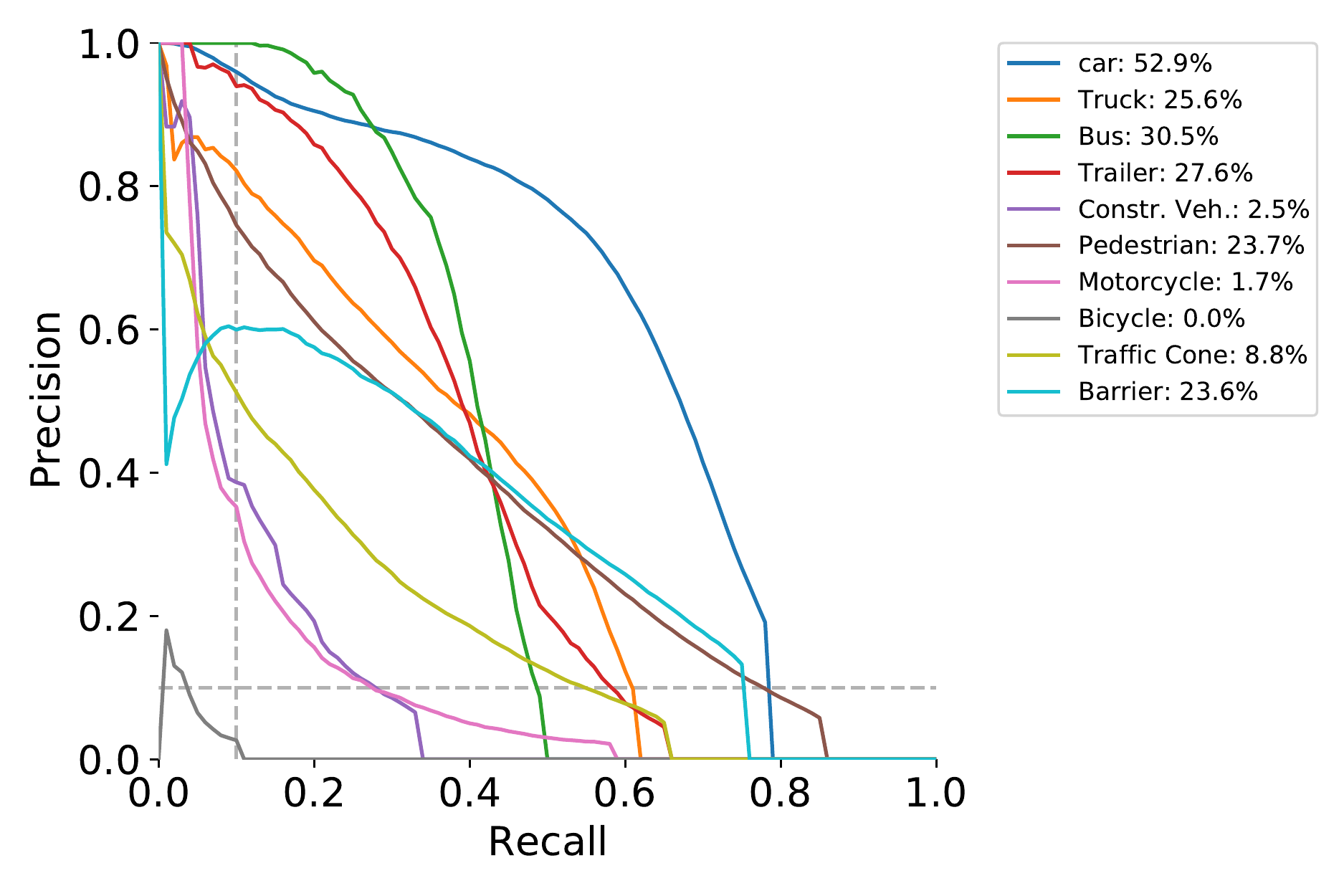}
				\\
				\includegraphics[width=1\columnwidth]{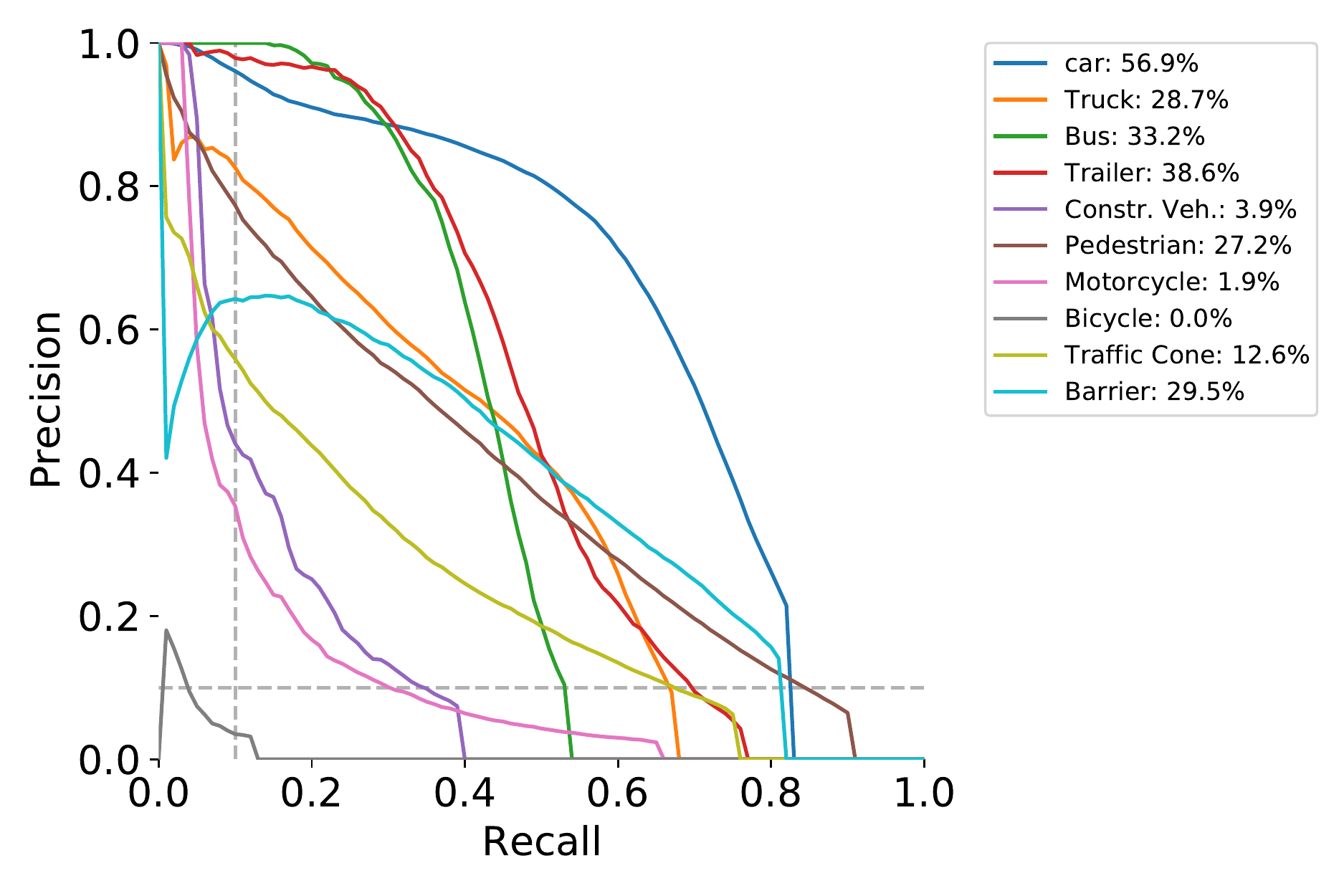}
			\end{minipage}
		}
		\subfigure[t] {  
			\begin{minipage}{0.236\textwidth}
				\centering
				\includegraphics[width=1\columnwidth]{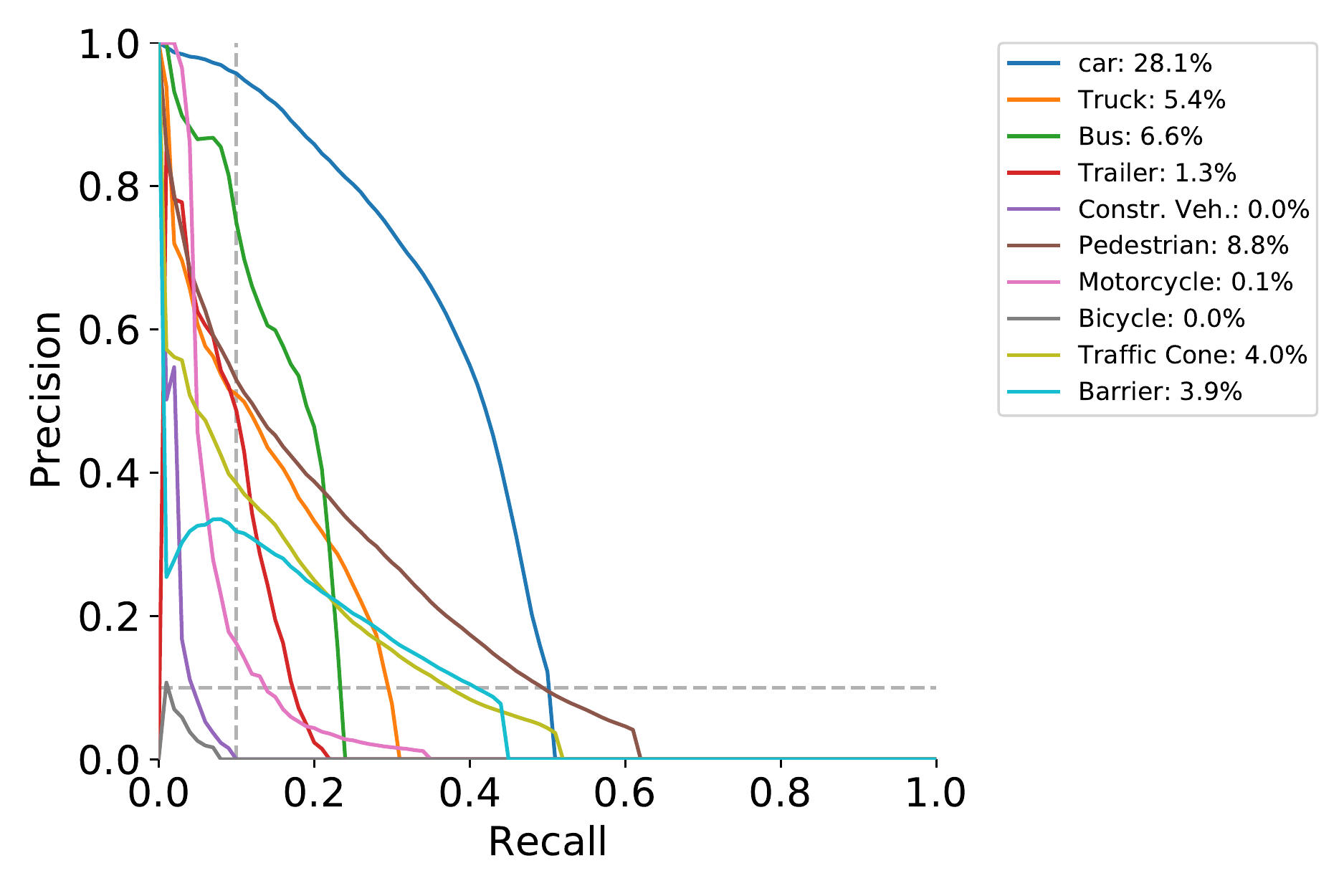}
				\\
				\includegraphics[width=1\columnwidth]{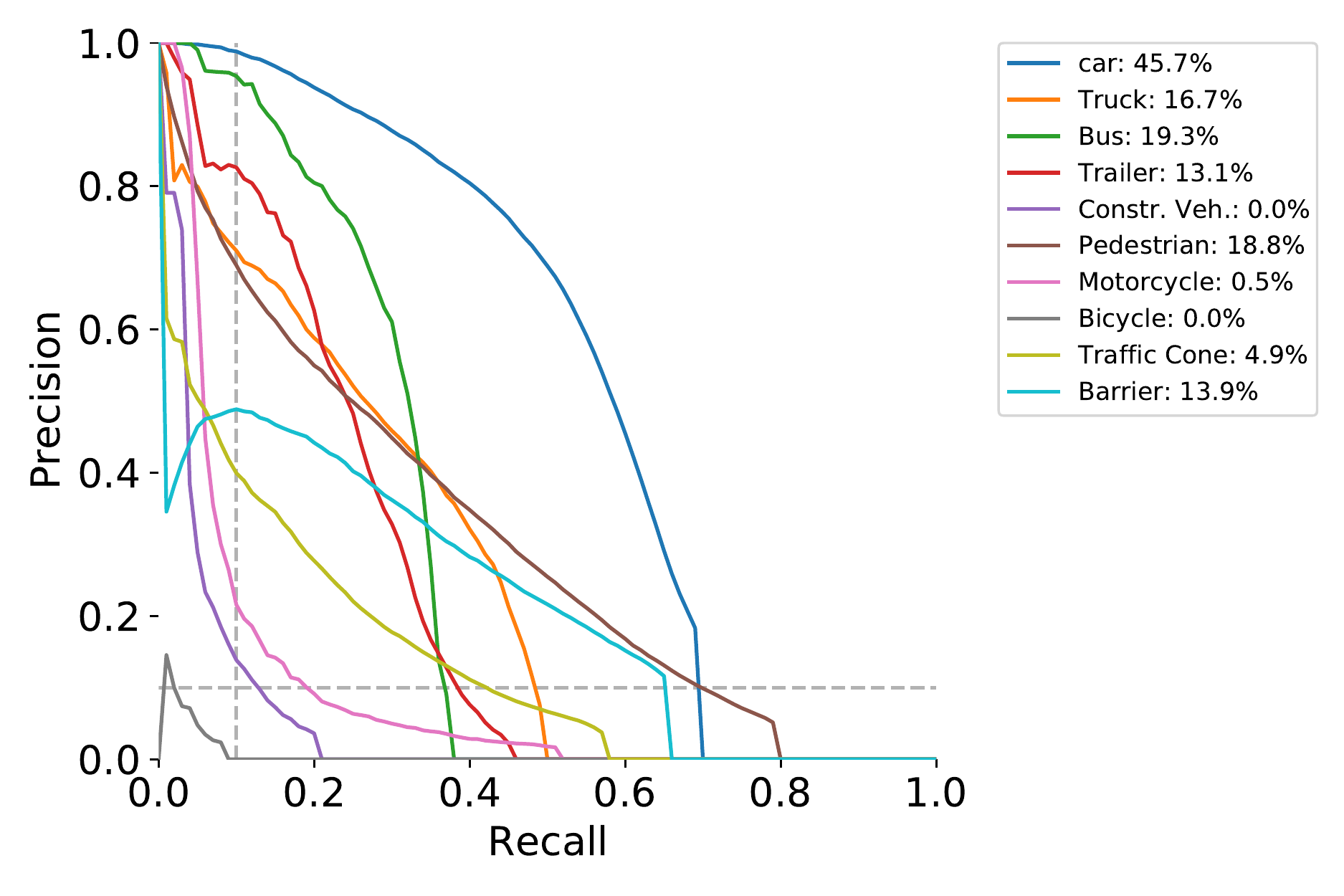}
				\\
				\includegraphics[width=1\columnwidth]{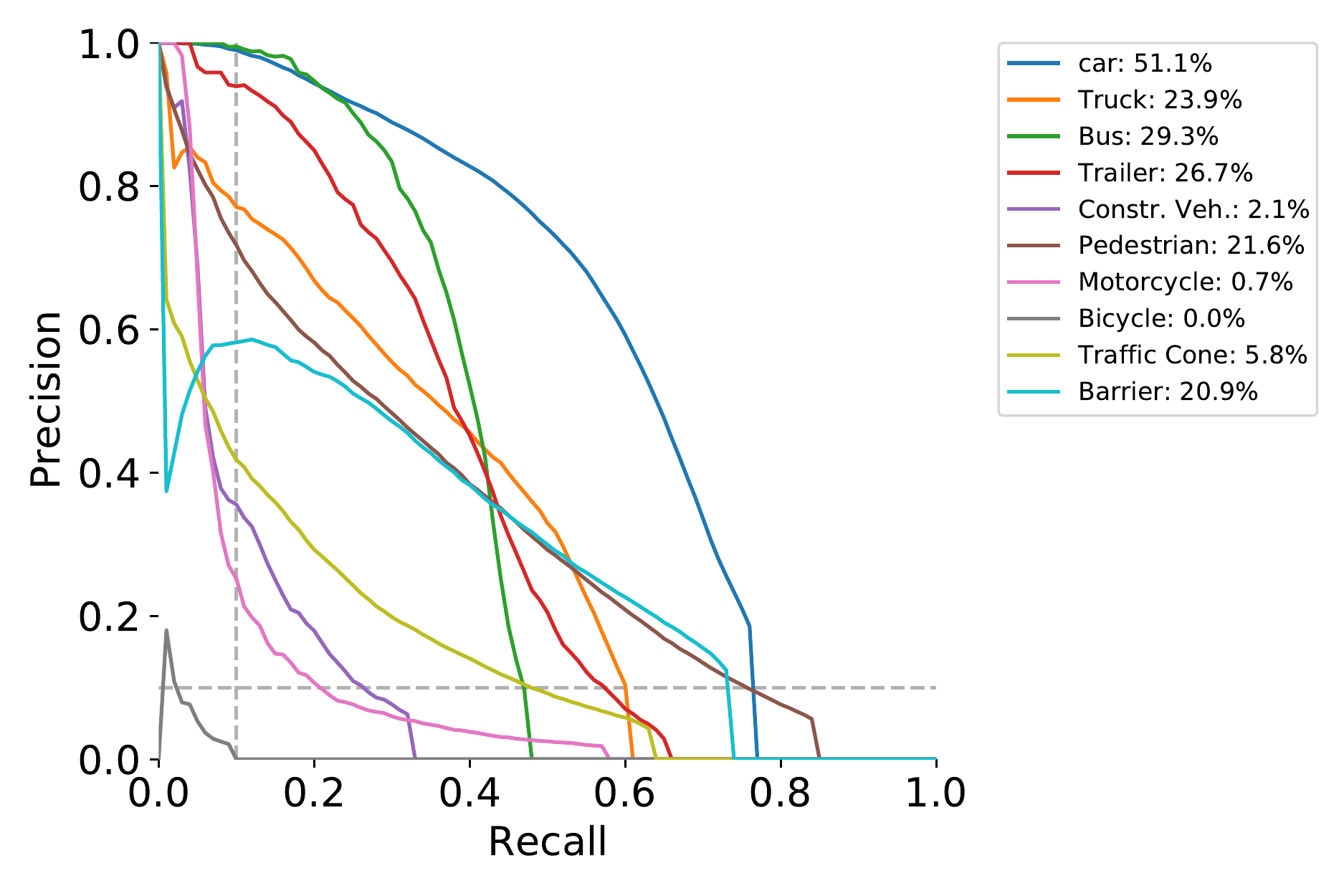}
				\\
				\includegraphics[width=1\columnwidth]{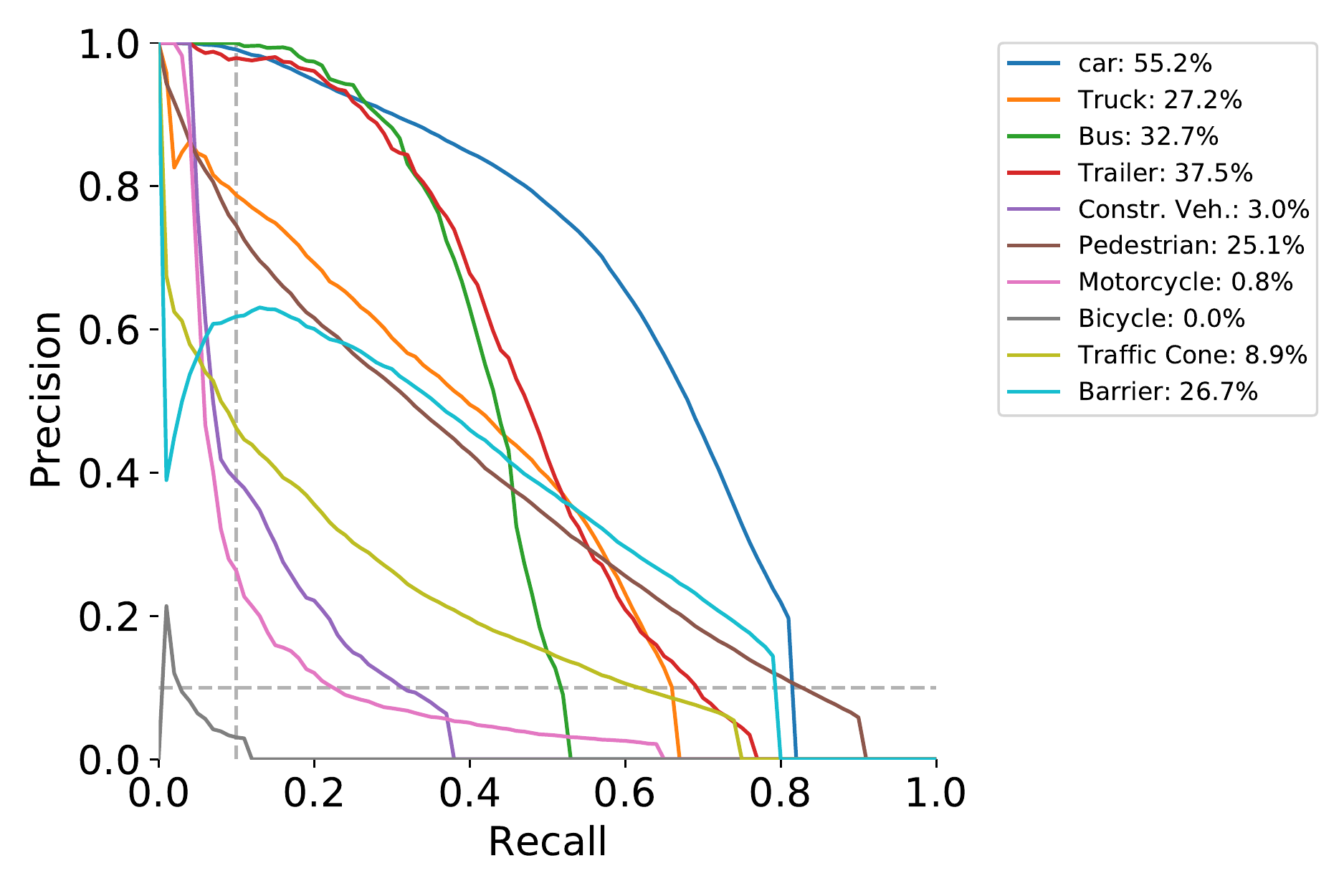}	
			\end{minipage}
		}
	\end{center}
	
	\caption{Per category precision-recall plot of \textbf{original detector}, attacking \textbf{translation} only, attacking \textbf{rotation} only, attacking \textbf{full trajectory} (black box) on the nuScenes validation set [2] (from left to right). From top to bottom is respectively the results from four thresholds (0.5m, 1.0m, 2.0m, 4.0m). From left to right, the curve shifts to the lower left when increasing the attack magnitude.}
	\label{fig:overall}
\end{figure*}

The nuScenes dataset [2] employs 2D center distance (0.5, 1, 2, 4 meters) as the matching threshold when calculating the Average Precision (AP). The per category precision-recall plots of the original detector as well as four attack settings are shown from Fig.~\ref{fig:black_original} - Fig.~\ref{fig:black_GNSS_INS}. The performance drop is the largest when the threshold is 0.5m (the highest precision 
standard), \eg, the AP@0.5m in car category is decreased by $23.1$($33.1\%$), $36.5$($52.4\%$), $41.5$($59.6\%$) while attacking translation, rotation and the full trajectory. As shown in Fig.~\ref{fig:overall}, the curve shifts to the lower left when increasing the attack magnitude.

\section{White Box Attack}\label{sec:white}
\subsection{Qualitative Evaluation}
\begin{figure*}[t]
	\includegraphics[width=1\textwidth]{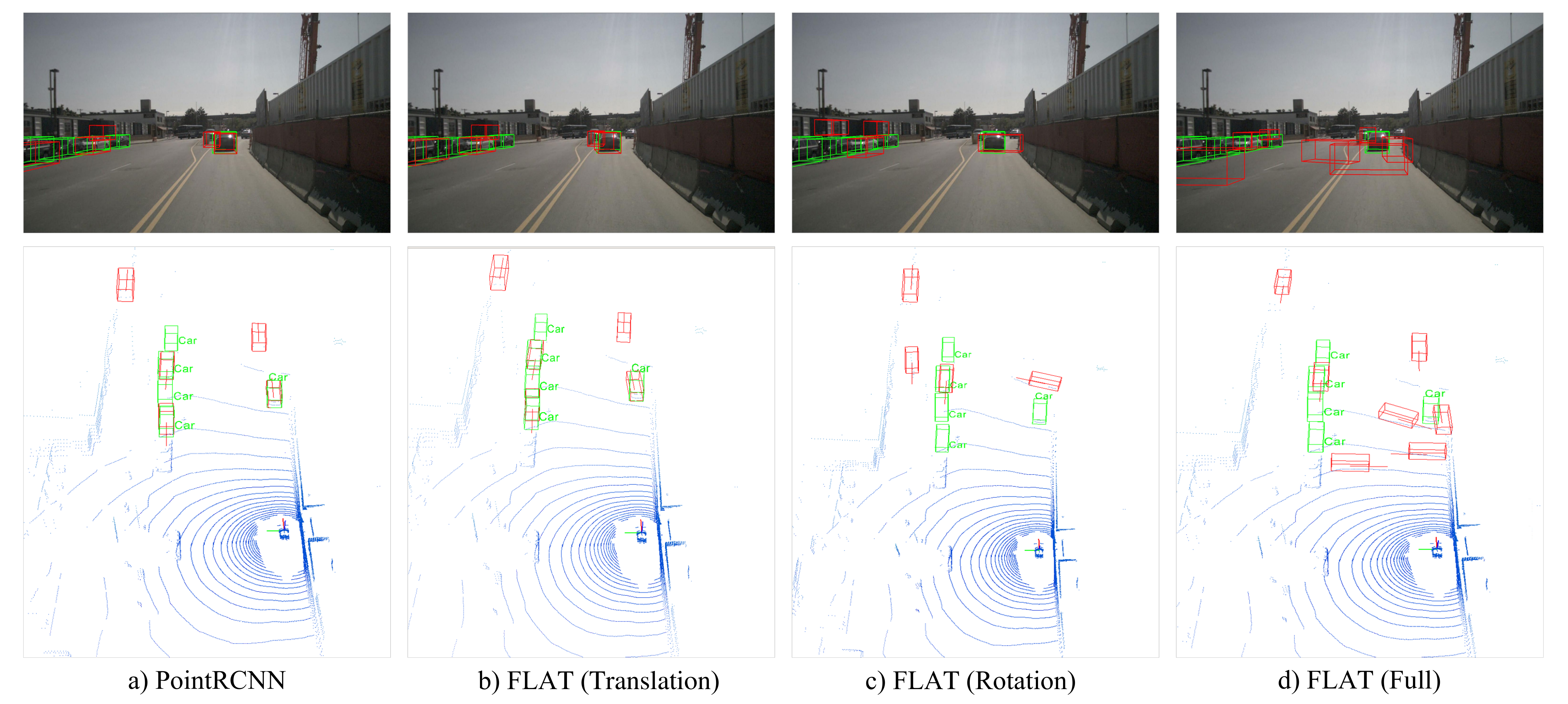}
	\caption{Qualitative evaluations of white box attack. False positives are increased and predictions are drifted by carefully crafting the vehicle trajectory.}
	\label{fig:white_supp1}
\end{figure*}

\begin{figure*}[t]
	\includegraphics[width=1\textwidth]{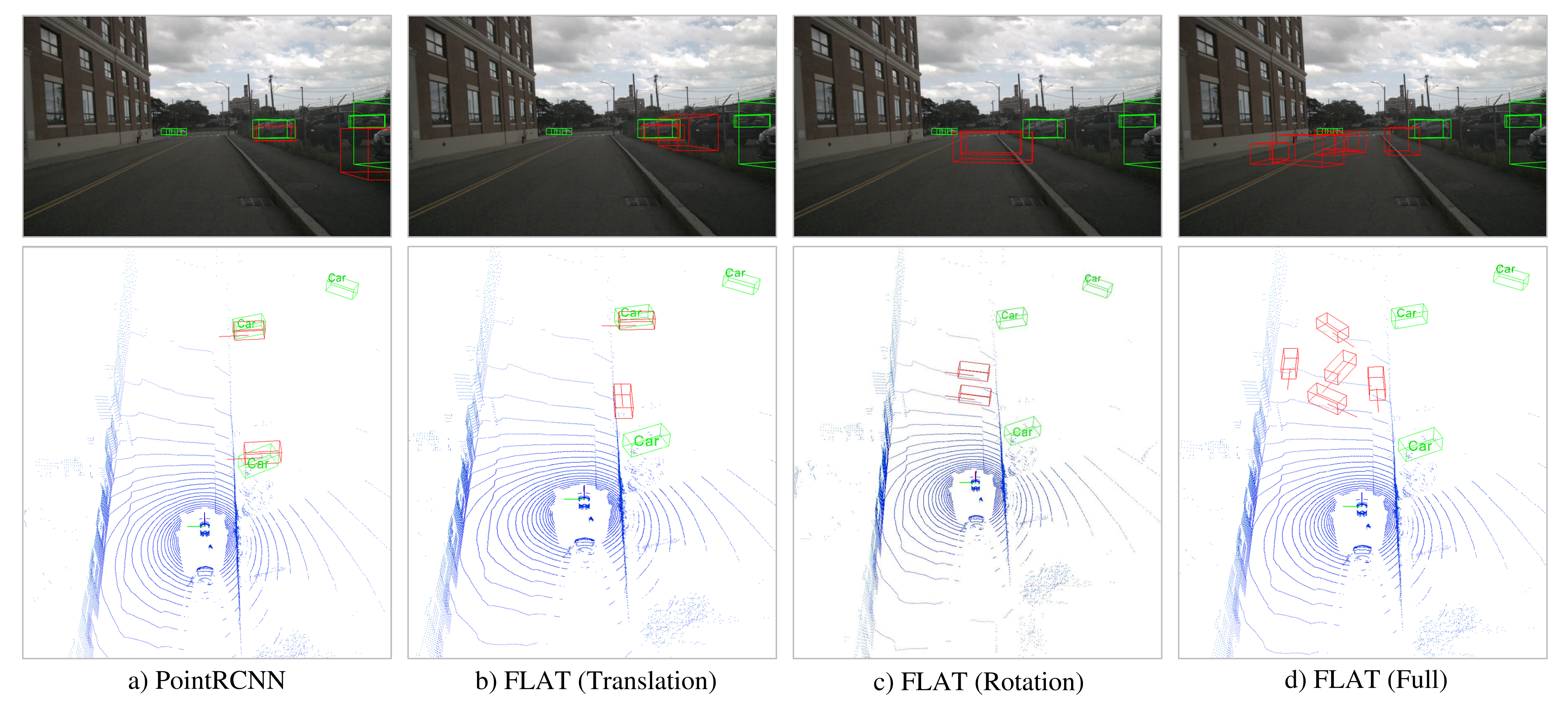}
	\caption{Qualitative evaluations of white box attack. False positives are increased and predictions are drifted by carefully crafting the vehicle trajectory.}
	\label{fig:white_supp2}
\end{figure*}

\begin{figure*}[t]
	\includegraphics[width=1\textwidth]{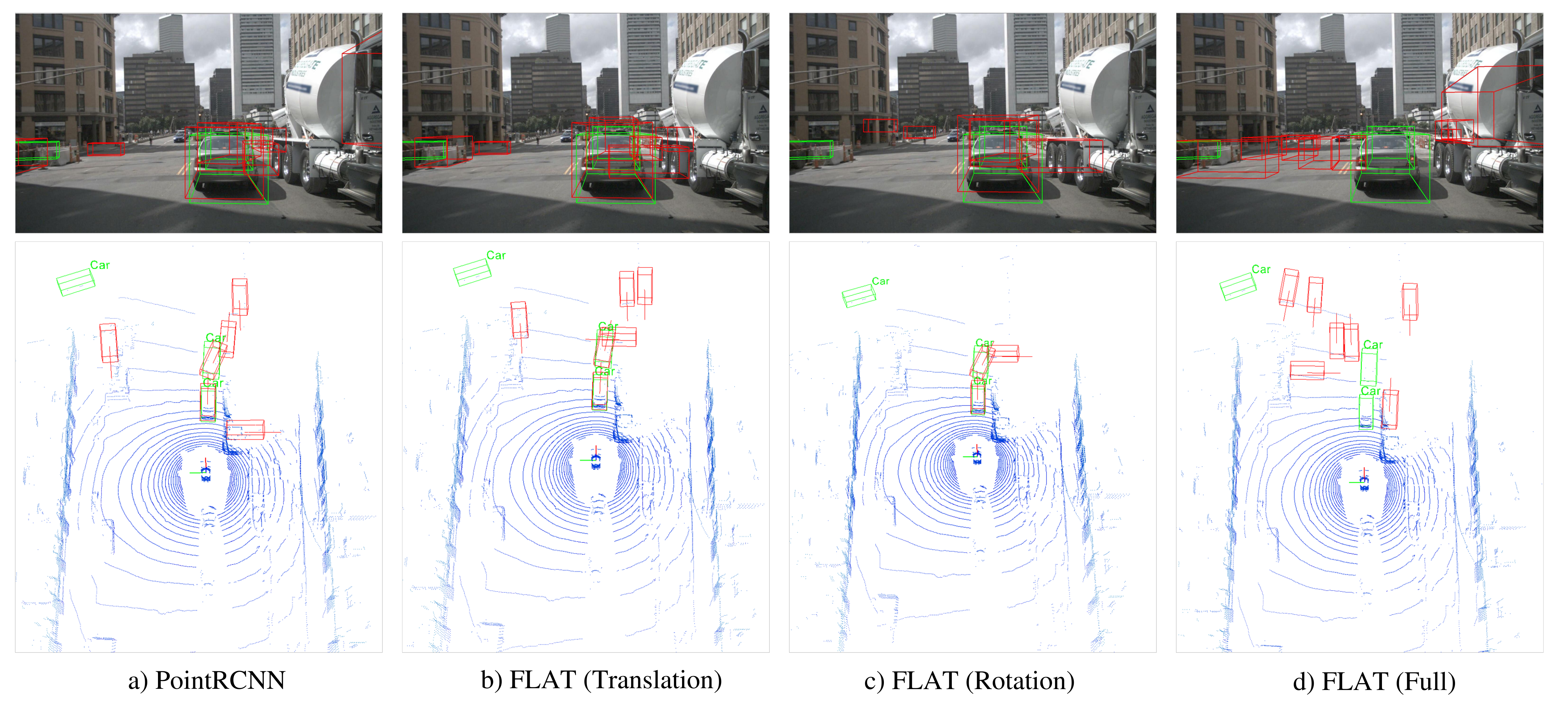}
	\caption{Qualitative evaluations of white box attack. False positives are increased and predictions are drifted by carefully crafting the vehicle trajectory.}
	\label{fig:white_supp3}
\end{figure*}

\begin{figure*}[t]
	\includegraphics[width=1\textwidth]{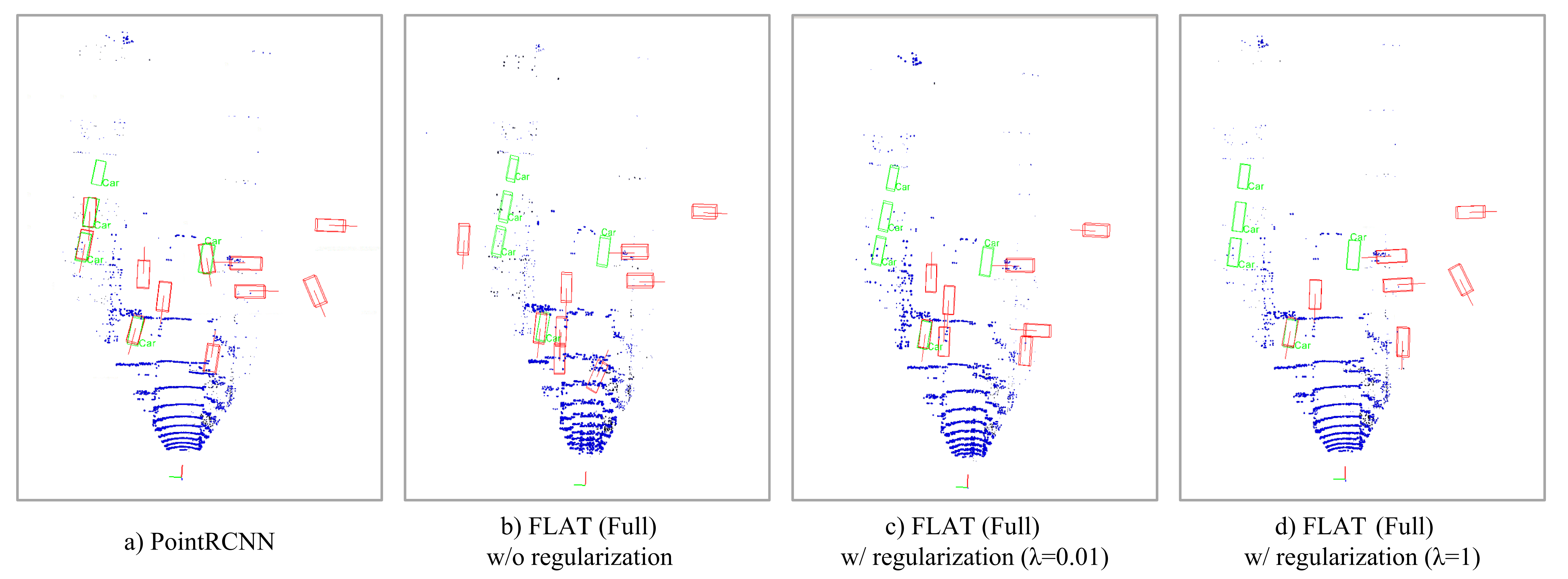}
	\caption{Qualitative example of FLAT attack with and without regularization. A minor perturbation in point cloud can fool the detector.}
	\label{fig:reg_supp1}
\end{figure*}

\begin{figure*}[t]
	\includegraphics[width=1\textwidth]{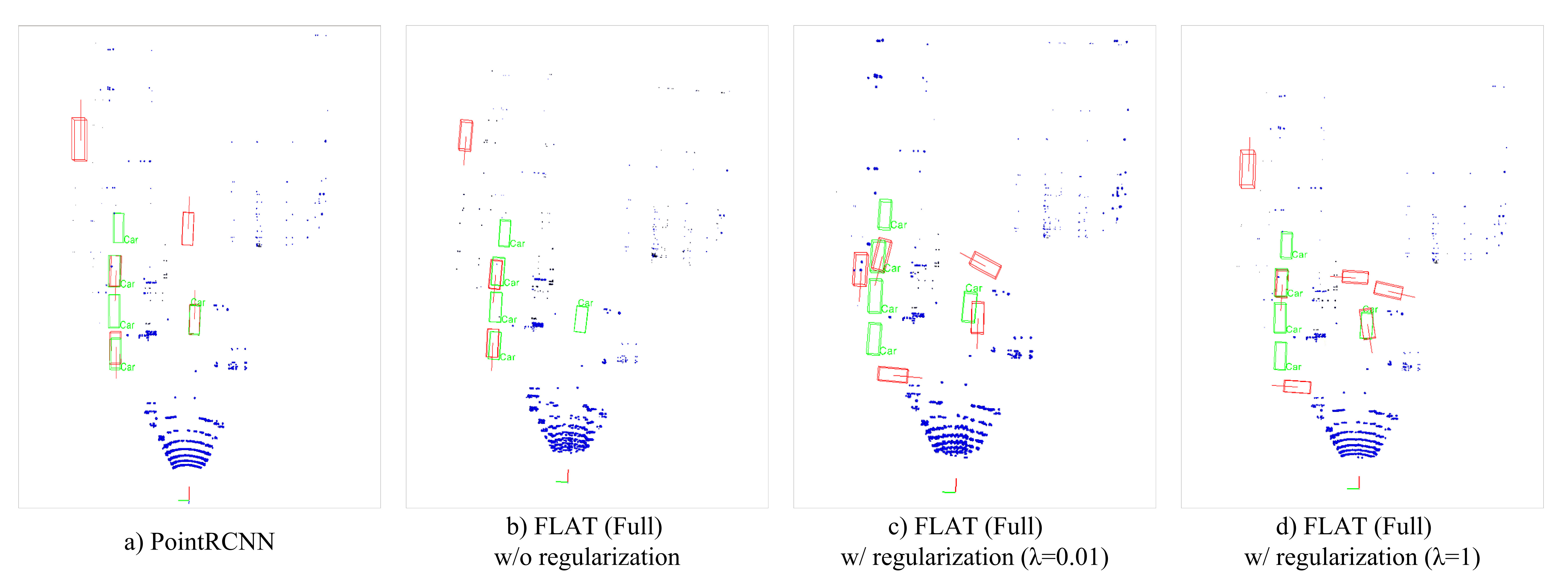}
	\caption{Qualitative example of FLAT attack with and without regularization. A minor perturbation in point cloud can fool the detector.}
	\label{fig:reg_supp2}
\end{figure*}

\begin{figure*}[t]
	\centering
	\includegraphics[width=0.94\textwidth]{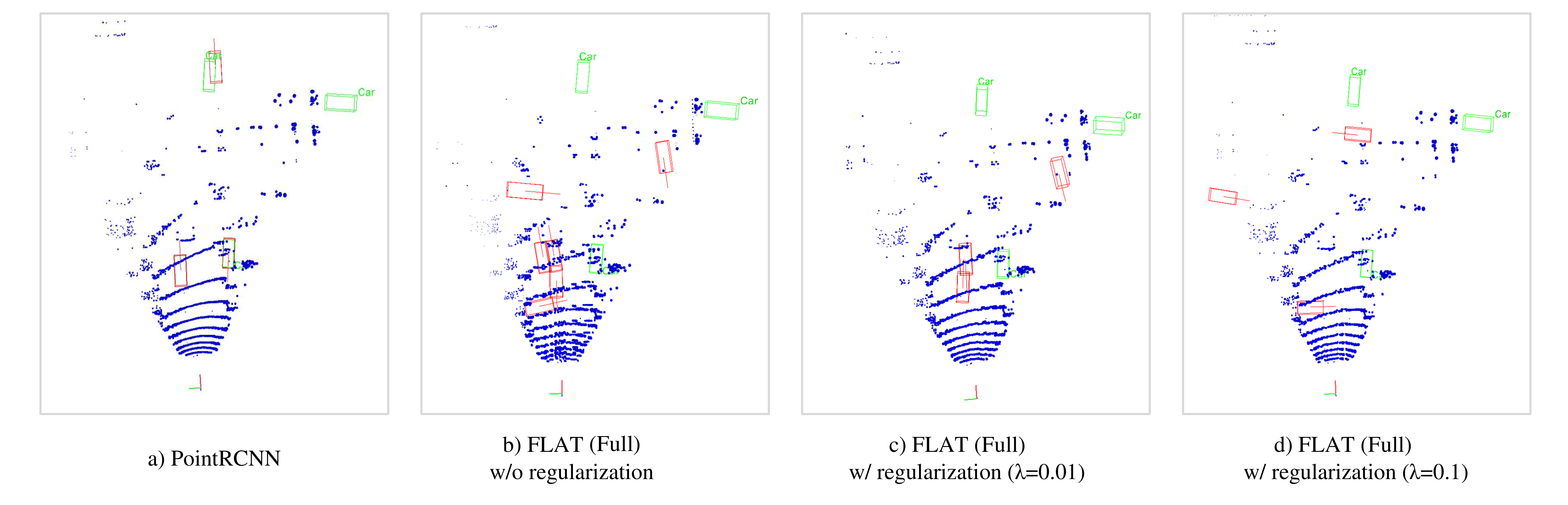}
	\caption{Qualitative example of FLAT attack with and without regularization. A minor perturbation in point cloud can fool the detector.}
	\label{fig:reg}
\end{figure*}

Three qualitative examples are demonstrated in Fig.~\ref{fig:white_supp1} - Fig.~\ref{fig:white_supp3}. In Fig.~\ref{fig:white_supp1}, original PointRCNN can perceive three our of five cars, and attacking translation only has slightly shifted the predictions. Attacking rotation has a stronger impact and cause two objects to be undetectable, while attacking the full trajectory further creates four false positives. In Fig.~\ref{fig:white_supp2}, PointRCNN can successfully detect two out of three surrounding cars, then adversarial translation perturbation make the detection of the nearest car drift. Both two detections are drifted in the scenario of adversarial rotation perturbation. Finally, when attacking both translation and rotation, five false positives have emerged, severely degrading the perception capability of the self-driving car. Similar results can be found in Fig.~\ref{fig:white_supp3}. Moreover, three qualitative examples of attacking with and without regularization are presented in Fig.~\ref{fig:reg_supp1}, Fig.~\ref{fig:reg_supp2} and Fig.~\ref{fig:reg}. As the strength of regularization is enlarged, the variation in the point space is reduced, enabling less perceptible attack. 
\subsection{Different Parameters}
The detection performances under different parameter settings are reported in Table~\ref{tab:trans-only} - Table~\ref{tab:trans-rot}. In the case of attacking the translation (fooling classification in stage-2), AP of easy case in six parameter settings fluctuate between $11.72$ and $13.89$, which means that all the parameter configurations can produce a high-quality adversarial perturbation. In the situation of attacking the full trajectory (with the aim of attacking regression branch in stage-1), AP of easy case in six parameter settings fluctuate between $0.53$ and $2.44$, proving that all the six settings can maintain the attacking quality at a high level. To sum up, the number of iteration and the step size for each iteration has a relatively small impact on our FLAT attack. As for the interpolation step, when it is increased, there will be more sectors, \ie, more point cloud groups, and the number of trajectory points which can be attacked are also increased, easing the learning of an effective adversarial perturbation. Consequently, when the interpolation step is increased from 50 to 1000, the AP in easy scenarios can be lowered by $5.04$ ($36.7\%$) while attacking the translation, by $3.70$ ($78.2\%$) while attacking the rotation, and by $0.51$ ($87.9\%$) while attacking the full trajectory, as shown in  Table~\ref{tab:interpolation_step}. 

Noted that our simulation of motion distortion/correction accurately matches with the real-world scenarios, \eg, in Fig.~2 of [29]: ``the Velodyne VLP-16 software used in our electric vehicle platform produces 76 packets for each full revolution scan. Each packet covers an azimuth angle of approximately $4.74^\circ$.'' Noted that the number of point cloud sector/packet is adjustable depending on vehicle platforms, and the attack performance keeps at a high level under different number of sector/interpolation step, as shown in  Table~\ref{tab:interpolation_step}.

\section{Polynomial Trajectory Perturbation}\label{sec:polynomial}
\begin{figure*}[t]
	\centering
	\includegraphics[width=0.94\textwidth]{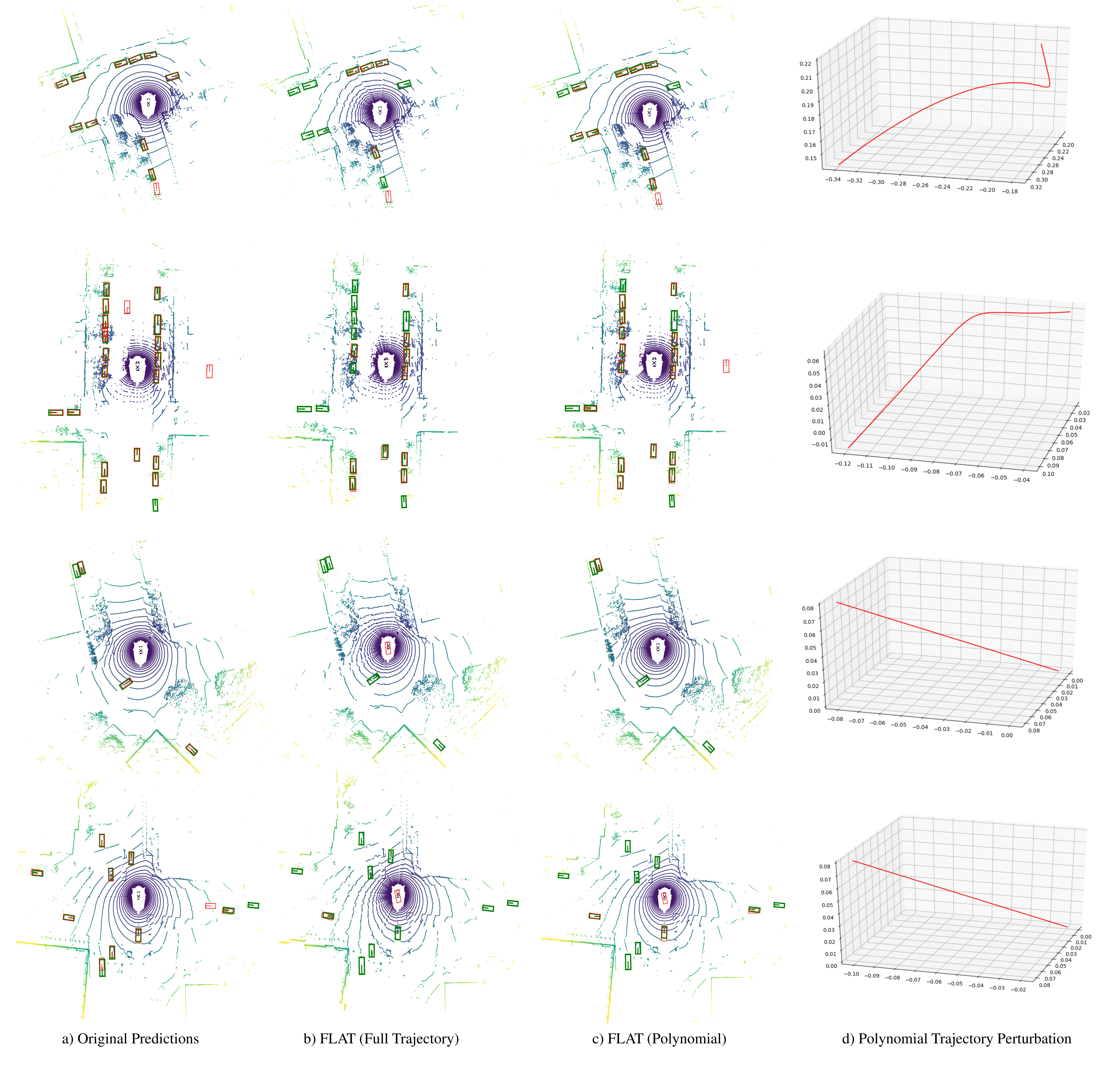}
	\caption{Point cloud visualization and qualitative results of black box attack. a) Raw detections of the original detector PointPillar++ [11]. b) The output of the detector after attacking the full trajectory. c) The output of the detector after polynomial trajectory perturbation in the euclidean space. d) The polynomial translation perturbation visualized in $xyz$ space, the units of three axes are all meters.}
	\label{fig:traj_supp}
\end{figure*}

Our trajectory attack is feasible, \eg, through GNSS spoofing as proven in [23]: ``Today, it is feasible to execute GNSS spoofing attacks with less than $\$100$ of equipment. GNSS signal generators can be programmed to transmit radio frequency signals corresponding to a static position, or simulate entire trajectories.'' We have verified the effectiveness of the discrete adversarial trajectory perturbation in both white box and black box attack. To achieve a temporally-smooth attack which is less perceptible, we implement a polynomial regression before the generation of perturbation and attack the polynomial coefficients instead of the trajectory itself. In this scenario, we only need to manipulate several key points to bend a polynomial-parameterized trajectory which can be easily achieved in reality, realizing a real-time and high-quality attack. Several qualitative examples of the polynomial trajectory perturbation are shown in Fig.~\ref{fig:traj_supp}. Although the attack performance is inferior to the full trajectory attack, the detector still missed many safety-critical objects yet the perturbation in point cloud space is highly imperceptible: \textit{even for human eyes, it is quite difficult to distinguish the point cloud before and after the polynomial trajectory perturbation. }

%For the experiment 1), the results are presented in Table~\ref{tab:polynomial_after_attack}, the attack quality is maintained at a satisfactory level after the implementation of polynomial interpolation. Besides, in the three settings (degree of polynomial equals to 3, 5, 7), the performances have little fluctuations, and compared to discrete perturbation before interpolation, the loss of performance is acceptable (from 14.8 to 16.9). Meanwhile, the smoothness becomes much better compared to the discrete settings (as shown in Fig.~\ref{fig:polynomial_after_attack}), further increasing the defense difficulty. As for the experiment 2), attacking polynomial coefficients is on par with attacking each trajectory point, as shown in Table~\ref{tab:polynomial_before_attack}, validating that a more advanced attack with good trajectory smoothness can be realized using polynomial interpolation.

\section{Summary}\label{sec:summary}
In this supplementary, we presented more qualitative evaluations of both white box and black box attack, to validate the effectiveness of our attack pipeline. Besides, we found that the number of PGD iteration and the step size for each iteration have a relatively small impact on the attack quality, but the interpolation step, \ie, the number of LiDAR packets, can have a relatively large influence on the attack performance, because increasing trajectory points which can be deliberately 
modified can facilitate the adversarial learning, resulting in a stronger adversarial perturbation. Finally, more qualitative examples of the polynomial trajectory perturbation are demonstrated to validate the imperceptibility of out attack.

\begin{table*}[t]
	\setlength{\tabcolsep}{2.5mm}
	\begin{center}
		\caption{3D detection performance of white-box \textbf{translation} attack. We report the average precision of the 3D bounding box for the car category with the IoU threshold 0.7, under different levels of difficulty and ranges of depth following [28]. iter\_eps and nb\_iter respectively denote the step size of each iteration and the number of iteration. The best and second best attack qualities are respectively highlighted using \tbred{red} and \tbblue{blue} color.
		}\label{tab:trans-only}
		\begin{tabular}{c|c|cc|ccc|ccc}
			\hline
			\multicolumn{2}{c|}{Attack Setting \quad  $\backslash$ \quad  Case} & iter\_eps & nb\_iter & Easy & Moderate & Hard & 0-30m & 30-50m & 50-70m \\ \hline
			\multicolumn{2}{c|}{None} & - & - & 47.44 & 21.56 & 20.91 & 47.44 & 2.16 & 0.17 \\ 
			\multicolumn{2}{c|}{Coordinate Attack} & - & - & 16.42 & 6.58 & 5.90 & 15.20 & 0.48 & 0.03 \\ 
			\multicolumn{2}{c|}{Random Attack} & - & - & 17.00 & 8.58 & 8.90 & 20.43 & 1.09 & 0.09 \\ \hline
			\multirow{12}{*}{Classification} & \multirow{6}{*}{Stage-1} & 0.05 & 10 & 15.38 & 7.32 & 7.66 & 18.12 & 0.93 & \tbred{0.03} \\
			&  & 0.05 & 20 & 16.85 & 8.49 & 8.69 & 19.87 & 1.29 & 0.05 \\
			&  & 0.05 & 30 & 17.60 & 8.81 & 8.95 & 20.60 & 0.82 & 0.05 \\
			&  & 0.1 & 10 & 13.12 & 6.49 & 6.79 & 15.83 & \tbblue{0.71} & \tbblue{0.04} \\
			&  & 0.1 & 20 & 12.94 & 6.58 & 7.22 & 16.82 & 0.86 & 0.06 \\
			&  & 0.1 & 30 & 14.77 & 7.14 & 7.69 & 18.57 & 0.74 & 0.05 \\ \cline{2-10} 
			& \multirow{6}{*}{Stage-2} & 0.05 & 10 & 12.47 & 6.72 & 7.29 & 16.13 & 1.16 & 0.14 \\
			&  & 0.05 & 20 & 13.89 & 6.82 & 6.96 & 16.11 & 0.95 & 0.17 \\
			&  & 0.05 & 30 & \tbblue{11.87} & \tbblue{5.98} & \tbblue{6.40} & \tbblue{14.87} & \tbred{0.67} & 0.05 \\
			&  & 0.1 & 10 & 12.58 & 6.40 & 6.77 & 15.25 & 0.95 & 0.07 \\
			&  & 0.1 & 20 & \tbred{11.72} & \tbred{5.87} & \tbred{6.06} & \tbred{13.91} & 0.87 & \tbblue{0.04} \\
			&  & 0.1 & 30 & 12.54 & 6.52 & 6.69 & 15.14 & 0.96 & 0.09 \\ \hline
			\multirow{12}{*}{Regression} & \multirow{6}{*}{Stage-1} & 0.05 & 10 & 17.45 & 8.43 & 8.42 & 19.53 & 1.12 & \tbblue{0.04} \\
			&  & 0.05 & 20 & 16.52 & 8.53 & 8.75 & 19.69 & 1.37 & \tbblue{0.04} \\
			&  & 0.05 & 30 & 17.60 & 8.81 & 9.11 & 21.19 & 0.85 & \tbblue{0.04} \\
			&  & 0.1 & 10 & 14.47 & 7.51 & 7.88 & 17.77 & 1.07 & 0.05 \\
			&  & 0.1 & 20 & 17.46 & 8.24 & 8.57 & 19.36 & 1.09 & \tbred{0.03} \\
			&  & 0.1 & 30 & 14.42 & 7.04 & 7.31 & 16.44 & 1.02 & 0.13 \\ \cline{2-10}
			& \multirow{6}{*}{Stage-2} & 0.05 & 10 & 26.07 & 12.76 & 12.51 & 27.19 & 2.16 & 0.17 \\
			&  & 0.05 & 20 & 26.05 & 12.75 & 12.50 & 27.19 & 2.16 & 0.15 \\
			&  & 0.05 & 30 & 26.07 & 12.76 & 12.51 & 27.19 & 2.16 & 0.17 \\
			&  & 0.1 & 10 & 26.07 & 12.76 & 12.51 & 27.19 & 2.16 & 0.17 \\
			&  & 0.1 & 20 & 26.09 & 12.78 & 12.53 & 27.15 & 2.17 & 0.17 \\
			&  & 0.1 & 30 & 26.07 & 12.76 & 12.51 & 27.19 & 2.16 & 0.17 \\ \hline
		\end{tabular}
	\end{center}
\end{table*}

\begin{table*}[t]
	\setlength{\tabcolsep}{2.5mm}
	\begin{center}
		\caption{3D object detection performance of white-box \textbf{rotation} attack. Other settings are similar to Table~\ref{tab:trans-only}.
		}\label{tab:rot-only}
		\begin{tabular}{c|c|cc|ccc|ccc}
			\hline
			\multicolumn{2}{c|}{Attack Setting \quad  $\backslash$ \quad  Case} & iter\_eps & nb\_iter & Easy & Moderate & Hard & 0-30m & 30-50m & 50-70m \\ \hline
			\multicolumn{2}{c|}{None} & - & - & 47.44 & 21.56 & 20.91 & 47.44 & 2.16 & 0.17 \\ 
			\multicolumn{2}{c|}{Random Attack} & - & - & 12.30 & 4.87 & 5.13 & 13.31 & \tbblue{0.01} & \tbred{0.00} \\ \hline
			\multirow{12}{*}{Classification} & \multirow{6}{*}{Stage-1} & 0.005 & 10 & 5.38 & 1.91 & 2.05 & 5.89 & \tbblue{0.01} & \tbred{0.00} \\
			&  & 0.005 & 20 & 7.16 & 2.94 & 2.81 & 7.75 & 0.02 & \tbred{0.00} \\
			&  & 0.005 & 30 & 6.17 & 2.41 & 2.49 & 7.02 & \tbblue{0.01} & \tbred{0.00} \\
			&  & 0.01 & 10 & 3.93 & 1.70 & 1.61 & 4.79 & \tbblue{0.01} & \tbred{0.00} \\
			&  & 0.01 & 20 & 6.32 & 2.43 & 2.51 & 7.32 & 0.02 & \tbred{0.00} \\
			&  & 0.01 & 30 & 6.83 & 2.53 & 2.27 & 6.87 & 0.02 & \tbred{0.00} \\ \cline{2-10} 
			& \multirow{6}{*}{Stage-2} & 0.005 & 10 & 4.32 & 1.48 & 1.35 & 4.08 & \tbred{0.00} & \tbred{0.00} \\
			&  & 0.005 & 20 & 3.39 & 1.20 & 1.14 & 3.39 & \tbblue{0.01} & \tbred{0.00} \\
			&  & 0.005 & 30 & 6.59 & 2.22 & 2.04 & 6.15 & \tbblue{0.01} & \tbred{0.00} \\
			&  & 0.01 & 10 & 3.18 & 1.20 & 1.17 & 3.44 & \tbred{0.00} & \tbred{0.00} \\
			&  & 0.01 & 20 & \tbred{2.35} & \tbred{0.80} & \tbred{0.61} & \tbred{2.03} & \tbblue{0.01} & \tbred{0.00} \\
			&  & 0.01 & 30 & \tbblue{2.36} & \tbblue{0.90} & \tbblue{0.94} & \tbblue{2.84} & \tbblue{0.01} & \tbred{0.00} \\ \hline
			\multirow{12}{*}{Regression} & \multirow{6}{*}{Stage-1} & 0.005 & 10 & 8.46 & 3.56 & 3.36 & 9.01 & 0.04 & \tbred{0.00} \\
			&  & 0.005 & 20 & 8.11 & 3.11 & 3.06 & 7.96 & 0.02 & 0.03 \\
			&  & 0.005 & 30 & 8.03 & 3.25 & 3.02 & 8.27 & \tbblue{0.01} & \tbred{0.00} \\
			&  & 0.01 & 10 & 5.02 & 1.79 & 1.61 & 4.79 & \tbblue{0.01} & 0.01 \\
			&  & 0.01 & 20 & 5.50 & 1.87 & 1.76 & 5.45 & 0.02 & \tbred{0.00} \\
			&  & 0.01 & 30 & 4.17 & 1.53 & 1.47 & 4.60 & \tbblue{0.01} & \tbred{0.00} \\ \cline{2-10} 
			& \multirow{6}{*}{Stage-2} & 0.005 & 10 & 25.99 & 12.71 & 12.49 & 27.07 & 2.16 & 0.17 \\
			&  & 0.005 & 20 & 26.07 & 12.76 & 12.51 & 27.19 & 2.16 & 0.17 \\
			&  & 0.005 & 30 & 26.07 & 12.76 & 12.51 & 27.19 & 2.16 & 0.17 \\
			&  & 0.01 & 10 & 26.07 & 12.76 & 12.51 & 27.19 & 2.16 & 0.17 \\
			&  & 0.01 & 20 & 26.30 & 12.89 & 12.59 & 27.35 & 2.17 & 0.17 \\
			&  & 0.01 & 30 & 26.18 & 12.75 & 12.53 & 27.24 & 2.16 & 0.17 \\ \hline
		\end{tabular}
	\end{center}
\end{table*}

\begin{table*}[t]
	\setlength{\tabcolsep}{1.5mm}
	\begin{center}
		\caption{3D detection performance of white-box \textbf{full trajectory} attack. Other settings are similar to Table~\ref{tab:trans-only}. 
		}
		\label{tab:trans-rot}
		\begin{tabular}{c|c|ccc|ccc|ccc}
			\hline
			\multicolumn{2}{c|}{Attack Setting \quad  $\backslash$ \quad  Case} & iter\_eps & iter\_eps2 & nb\_iter & Easy & Moderate & Hard & 0-30m & 30-50m & 50-70m \\ \hline
			\multicolumn{2}{c|}{None} & - & - & - & 47.44 & 21.56 & 20.91 & 47.44 & 2.16 & \tbblue{0.17} \\
			\multicolumn{2}{c|}{Random Attack} & - & - & - & 5.66 & 2.43 & 2.78 & 7.67 & 0.02 & \tbred{0.00} \\ \hline
			\multirow{12}{*}{Classification} & \multirow{6}{*}{Stage-1} & 0.005 & 0.05 & 10 & 1.11 & 0.25 & 0.23 & 1.19 & 0.02 & \tbred{0.00} \\
			&  & 0.005 & 0.05 & 20 & 2.89 & 1.02 & 1.15 & 3.52 & \tbblue{0.01} & \tbred{0.00} \\
			&  & 0.005 & 0.05 & 30 & 0.65 & 0.21 & 0.25 & 0.88 & \tbblue{0.01} & \tbred{0.00} \\
			&  & 0.01 & 0.1 & 10 & 0.57 & 0.16 & 0.16 & 0.67 & \tbblue{0.01} & \tbred{0.00} \\
			&  & 0.01 & 0.1 & 20 & 1.52 & 0.45 & 0.51 & 1.71 & \tbblue{0.01} & \tbred{0.00} \\
			&  & 0.01 & 0.1 & 30 & 0.84 & 0.20 & 0.24 & 1.03 & \tbred{0.00} & \tbred{0.00} \\ \cline{2-11} 
			& \multirow{6}{*}{Stage-2} & 0.005 & 0.05 & 10 & 0.37 & 0.09 & 0.07 & 0.43 & \tbred{0.00} & \tbred{0.00} \\
			&  & 0.005 & 0.05 & 20 & 0.42 & 0.15 & \tbblue{0.12} & 0.56 & \tbred{0.00} & \tbred{0.00} \\
			&  & 0.005 & 0.05 & 30 & \tbred{0.09} & \tbred{0.01} & \tbred{0.02} & \tbred{0.12} & \tbred{0.00} & \tbred{0.00} \\
			&  & 0.01 & 0.1 & 10 & 0.22 & \tbblue{0.02} & \tbred{0.02} & \tbblue{0.21} & \tbred{0.00} & \tbred{0.00} \\
			&  & 0.01 & 0.1 & 20 & \tbblue{0.19} & \tbred{0.01} & \tbred{0.02} & 0.26 & \tbred{0.00} & \tbred{0.00} \\
			&  & 0.01 & 0.1 & 30 & 0.54 & 0.15 & 0.15 & 0.54 & 0.03 & \tbred{0.00} \\ \hline
			\multirow{12}{*}{Regression} & \multirow{6}{*}{Stage-1} & 0.005 & 0.05 & 10 & 1.20 & 0.33 & 0.32 & 1.27 & \tbblue{0.01} & \tbred{0.00} \\
			&  & 0.005 & 0.05 & 20 & 2.44 & 0.90 & 0.88 & 2.72 & 0.02 & \tbred{0.00} \\
			&  & 0.005 & 0.05 & 30 & 2.07 & 0.77 & 0.76 & 2.68 & \tbblue{0.01} & \tbred{0.00} \\
			&  & 0.01 & 0.1 & 10 & 0.90 & 0.36 & 0.35 & 1.18 & \tbblue{0.01} & \tbred{0.00} \\
			&  & 0.01 & 0.1 & 20 & 1.01 & 0.35 & 0.32 & 1.27 & \tbblue{0.01} & \tbred{0.00} \\
			&  & 0.01 & 0.1 & 30 & 0.53 & 0.11 & 0.13 & 0.80 & 0.03 & \tbred{0.00} \\ \cline{2-11} 
			& \multirow{6}{*}{Stage-2} & 0.005 & 0.05 & 10 & 26.07 & 12.76 & 12.51 & 27.19 & 2.16 & \tbblue{0.17} \\
			&  & 0.005 & 0.05 & 20 & 26.08 & 12.76 & 12.52 & 27.19 & 2.17 & \tbblue{0.17} \\
			&  & 0.005 & 0.05 & 30 & 26.06 & 12.75 & 12.51 & 27.17 & 2.16 & \tbblue{0.17} \\
			&  & 0.01 & 0.1 & 10 & 26.07 & 12.76 & 12.51 & 27.19 & 2.16 & \tbblue{0.17} \\
			&  & 0.01 & 0.1 & 20 & 26.03 & 12.70 & 12.48 & 27.13 & 2.16 & \tbblue{0.17} \\
			&  & 0.01 & 0.1 & 30 & 26.07 & 12.76 & 12.51 & 27.19 & 2.16 & \tbblue{0.17} \\ \hline
		\end{tabular}
	\end{center}
\end{table*}

% Please add the following required packages to your document preamble:
% \usepackage{multirow}
\begin{table*}[t]
	\begin{center}
		\caption{3D object detection performance under different interpolation steps (attacking the \textbf{classification} in \textbf{stage-2}). The number of iteration and the step size for each iteration is fixed as 20 and 0.1/0.01 (translation/rotation), respectively.
		}
		\label{tab:interpolation_step}
		\begin{tabular}{c|c|ccc|ccc}
			\hline
			Attack Setting                    & Interpolation Step & Easy & Moderate & Hard& 0-30m & 30-50m & 50-70m \\ \hline
			\multirow{5}{*}{\textbf{FLAT (Translation)}}      & 25                 & 16.97 & 7.71 & 7.99 & 18.12 & 0.93 & 0.13\\
			& 50                 & 13.73 & 4.75 & 6.33 & 16.33 & 0.34 & 0.00 \\
			& 100                &   11.72 & 5.87 & 6.06 & 13.91 & 0.87 & 0.04  \\
			& 500                &    10.29 & 4.64 & 5.94 & 17.35 & 0.73 & 0.00  \\
			& 1000               &   8.69 & 3.02 & 5.52 & 14.62 & 1.10 & 0.00\\ \hline
			\multirow{5}{*}{\textbf{FLAT (Rotation)}}       & 25                 &2.53 & 0.89 & 0.75 & 2.18 & 0.00 & 0.00 \\
			& 50                 & 4.73 & 1.66 & 1.59 & 4.30 & 0.01 & 0.00 \\
			& 100             &  2.35 & 0.80 & 0.61 & 2.03 & 0.01 & 0.00  \\
			& 500                & 1.77 & 0.62 & 0.75 & 2.70 & 0.01 & 0.00  \\
			& 1000               &   
			1.03 & 0.27 & 0.21 & 1.11 & 0.00 & 0.00 \\ \hline
			\multirow{5}{*}{\textbf{FLAT (Full)}} & 25& 0.36 & 0.13 & 0.14 & 0.40 & 0.00 & 0.00 \\
			& 50                 & 0.58 & 0.18 & 0.17 & 0.58 & 0.00 & 0.00\\
			& 100                & 0.19 & 0.01 & 0.02 & 0.26 & 0.00 & 0.00\\
			& 500                & 0.17 & 0.01 & 0.02 & 0.09 & 0.00 & 0.00 \\ 
			& 1000               & 
			0.07 & 0.02 & 0.02 & 0.11 & 0.01 & 0.00\\  \hline
		\end{tabular}
	\end{center}
\end{table*}

\end{document}